%% file: main.tex
\documentclass[11pt]{article}
\usepackage[final]{acl}
\usepackage{times}
\usepackage{latexsym}
\usepackage[T1]{fontenc}
\usepackage[utf8]{inputenc}
\usepackage{microtype}
\usepackage{inconsolata}
\usepackage{graphicx}
\usepackage{booktabs}
\usepackage{array}
\usepackage{multirow}
\usepackage{amsmath,amssymb}
\usepackage{enumitem}
\usepackage{xcolor}

\newcommand{\lpp}{\textsc{LPP}}
\input{macros}  

\title{Latent-Predisposition Personas for Heterogeneous and Reliable LLM Survey Simulation}

\author{
  Geonwoo Bang \\
  ETH Zürich \\
  \texttt{gebang@ethz.ch}
  \And
  Dongho Kim \\
  Sungkyunkwan University \\
  \texttt{dpl9753@g.skku.edu}
  \And
  Moohong Min\thanks{Corresponding author.} \\
  Sungkyunkwan University \\
  \texttt{iceo@skku.edu}
}

\begin{document}
\maketitle
\begin{abstract}
Simulated publics from LLM personas should match more than marginal distributions; they also need within-respondent coherence and stability.
Drawing on research into mass belief systems, we show that standard persona prompting produces overly constrained publics. Across three model families and three U.S.\ surveys, opinions on unrelated issues are \DIAGcross\ as correlated as in human samples, and party identification explains \DIAGpid\ as much response variance.
Common remedies (temperature, diversity instructions, backstories) weaken these correlations mainly by adding response noise. On Qwen, their ten-day test-retest reliability drops to \N{qwen-sel-rel-pct} of the human level, and forcing consistent answers undoes their correlation gains.
We propose Latent-Predisposition Personas (LPP), which introduces stable individual differences rather than noise. Each persona receives political engagement, value priorities, and party-relative issue deviations that occur more often as engagement falls. Response distributions are sharpened for consistency and calibrated to toplines and party crosstabs.
At selected operating points, LPP is never significantly worse than any baseline, and is far more reliable than any baseline that matches its correlation error. At matched reliability (human-level or lower), it achieves the lowest correlation error across all six open-model settings.
Even so, errors still exceed those between human samples, and gains on framing effects and a frontier closed model remain partial.
\end{abstract}

\begin{figure}[t]
\centering
\includegraphics[width=0.92\columnwidth]{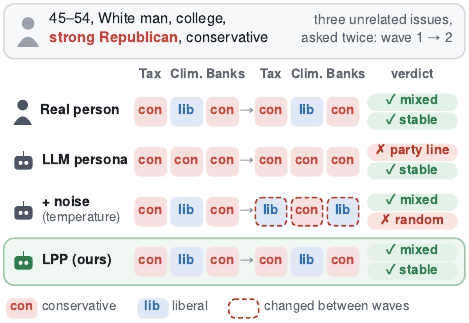}
\caption{\textbf{Standard personas follow the party line; adding noise trades rigidity for instability.} Schematic responses of one respondent asked about three issues twice. \lpp\ instead assigns each persona stable, issue-specific deviations.}
\label{fig:teaser}
\end{figure}

\section{Introduction}
Large language models are increasingly used to simulate survey respondents \citep{argyle2023out,aher2023using}. A model is told who a person is (``You are a 45-year-old White man \dots\ a strong Republican'') and asked the questions a pollster would ask.
Aggregated across many personas, its responses form a \emph{simulated public} for pre-testing questionnaires or anticipating reactions to messages.
Beyond marginal counts, such a public is useful only if it captures how opinions hang together within individuals: which views co-occur, how much partisans disagree, and how stable opinions stay over time.

Persona-conditioned LLMs get this internal structure wrong in a way that public-opinion research directly explains.
In real mass publics, belief systems are loosely constrained: most citizens form opinions issue by issue, while ideological consistency is concentrated among the politically engaged \citep{converse1964nature,zaller1992nature,baldassarri2008partisans}.
Simulated publics invert this pattern. Across three model families and three surveys, opinions on unrelated issues are \DIAGcross\ as correlated as in human samples, party identification explains \DIAGpid\ as much response variance, and within-party heterogeneity is compressed (Section~\ref{sec:diagnosis}).

Common remedies introduce a second problem.
Raising the sampling temperature, instructing the model to avoid stereotypes, or adding backstories lowers cross-issue correlations mainly by making each persona answer more randomly. Real respondents, by contrast, mostly give the same answer when asked twice (Figure~\ref{fig:teaser}).
On the ten-day CEK panel (Section~\ref{sec:setup}), these remedies achieve only \N{qwen-sel-rel-pct} of human test-retest reliability on Qwen, causing a simulated two-wave panel to show roughly twice as much opinion change as humans do (Section~\ref{sec:reliability}). Once constrained to answer consistently, the same personas lose their correlation gains (Figure~\ref{fig:frontier}).
Realistic heterogeneity therefore requires stable differences across individuals rather than within-respondent noise.

We propose Latent-Predisposition Personas (\lpp).\footnote{Code: \url{https://github.com/g7199/lpp-personas}}
Beyond a demographic and partisan profile, each simulated respondent receives three latent predispositions drawn from a fixed prior: a level of political engagement, priorities over six basic values \citep{schwartz1992universals}, and issue-specific deviations from their party's line. Following Converse's account, these deviations occur more frequently as engagement falls.
We then sharpen each persona's response distribution for consistency and apply a monotone transport map to match population toplines and party crosstabs. No individual-level survey answers enter the personas.

Our contributions are:
\begin{enumerate}[leftmargin=*,nosep]
\item A belief-systems diagnosis of persona simulation showing excess cross-issue constraint, inflated party separation, and compressed within-party heterogeneity.
\item Test-retest reliability as an overlooked evaluation axis for simulated publics, examined by tracing correlation error against reliability.
\item \lpp, which achieves the lowest correlation error at matched reliability across all six open-model settings (two model families, three surveys) compared with six baseline persona types given the same post-processing options. Tuned on a single development split, its configuration transfers to two additional surveys and a second model family with only the issue-domain names changed.
\end{enumerate}

\section{Related Work}
\label{sec:related}

\paragraph{Simulated respondents.}
Conditioning LLMs on respondent covariates to create ``silicon samples'' was popularized by \citet{argyle2023out} and \citet{aher2023using}, alongside opinion-alignment benchmarks such as OpinionQA and GlobalOpinionQA \citep{santurkar2023whose,durmus2023towards}.
Subsequent work shows that simulated samples understate variance, flatten or caricature groups \citep{bisbee2024synthetic,cheng2023compost,wang2025large}, and remain sensitive to survey-design artifacts \citep{dominguez2024questioning,tjuatja2024llms}, while LLMs describe group distributions better than they simulate them \citep{meister2025benchmarking}.
Fine-tuning on survey data \citep{suh2025language} and interview-grounded agents \citep{park2024generative} improve fidelity using respondent-level records; we instead target joint public structure without them.

\paragraph{Correlation structure.}
Although LLMs predict aggregate pairwise attitude associations fairly well \citep{ma2025pairwise}, simulated publics struggle to match item correlations: backstory personas still leave correlation-matrix error \citep{moon2024virtual}, and neither persona prompting nor fine-tuning reproduces World Values Survey correlations \citep{williams2026beyond}.
Elsewhere, simulated philosophers over-correlate across questions with DPO helping only partly \citep{shi2026collapse}, models answer ``too consistently'' \citep{chidichimo2026artificial}, and an LLM-free Gaussian copula matches the best LLM respondents \citep{lukauskas2026psychometric}.
Prior attempts to add diversity rely on richer backstories \citep{argyle2023out,moon2024virtual}, instructions, sampling temperature, hidden-state interventions \citep{wang2026psii}, or importance sampling against reference psychometric distributions \citep{hu2025population}.
We diagnose this over-correlation through mass belief systems, connect it to a reliability deficit, and address both with a generative prior calibrated solely to aggregate tabulations.

\paragraph{Mass belief systems.}
Our diagnosis and prior draw on three findings in public-opinion research. \citet{converse1964nature} argued that mass belief systems are weakly constrained, with ideological constraint concentrated among the politically engaged. \citet{achen1975mass} and \citet{ansolabehere2008strength} attribute part of apparent response instability to measurement error rather than non-attitudes, while \citet{baldassarri2008partisans} document partisans who sort without becoming ideologically constrained.

\section{Data, Models, and Metrics}
\label{sec:setup}

\paragraph{Surveys.}
\textbf{CEK} is the control arm of \citet{coppock2018long} from the archive of \citet{coppock2023persuasion}: 622 MTurk respondents answered 14 seven-point items across five domains (climate, financial regulation, infrastructure spending, veterans' care, and taxes), with most retested ten days later.
\textbf{GSS} uses 856 national probability respondents from the 2024 General Social Survey \citep{gss2024} answering 12 items (2--7 categories) across five domains (redistribution, crime and guns, abortion and sexual morality, civil liberties, and confidence in institutions), paired with two-year retest reliabilities from the 2006--08 GSS panel \citep{gsspanel2006}.
\textbf{HM} is the control arm of \citet{hopkins2017assessing}, replicated on GfK and MTurk \citep{coppock2023persuasion}: 1,284 respondents rated federal spending across five domains, with most completing a follow-up wave; its four framed arms supply our treatment-effect test (Section~\ref{sec:dynamic}).
We randomly split each survey into development and test halves. All hyperparameters are chosen on the CEK development half; GSS and HM development halves are used only for calibration tabulations and the human floor.

\paragraph{Models.}
Our primary model is Qwen3.8-27B \citep{qwen38} (instruction-tuned, NF4 4-bit). Gemma-4-26B-A4B \citep{gemmateam2026gemma4} (MoE, 4B active parameters, FP8) tests transfer to a second open model family under identical hyperparameters. As a frontier closed model, we include GPT-6.1-Sol \citep{openai2026gpt61sol} via isolated, read-only OpenAI Codex CLI sessions (\texttt{gpt-6.1-sol}, October 2026, low reasoning effort; Appendix~\ref{app:impl}); without token probabilities, it answers all items per persona in one call and sharpening does not apply (Section~\ref{sec:transfer}). For open models, we read the next-token distribution over option digits (over 99\% of probability mass on average). Each model-survey pair is a \emph{setting}.

\paragraph{Structure metrics.}
\textbf{Correlation error} (corr$_F$), our primary metric, is the Frobenius norm between the upper triangles of simulated and human item correlation matrices.
\textbf{Party-conditional energy distance} (ED$_{\text{party}}$) averages the energy distance \citep{szekely2013energy} between simulated and human joint response vectors (standardized by human item SDs) within Democrats and Republicans (zero iff joint distributions coincide).
To pinpoint discrepancies, we also report a Converse decomposition of five ratios to human values (1 = human, below 1 = underestimated): cross-domain mean $|r|$ (``Cross''), within-domain mean $|r|$, the Democrat--Republican gap, within-party SD, and variance share explained by party ID.

\paragraph{Reliability.}
For item $j$, reliability is the test-retest correlation between two independent responses from one persona. After all post-processing, we compute it directly as the between-persona variance share $\operatorname{Var}_r(\mathbb E[x_{rj}])/(\operatorname{Var}_r(\mathbb E[x_{rj}])+\mathbb E_r\operatorname{Var}(x_{rj}))$, average over items, and divide by the human wave-1/wave-2 correlation (253 follow-up test respondents in CEK, 506 in HM, and the 2006--08 panel for GSS). Because human retests include genuine opinion change, this benchmark is a lower bound. We treat $\geq 0.9$ as \emph{human-level}, while our selection criterion (Section~\ref{sec:protocol}) penalizes deviations from 1 in either direction. We also retest with reworded questions (Section~\ref{sec:main}). Tables report this ratio as ``Rel.''

\paragraph{Uncertainty and floor.}
We compute all metrics across 200 random test half-samples, paired across methods, drawing one answer per persona per sample.
The human \emph{floor} applies the same metric between an equal-sized subsample from the development half and the same test half-sample, capturing the sampling error even a perfect simulator would incur.
Table markers ($^\dagger$) are one-sided and uncorrected. For the matched-reliability comparison, we instead report the share of half-samples in which \lpp\ beats every baseline simultaneously, bounding all pairwise comparisons jointly.

\section{Diagnosis}
\label{sec:diagnosis}

Each model is prompted with the demographic profile of every test respondent, namely age, gender, race, education, party identification, and ideology. We call this the Demographic persona. For each item, one answer is drawn from the model's response distribution. Table~\ref{tab:diag} reports four statistics of the resulting public as ratios to the human value, and its caption gives the range of the same ratio between two human samples (the human floor).

\begin{table}[t]\centering\small\setlength{\tabcolsep}{3pt}
\resizebox{\columnwidth}{!}{\begin{tabular}{lccccccc}
\toprule
 & Qwen & Qwen & Qwen & Gemma & Gemma & Gemma & GPT \\
 & CEK & GSS & HM & CEK & GSS & HM & CEK \\ \midrule
Cross-domain $|r|$ & 1.73 & 2.36 & 1.58 & 2.80 & 3.23 & 2.04 & 4.65 \\
Party gap & 1.84 & 1.39 & 1.55 & 1.76 & 1.56 & 1.76 & 2.23 \\
Within-party SD & 0.91 & 0.83 & 0.87 & 0.56 & 0.63 & 0.62 & 0.41 \\
Party $R^2$ & 3.18 & 2.46 & 2.74 & 4.30 & 3.21 & 3.70 & 7.56 \\
\bottomrule
\end{tabular}}
\caption{Demographic personas without post-processing: simulated-public statistics relative to the human test half (1 = human). Human floor (two disjoint human samples): Cross-domain $|r|$ 0.98--1.08; Party gap 0.98--1.09; Within-party SD 1.00--1.04; Party $R^2$ 1.01--1.20.}
\label{tab:diag}
\end{table}

Three patterns appear in every column.
\textbf{(i) Excess cross-issue constraint.} Responses across unrelated domains (e.g., climate policy and veterans' health care) are \DIAGcross\ as correlated as in human samples. Within-domain correlations, which are already high among humans, are inflated less (\DIAGwithin\ on CEK and GSS; HM has one item per domain).
\textbf{(ii) Inflated partisan separation.} The Democrat--Republican gap is overly wide, and party identification accounts for \DIAGpid\ as much response variance as in human data, making partisanship a near-sufficient statistic for each simulated respondent.
\textbf{(iii) Compressed within-party heterogeneity.} Within-party standard deviations fall below human levels, dropping to roughly 60\% on Gemma and under half on GPT for CEK.
In Converse's terms, the simulated public behaves like a population of ideologues.
Across our three models, this failure does not ease with scale: GPT-6.1-Sol, which answers all items for a persona in a single call, is the most extreme on every statistic, and Gemma-4-26B-A4B shows tighter cross-issue constraint than the similarly sized Qwen3.8-27B on every survey.

Neither diversity instructions (Appendix~\ref{app:diag}) nor marginal calibration fixes this structure. Aligning each item's marginal to the human topline still leaves cross-issue constraint well above human levels (\N{cg-qwen-pool-demo-cross}$\times$ on Qwen and \N{cg-gemma-pool-demo-cross}$\times$ on Gemma for CEK; Appendix Table~\ref{tab:main}). Calibrating within each party eliminates the excess party gap by construction, yet correlation error remains well above the human floor (Section~\ref{sec:results}). Because a per-item, per-party map only rescales marginals, it cannot supply within-party dependence structure that the personas themselves lack.

\section{Method}
\label{sec:method}

\begin{figure*}[t]
\centering
\includegraphics[width=0.92\textwidth]{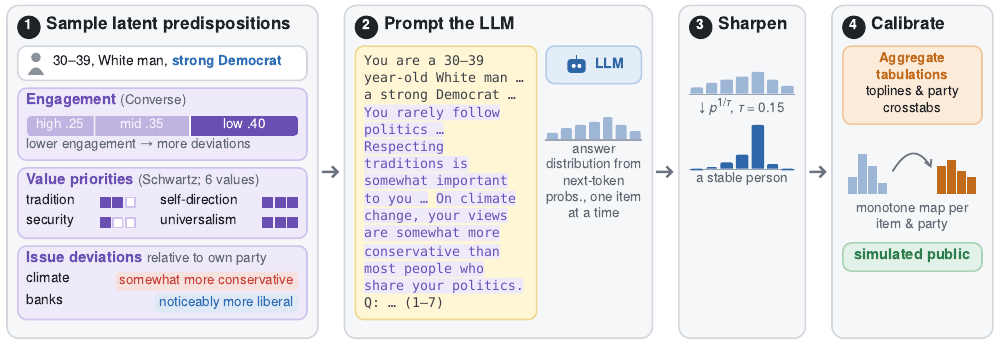}
\caption{\textbf{Latent-Predisposition Personas.} (1)~Latents drawn from a fixed prior: engagement, value priorities, and party-relative issue deviations. (2)~Response distributions read from next-token probabilities, (3)~sharpened for consistency, and (4)~calibrated to toplines and party crosstabs. No individual-level survey answers are used.}
\label{fig:overview}
\end{figure*}

Let $r$ index target respondents with observed covariates $c_r$ (age, gender, race, education, party identification, and ideology), and let $j=1,\dots,J$ index survey items with ordered options $\{1,\dots,K_j\}$.
A persona simulator maps $c_r$ to a prompt and reads out a response distribution $p_{rj}\in\Delta^{K_j-1}$ from the model (next-token probabilities over the option digits for open models, or a single sampled answer for API models).
The simulated population is formed by drawing one answer $x_{rj}\sim p_{rj}$ per persona and item.
\lpp\ retains this readout (step~2 in Figure~\ref{fig:overview}) and adds three components, each motivated by an empirical feature of public opinion.

\paragraph{Latent predispositions (step 1).}
For each respondent, we draw latent attributes from a fixed prior seeded by their identifier, independently of their survey answers:
\begin{itemize}[leftmargin=*,nosep]
\item \emph{Political engagement} $e_r\in\{\text{high},\text{mid},\text{low}\}$ with prior $(.25,.35,.40)$, expressed as a brief statement (e.g., ``You rarely follow politics. You form opinions issue by issue \dots'').
\item \emph{Value priorities} $v_{rk}$ over six broad values $k$ \citep{schwartz1992universals}: tradition, self-direction, security, benevolence, achievement, and universalism. Each value is drawn uniformly and independently as ``not very'', ``somewhat'', or ``very'' important. Values cut across party lines, giving issue positions a shared, non-partisan basis on which to covary.
\item \emph{Issue deviations} $z_{rd}\sim\mathcal N(0,1)$ for each issue domain $d$, discretized into five bins using engagement-dependent thresholds at $\pm0.4k_e$ and $\pm1.2k_e$ (defined below) and rendered as ``On \emph{[domain]}, your views are \{noticeably, somewhat\} more \{liberal, conservative\} than / about the same as most people who share your general politics''. Because deviations are relative to the respondent's own party, they preserve party structure while preventing the party line from acting as a sufficient statistic, echoing ``partisans without constraint'' \citep{baldassarri2008partisans}.
\end{itemize}
\emph{Engagement coupling.} Because ideological constraint concentrates among engaged citizens \citep{converse1964nature,zaller1992nature}, we tie $k_{e_r}$ to engagement: highly engaged personas state no deviations ($k_{\text{high}}=\infty$), moderately engaged personas deviate occasionally ($k_{\text{mid}}=1.6$, 48\% of domains ``about the same as''), and low-engagement personas deviate often ($k_{\text{low}}=0.6$, 19\% of domains). Low engagement thus yields idiosyncratic yet stable opinions rather than random non-attitudes \citep{achen1975mass,ansolabehere2008strength}, consistent with our data where human retest reliability does not decline among weaker partisans (Appendix~\ref{app:robust}). Latents are never fit to individuals; the prior's only free parameters are the engagement shares and $k_e$.

\paragraph{Sharpening (step 3).}
LLM output distributions are overly broad: two draws from one persona agree less often than a human asked twice (Section~\ref{sec:reliability}), and noise remedies worsen this. Once latents supply between-persona heterogeneity, we sharpen each distribution as $\tilde p_{rj}\propto p_{rj}^{1/\tau}$ with $\tau<1$.

\paragraph{Calibration (step 4).}
For each item, we compute the one-dimensional monotone (optimal) transport plan between the pooled simulated marginal $\bar p_j=\frac1n\sum_r\tilde p_{rj}$ and a reference human marginal $q_j$ \citep{peyre2019computational}, row-normalize it into a stochastic matrix $M_j$, and apply it to each persona as $\hat p_{rj}=\tilde p_{rj}M_j$.
Because this transport map is monotone, it preserves the stochastic ordering of personas on every item (if $\tilde p_{rj}$ first-order dominates $\tilde p_{sj}$, then $\hat p_{rj}$ dominates $\hat p_{sj}$). It therefore retains most of the dependence induced by the latents while matching the pooled marginal $q_j$ exactly.
Under party-crosstab calibration, the map is fit separately within Democrats, independents, and Republicans, using only aggregate tabulations from a separate human sample.

\section{Experimental Protocol}
\label{sec:protocol}

\paragraph{Baselines.}
\textbf{Demographic} personas use only the profile sentences. \textbf{Diversity instruction} appends ``real people often hold mixed views that do not follow a single party line \dots''; we also test the best of five \textbf{wordings} selected on the CEK development half (Appendix~\ref{app:prompts}). Two backstory baselines add a first-person \textbf{Argyle} backstory \citep{argyle2023out} or a profile-conditioned \textbf{life-story} backstory. \textbf{Anthology} \citep{moon2024virtual} generates unconditional ``Tell me about yourself'' backstories, infers demographics, and matches them to respondents via maximum-weight bipartite matching. Finally, we test \textbf{temperature} scaling on Demographic responses and aggregate-only \textbf{reweighting} of Demographic personas to match reference marginals \citep[in the spirit of][]{hu2025population}.

\paragraph{Calibration.}
All baselines receive the same monotone transport calibration as \lpp\ (Section~\ref{sec:method}) using standard aggregate tabulations: \emph{toplines} (per-item marginals) and \emph{party crosstabs} (marginals within Democrats, independents, and Republicans). We report party-crosstab results unless stated otherwise. Tabulations come from each survey's development half, which is optimistic relative to out-of-sample tabulations; we test cross-wave GSS tabulations in Section~\ref{sec:transfer}.

\paragraph{Selection.}
Both the \lpp\ configuration (latents, thresholds $k_e$, and sharpening, yielding $\tau{=}0.15$) and each baseline's post-processing are selected on the CEK development half by one criterion: the sum of correlation error and reliability error (plus ED$_{\text{party}}$ under topline calibration), each normalized against the calibrated Demographic baseline and averaged across both calibration levels (Appendix~\ref{app:impl}). For \lpp, we searched 32 persona configurations across the same 12 post-processing options given to every baseline; baseline texts have no free parameters beyond the five diversity wordings, and unselected \lpp\ variants perform similarly (Section~\ref{sec:ablations}). For GSS and HM, the configuration is frozen except for a short phrase naming each issue domain (Appendix~\ref{app:prompts}).

\begin{figure*}[t]
\centering
\includegraphics[width=0.82\textwidth]{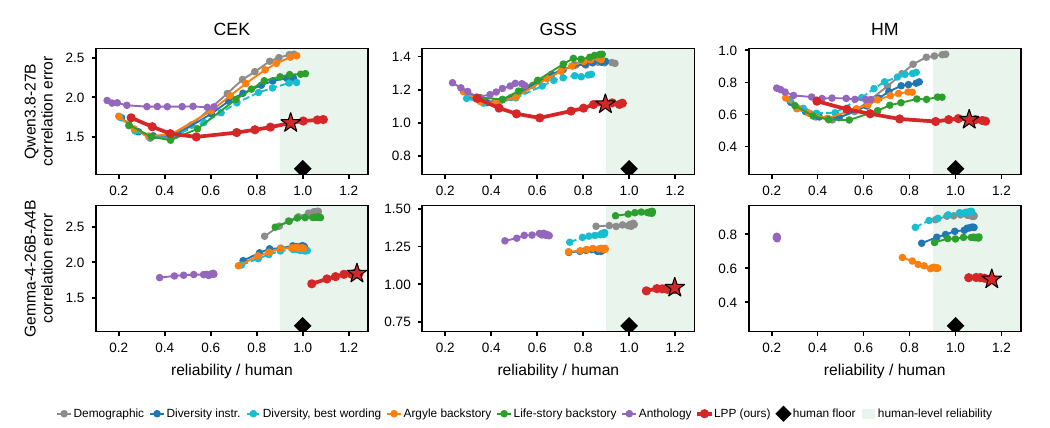}
\caption{Correlation error vs.\ reliability across all six settings (test halves, party-crosstab calibration). Curves sweep each calibrated persona from $T{=}3$ (left) to $\tau{=}0.01$ (right); on Gemma, temperature lowers reliability only slightly. Shaded: Rel.\ $\geq 0.9$; star: selected \lpp; diamond: human floor.}
\label{fig:frontier}
\end{figure*}

\paragraph{Comparison regimes.}
\emph{Selected baselines} use the post-processing chosen above ($T\in\{1.5,2,3,4\}$, none, or $\tau\in\{0.05,\dots,0.7\}$), whereas \emph{sharpened baselines} use the sharpening closest to human reliability on the development half ($\tau{=}0.05$ for all baselines). Our primary comparison is at \emph{matched reliability}. Random answering pulls correlations toward zero, lowering corr$_F$ in an over-constrained public without metric penalty; matching reliability removes this shortcut. Although reliability cannot be set directly, sweeping the temperature/sharpening scale from $T{=}3$ (softer, lower reliability) to $\tau{=}0.01$ (sharper, higher reliability) traces each persona's trade-off curve (Figure~\ref{fig:frontier}); we interpolate corr$_F$ at 0.9 of human reliability and at 0.8.

\section{Results}
\label{sec:results}

\subsection{Main results}
\label{sec:main}

Table~\ref{tab:main-otp} compares all methods on the CEK test half under party-crosstab calibration. Because calibration matches each method's item marginals within each party by construction, differences across methods reflect dependence structure and reliability.

\paragraph{Selected operating points.}
On Qwen, even though the selection criterion penalizes low reliability, every selected baseline remains well below human reliability (\N{cg-qwen-otp-sel-rel} of the human level), reaching correlation errors of \N{cg-qwen-otp-sel-corr}. \lpp\ achieves a correlation error of \N{cg-qwen-otp-lpp-corr} (human floor \N{cg-qwen-otp-floor-corr}) at \N{cg-qwen-otp-lpp-rel} of human reliability, while keeping cross-domain constraint close to humans (\N{cg-qwen-otp-lpp-cross} vs.\ a human floor of \N{cg-qwen-otp-floor-cross}).
Although the Qwen/CEK baselines fall within sampling noise of \lpp's correlation error (paired differences \N{cg-qwen-otp-sel-drange}, all 95\% intervals including zero), they do so only by answering \N{cg-qwen-otp-sel-relx}$\times$ less consistently. Across all six settings (Table~\ref{tab:transfer-otp}), \lpp\ ties for or achieves the lowest mean error in five; its gain is significant on GSS with Gemma, and it is never significantly worse than any selected baseline. By contrast, an LLM-free sampler drawing independently from each party's reference marginal severely under-correlates (cross-domain constraint \N{cg-qwen-otp-agg-cross}) and trails \lpp\ in every setting (Appendix~\ref{app:diag}).

\paragraph{Sharpened baselines.}
Sharpening baseline personas to human-level reliability (lower block of Table~\ref{tab:main-otp}) collapses each party onto its modal response pattern. On Qwen, correlation error climbs to \N{cg-qwen-otp-shp-corr}, with \lpp\ outperforming every sharpened baseline in \N{cg-qwen-otp-shp-pwinphrase} paired half-samples.
Figure~\ref{fig:frontier} traces the full trade-off across all six settings. On Qwen, baseline curves are either U-shaped with minima well below human reliability or terminate before reaching it; on Gemma, whose raw response distributions are already sharper, baseline errors generally rise with reliability. By contrast, the \lpp\ curve stays relatively flat across sharpening levels, reaches human reliability, and sits below every baseline there in all six settings. Sharpening alone cannot produce this gain (Demographic + sharpening in Table~\ref{tab:abl}), as it only preserves existing heterogeneity across personas.

\paragraph{Matched reliability.}
Interpolated at \MATCHrt\ of human reliability (Table~\ref{tab:transfer-otp}), \lpp\ achieves the lowest correlation error in all six settings, beating every reaching baseline in at least 95\% of paired half-samples in five settings and in \MATCHminp\ on HM with Gemma (where the Argyle backstory comes closest). The 95\% interval for the difference against the best reaching baseline excludes zero on CEK and GSS for both models (Appendix Table~\ref{tab:matched}). In \MATCHnun\ of the \MATCHnb\ baseline-setting pairs, no post-processing down to $\tau{=}0.01$ reaches 0.9 reliability (because calibration reintroduces noise to match the target marginal; Appendix~\ref{app:robust}), though at least one baseline reaches it in every setting. At a lower target of 0.8, which five of six baselines reach in every setting, \lpp\ again posts the lowest error across all six settings (Appendix Table~\ref{tab:matched}).

\paragraph{Robustness.}
All \lpp\ results average over six independent latent draws (half-sample $i$ uses draw $i \bmod 6$), so every uncertainty interval incorporates seed variance. On CEK/Qwen, individual draws span \RB{seed-qwen-min}--\RB{seed-qwen-max}, well below the best matched-reliability baseline (\RB{qwen-bestbase}). Under retest with reworded questions, every persona type loses some reliability, yet \lpp\ retains the most (reaching \RB{para-qwen-lppmax} vs.\ $\leq$\RB{para-qwen-blmax} for baselines on Qwen, and \RB{para-gemma-lppmax} vs.\ $\leq$\RB{para-gemma-blmax} on Gemma). Its stability therefore does not depend on repeating the exact same wording.

\begin{table}[t]\centering\small\setlength{\tabcolsep}{3.5pt}
\resizebox{\columnwidth}{!}{\begin{tabular}{lcccccc}
\toprule
 & \multicolumn{3}{c}{Qwen3.8-27B} & \multicolumn{3}{c}{Gemma-4-26B-A4B} \\
 & corr$_F$ & Rel. & Cross & corr$_F$ & Rel. & Cross \\ \midrule
Human floor & 1.10 & 1.00 & 1.06 & 1.10 & 1.00 & 1.07 \\ \midrule
Aggregate sampler (no LLM) & 1.97 & 0.13$^\dagger$ & 0.49 & 1.97 & 0.13$^\dagger$ & 0.49 \\
Demographic & 1.65 & 0.52$^\dagger$ & 1.21 & 2.64$^\dagger$ & 0.98 & 2.03 \\
\multicolumn{7}{l}{\emph{Selected baselines}} \\
\quad Demographic + reweighting & 1.67 & 0.52$^\dagger$ & 1.23 & 2.66$^\dagger$ & 0.98 & 2.04 \\
\quad Demographic + temperature & 1.48 & 0.34$^\dagger$ & 0.86 & 2.51$^\dagger$ & 0.90 & 1.92 \\
\quad Diversity instruction & 1.52 & 0.44$^\dagger$ & 1.06 & 2.20$^\dagger$ & 0.90 & 1.72 \\
\quad Diversity, best of 5 wordings & 1.49 & 0.42$^\dagger$ & 1.00 & 2.15 & 0.90 & 1.72 \\
\quad Argyle backstory & 1.51 & 0.33$^\dagger$ & 0.86 & 2.14 & 0.85 & 1.68 \\
\quad Life-story backstory & 1.46 & 0.43$^\dagger$ & 1.02 & 2.62$^\dagger$ & 0.98 & 1.95 \\
\quad Anthology & 1.87 & 0.59$^\dagger$ & 0.66 & 1.84 & 0.61$^\dagger$ & 0.69 \\
\multicolumn{7}{l}{\emph{Sharpened baselines ($\tau{=}0.05$)}} \\
\quad Demographic & 2.54$^\dagger$ & 0.94 & 1.66 & 2.71$^\dagger$ & 1.07 & 2.08 \\
\quad Diversity instruction & 2.24$^\dagger$ & 0.93 & 1.59 & 2.23 & 1.00 & 1.73 \\
\quad Argyle backstory & 2.51$^\dagger$ & 0.95 & 1.86 & 2.20 & 1.00 & 1.69 \\
\quad Life-story backstory & 2.29$^\dagger$ & 0.99 & 1.65 & 2.63$^\dagger$ & 1.08 & 1.95 \\
\midrule
\lpp\ (ours) & 1.68 & 0.95 & 1.10 & 1.84 & 1.24 & 1.30 \\
\bottomrule
\end{tabular}}
\caption{CEK test half under party-crosstab calibration (means over 200 half-samples). Rel.\ and Cross are ratios to human values (1 = human). $^\dagger$: \lpp\ outperforms the baseline in $\geq 97.5\%$ of paired half-samples (for Rel., closer to 1).}
\label{tab:main-otp}
\end{table}

\begin{table*}[t]\centering\small\setlength{\tabcolsep}{5pt}
\begin{tabular}{llc cc cc ccc cc}
\toprule
 & & Floor & \multicolumn{2}{c}{Selected baselines} & \multicolumn{2}{c}{Sharpened baselines} & \multicolumn{3}{c}{\lpp} & \multicolumn{2}{c}{At Rel.\ 0.9} \\
\cmidrule(lr){4-5}\cmidrule(lr){6-7}\cmidrule(lr){8-10}\cmidrule(lr){11-12}
Survey & Model & corr$_F$ & corr$_F$ & Rel. & corr$_F$ & Rel. & corr$_F$ & Rel. & Cross & \lpp & Best bl. \\ \midrule
CEK & Qwen & 1.10 & 1.46--1.87 & 0.33--0.59 & 2.24--2.54$^\dagger$ & 0.93--0.99 & {1.68} & 0.95 & 1.10 & \textbf{1.65} & 2.15 \\
GSS & Qwen & 0.72 & 1.12--1.24 & 0.41--0.58 & 1.36--1.41$^\dagger$ & 0.87--0.92 & \textbf{1.11} & 0.90 & 1.10 & \textbf{1.11} & 1.36 \\
HM & Qwen & 0.26 & 0.57--0.69 & 0.36--0.61 & 0.71--0.97 & 0.79--0.94 & \textbf{0.57} & 1.06 & 0.91 & \textbf{0.56} & 0.70 \\
CEK & Gemma & 1.10 & 1.84--2.66 & 0.61--0.98 & 2.20--2.71 & 1.00--1.08 & \textbf{1.84} & 1.24 & 1.30 & \textbf{1.70} & 2.16 \\
GSS & Gemma & 0.72 & 1.23--1.47$^\dagger$ & 0.65--1.02 & 1.23--1.48$^\dagger$ & 0.88--1.10 & \textbf{0.98} & 1.20 & 1.03 & \textbf{0.95} & 1.39 \\
HM & Gemma & 0.26 & 0.62--0.92 & 0.22--1.02 & 0.60--0.90 & 0.92--1.10 & \textbf{0.54} & 1.16 & 0.97 & \textbf{0.54} & 0.60 \\
\bottomrule
\end{tabular}
\caption{All six settings (test halves, party-crosstab calibration; \lpp\ frozen from CEK/Qwen). Ranges span seven selected and four sharpened baselines; GSS Rel.\ uses a two-year panel. $^\dagger$: \lpp\ beats every baseline in the group in $\geq 97.5\%$ of paired half-samples. ``At Rel.\ 0.9'': corr$_F$ interpolated on Figure~\ref{fig:frontier} for \lpp\ and the best reaching baseline.}
\label{tab:transfer-otp}
\end{table*}

\subsection{Spurious opinion change}
\label{sec:reliability}

In a simulated two-wave panel, low reliability manifests as spurious opinion change. For each persona and item, we check whether two independent responses fall on opposite sides of the scale midpoint, and compare this switching rate against CEK respondents surveyed ten days apart.
Humans switch sides in \N{churn-human} of cases; on Qwen, the selected baselines switch in \N{churn-sel} (\N{churn-sel-ratio}$\times$ as often), meaning a simulated panel built on them would register substantial opinion change where none occurred. \lpp\ personas switch in \N{churn-lpp} of cases (\N{churn-lpp-ratio}$\times$ the human rate). On Gemma, whose raw distributions are sharper, every persona type except Anthology switches less often than humans (0.4\% for \lpp).

\subsection{Generalization}
\label{sec:transfer}

Table~\ref{tab:transfer-otp} transfers the configuration selected on CEK/Qwen, unchanged apart from one phrase naming each issue domain (Appendix~\ref{app:prompts}), to two additional surveys and a second open model family.

\paragraph{New surveys.}
On GSS, whose items differ from CEK in topic, format, and decade, \lpp\ achieves the lowest correlation error on both models (Table~\ref{tab:transfer-otp}). When calibrated instead to tabulations from the 2022 GSS wave, it retains the lowest error at human-level reliability on both models, as well as at the selected operating point on Gemma (Appendix~\ref{app:robust}).
On HM, which has one item per domain, \lpp\ operates at or slightly above human reliability: on Qwen it matches the best selected baseline's correlation error at much higher reliability, and on Gemma it beats every selected and sharpened baseline (\N{hm-gemma-otp-lpp-corr} vs.\ $\geq 0.60$).

\paragraph{Other model families.}
On CEK, Gemma's \lpp\ personas exceed human reliability (\N{cg-gemma-otp-lpp-rel}) at a correlation error of \N{cg-gemma-otp-lpp-corr}, matching or beating every selected baseline (\N{cg-gemma-otp-sel-corr}). The only baseline that ties it, Anthology, reaches just \N{cg-gemma-otp-sel-anth-rel} of human reliability; at human-level reliability, every baseline has an interpolated error of at least \RB{gemma-bestbase} (Table~\ref{tab:transfer-otp}). On GPT-6.1-Sol, where responses are sampled directly and cannot be sharpened, \lpp\ wins under topline calibration (\N{cg-gpt-pool-lpp-corr} vs.\ \N{cg-gpt-pool-sel-div-corr}) but trails the diversity instruction under party crosstabs (\N{cg-gpt-otp-lpp-corr} vs.\ \N{cg-gpt-otp-sel-div-corr}): because the model applies each stated deviation uniformly across all items in a domain, it inflates within-domain correlations (Appendix~\ref{app:diag}).

\subsection{Ablations}
\label{sec:ablations}

Appendix Table~\ref{tab:abl} removes or replaces one component of \lpp\ at a time on the CEK test half. Most of the correlation gain comes from persistent individual-level issue deviations, while sharpening supplies stability and engagement coupling safeguards reliability across surveys.
Removing sharpening drops Qwen's reliability to \N{cg-qwen-otp-abl-nosharp-rel}.
Removing issue deviations raises correlation error on both models (to \N{cg-qwen-otp-abl-nodev-corr} on Qwen and \N{cg-gemma-otp-abl-nodev-corr} on Gemma) and restores excess cross-domain constraint.
Value priorities help on Qwen (error rises to \N{cg-qwen-otp-abl-novals-corr} without them) but slightly hurt on Gemma (\N{cg-gemma-otp-abl-novals-corr} without them).
Engagement coupling adds a small correlation penalty on CEK (without it, error falls to \N{cg-qwen-otp-abl-k16-corr} on Qwen and \N{cg-gemma-otp-abl-k16-corr} on Gemma), yet it matters for generalization: without coupling, the GSS/Qwen frontier never reaches human-level reliability, and on HM (where every pair of items crosses domains) uniform deviations pull cross-domain constraint down to \N{hm-qwen-otp-abl-k16-cross} of the human level, whereas coupling keeps it close to human levels (\N{hm-qwen-otp-lpp-cross}, difference \N{hm-qwen-otp-abl-k16-dcross}~\N{hm-qwen-otp-abl-k16-dcross-ci}).
At matched reliability, this unselected uncoupled variant lowers error by at most 0.06 in three settings and is either higher or unable to reach the target in the other three (Appendix Table~\ref{tab:matched}).

\paragraph{Controls.}
The gain is not an artifact of prompt length: replacing latents with placebo attributes of the same format (hobbies and daily routines) raises correlation error on Qwen to \N{cg-qwen-otp-abl-placebo-corr}, trailing \lpp\ in \N{cg-qwen-otp-abl-placebo-pwin} of paired half-samples. Stating deviations in absolute rather than party-relative terms yields a similar error on Qwen (\N{cg-qwen-otp-abl-abs-corr}; Appendix~\ref{app:robust}). Because toplines and crosstabs carry no cross-item dependence information, the closest non-prompting alternative is to apply the same latents as direct numerical offsets to the Demographic response distributions. This recovers most of the gain on Qwen with or without engagement coupling (\RB{off-qwen} at human-level reliability vs.\ \RB{off-qwen-lpp} for \lpp\ and \RB{off-qwen-base} for Demographic), but almost none on Gemma (\RB{off-gemma} vs.\ \RB{off-gemma-lpp} and \RB{off-gemma-base}). Persistent individual-level structure is thus the central ingredient, though on Gemma only the model itself translates these predispositions into coherent responses. Finally, giving each persona a single continuous ideology score matches \lpp's correlation error only far below human reliability on Qwen, and never on Gemma. Additional checks appear in Appendix~\ref{app:robust}.

\subsection{Framing effects}
\label{sec:dynamic}

We simulate 400 respondents per arm across the four HM framing arms \citep{hopkins2017assessing,coppock2023persuasion} (news frames on crime, terrorism, health care, and economic stimulus). Calibration is fit solely on the simulated control arm and applied unchanged to all arms so it cannot create or erase effects. Framed-minus-control mean differences (in human control-arm SD units) form a $4\times5$ effect matrix compared against humans (Appendix Table~\ref{tab:hmdyn}).

Human effects are small (mean on-topic effect \HMDYNhuman\ SD). On Qwen, Demographic personas match the relative pattern ($r{=}\HMDYNbaser$) but overreact more than fivefold (\HMDYNbasedirect~SD). \lpp\ preserves the pattern (\HMDYNlppr) while moderating overreaction (\HMDYNlppdirect\ SD), cutting RMSE from \HMDYNbaseraw\ to \HMDYNlppraw\ without post-processing and from \HMDYNbasesharp\ to \HMDYNlppsharp\ under \lpp's post-processing. Gemma's Demographic personas overreact less (\HMDYNGbasedirect\ SD), and \lpp\ again dampens responses (\HMDYNGlppdirect\ SD), lowering RMSE from \HMDYNGbaseraw\ to \HMDYNGlppraw\ without post-processing and from \HMDYNGbasesharp\ to \HMDYNGlppsharp\ under \lpp's post-processing. All four paired bootstrap differences exclude zero, yet simulated effects remain several times too large across models and methods.

\section{Conclusion}
\label{sec:conclusion}

Persona-conditioned LLMs produce simulated publics that are more constrained and polarized, yet less heterogeneous within parties, than real respondents. Adding response noise does not resolve this trade-off; realistic publics instead require stable individual differences within each party. At selected operating points, no baseline significantly outperforms \lpp, and at matched reliability \lpp\ achieves the lowest correlation error in all six open-model settings. More broadly, evaluations of simulated publics should report test-retest reliability alongside aggregate correlation fit.

\section*{Limitations}

\paragraph{Scope.} All three datasets are English-language U.S.\ political surveys in which partisanship is the dominant cleavage. Whether our prior transfers to multi-party systems or non-political attitudes remains untested.

\paragraph{Operating points.} At the selected operating points, \lpp's advantage in correlation error is statistically significant only on GSS with Gemma. On CEK with Qwen, the life-story backstory achieves a numerically lower error, albeit at less than half of \lpp's reliability, and statistical power is limited with 155--321 respondents per half-sample. On HM with Gemma, \lpp\ achieves the lowest matched-reliability error in \MATCHminp\ of half-samples. In addition, freezing Qwen's sharpening parameter leaves Gemma's personas more stable than human respondents, whose opinions also shift over time; modeling genuine opinion change is beyond our scope.

\paragraph{Requirements.} Each issue domain must be named explicitly, as stated deviations do not carry over to unstated domains (Appendix~\ref{app:diag}). Calibration requires target-population toplines, and our main results also rely on party crosstabs. For closed API models that do not expose token probabilities, sharpening is unavailable and the deviation scale had to be set per model. Our open-model experiments also used quantized weights (4-bit for Qwen and 8-bit for Gemma). Finally, while selecting the \lpp\ configuration used individual-level answers from one development split (CEK), the frozen configuration requires none thereafter.

\paragraph{Reference data.} Human retest benchmarks (the ten-day CEK panel, the HM follow-up, and the two-year GSS panel) reflect genuine opinion change as well as measurement noise and therefore act as lower bounds. CEK and the MTurk subset of HM are convenience samples, though results on the national probability sample in GSS are at least as favorable.

\paragraph{What is matched.} Our approach targets aggregate distributions, item correlations, test-retest reliability, and average treatment effects rather than individual respondents. A simulated public is no substitute for surveying real people in substantive inference.

\section*{Ethics Statement}
All human datasets are public, de-identified replication files released by the original authors or by NORC (GSS), used in accordance with their terms; no new human data were collected.
Simulated publics carry misuse risks, such as pre-testing manipulative messaging at scale or presenting synthetic responses as genuine public opinion. We highlight the remaining discrepancies against human data so that our results are not overstated, and we caution against substituting simulated samples for real surveys, particularly for groups underrepresented in model training data.
Persona prompts include demographic and partisan attributes strictly as survey covariates; we neither evaluate nor generate content about real individuals.

\bibliography{refs}

\appendix
\setlength{\floatsep}{8pt plus 2pt}\setlength{\dblfloatsep}{8pt plus 2pt}\setlength{\textfloatsep}{10pt plus 2pt}\setlength{\dbltextfloatsep}{10pt plus 2pt}\setcounter{dbltopnumber}{4}
\section{Persona Prompts}
\label{app:prompts}
\begin{table*}[t]\centering\small
\begin{tabular}{llc cc cc ccc}
\toprule
 & & Floor & \multicolumn{2}{c}{Selected baselines} & \multicolumn{2}{c}{Sharpened baselines} & \multicolumn{3}{c}{\lpp} \\
\cmidrule(lr){4-5}\cmidrule(lr){6-7}\cmidrule(lr){8-10}
Survey & Model & corr$_F$ & corr$_F$ & Rel. & corr$_F$ & Rel. & corr$_F$ & Rel. & Cross \\ \midrule
CEK & Qwen & 1.10 & 1.59--2.40 & 0.39--0.62 & 2.88--3.77$^\dagger$ & 1.01--1.12 & {1.85} & 0.96 & 1.17 \\
GSS & Qwen & 0.72 & 1.26--1.57 & 0.42--0.62 & 1.68--2.51$^\dagger$ & 0.98--1.21 & \textbf{1.26} & 0.84 & 0.88 \\
HM & Qwen & 0.26 & 0.55--0.85 & 0.44--0.68 & 1.05--1.33$^\dagger$ & 0.97--1.06 & {0.65} & 1.02 & 0.71 \\
CEK & Gemma & 1.10 & 2.26--3.65 & 0.57--1.09 & 2.98--3.73$^\dagger$ & 1.10--1.18 & \textbf{2.05} & 1.25 & 1.44 \\
GSS & Gemma & 0.72 & 1.63--2.47$^\dagger$ & 0.55--1.22 & 1.68--2.50$^\dagger$ & 1.01--1.30 & \textbf{1.08} & 1.24 & 0.83 \\
HM & Gemma & 0.26 & 1.07--1.37$^\dagger$ & 0.02--1.11 & 1.09--1.30$^\dagger$ & 1.11--1.19 & \textbf{0.52} & 1.16 & 0.84 \\
\bottomrule
\end{tabular}
\caption{All settings, test halves, topline calibration; configuration and post-processing frozen from CEK/Qwen. Ranges over the seven selected and four sharpened baselines. GSS Rel.\ is relative to a two-year panel. $^\dagger$: \lpp\ lower than every baseline of the group in $\geq$97.5\% of paired half-samples.}
\label{tab:transfer}
\end{table*}
\begin{table*}[t]\centering\small\setlength{\tabcolsep}{7pt}
\resizebox{\textwidth}{!}{\begin{tabular}{lcccccccc}
\toprule
 & \multicolumn{4}{c}{Qwen3.8-27B} & \multicolumn{4}{c}{Gemma-4-26B-A4B} \\
 & ED$_{\text{party}}$ & corr$_F$ & Rel. & Cross & ED$_{\text{party}}$ & corr$_F$ & Rel. & Cross \\ \midrule
Human floor & -0.009 & 1.10 & 1.00 & 1.06 & -0.011 & 1.10 & 1.00 & 1.07 \\ \midrule
Aggregate sampler (no LLM) & 0.365$^\dagger$ & 2.58$^\dagger$ & 0.00$^\dagger$ & 0.37 & 0.363$^\dagger$ & 2.57$^\dagger$ & 0.00$^\dagger$ & 0.37 \\
Demographic & 0.252$^\dagger$ & 2.19$^\dagger$ & 0.62$^\dagger$ & 1.54 & 0.424$^\dagger$ & 3.42$^\dagger$ & 1.09 & 2.43 \\
\multicolumn{9}{l}{\emph{Selected baselines}} \\
\quad Demographic + reweighting & 0.246 & 2.19 & 0.62$^\dagger$ & 1.54 & 0.452$^\dagger$ & 3.42$^\dagger$ & 1.09 & 2.43 \\
\quad Demographic + temperature & 0.128 & 1.59 & 0.39$^\dagger$ & 1.04 & 0.392$^\dagger$ & 3.28$^\dagger$ & 1.01 & 2.34 \\
\quad Diversity instruction & 0.203$^\dagger$ & 1.90 & 0.54$^\dagger$ & 1.36 & 0.316$^\dagger$ & 2.95$^\dagger$ & 1.00 & 2.18 \\
\quad Diversity, best of 5 wordings & 0.168 & 1.75 & 0.49$^\dagger$ & 1.25 & 0.312$^\dagger$ & 2.92$^\dagger$ & 1.01 & 2.22 \\
\quad Argyle backstory & 0.295$^\dagger$ & 1.83 & 0.48$^\dagger$ & 1.28 & 0.721$^\dagger$ & 3.65$^\dagger$ & 1.07 & 2.64 \\
\quad Life-story backstory & 0.112 & 1.67 & 0.49$^\dagger$ & 1.25 & 0.413$^\dagger$ & 3.40$^\dagger$ & 1.06 & 2.44 \\
\quad Anthology & 0.229 & 2.40$^\dagger$ & 0.52$^\dagger$ & 0.55 & 0.205 & 2.26 & 0.57$^\dagger$ & 0.62 \\
\multicolumn{9}{l}{\emph{Sharpened baselines ($\tau{=}0.05$)}} \\
\quad Demographic & 0.402$^\dagger$ & 3.09$^\dagger$ & 1.01 & 1.93 & 0.441$^\dagger$ & 3.49$^\dagger$ & 1.18 & 2.48 \\
\quad Diversity instruction & 0.383$^\dagger$ & 2.88$^\dagger$ & 1.01 & 1.91 & 0.338$^\dagger$ & 2.98$^\dagger$ & 1.10 & 2.20 \\
\quad Argyle backstory & 0.824$^\dagger$ & 3.77$^\dagger$ & 1.12$^\dagger$ & 2.54 & 0.741$^\dagger$ & 3.73$^\dagger$ & 1.18 & 2.70 \\
\quad Life-story backstory & 0.348$^\dagger$ & 2.98$^\dagger$ & 1.07 & 2.04 & 0.431$^\dagger$ & 3.44$^\dagger$ & 1.15 & 2.44 \\
\midrule
\lpp\ (ours) & 0.079 & 1.85 & 0.96 & 1.17 & 0.071 & 2.05 & 1.25 & 1.44 \\
\bottomrule
\end{tabular}}
\caption{CEK test half, topline calibration. ED$_{\text{party}}$: party-conditional energy distance (bias-corrected, can be $<0$); corr$_F$: correlation-matrix error; Rel.: test-retest reliability / human; Cross: cross-domain $|r|$ / human (ideal 1). Means over 200 half-samples; $^\dagger$: \lpp\ better in $\geq$97.5\% of paired half-samples (Rel.: closer to 1).}
\label{tab:main}
\end{table*}
\begin{table*}[t]\centering\small\setlength{\tabcolsep}{7pt}
\resizebox{\textwidth}{!}{\begin{tabular}{lcccccccccc}
\toprule
 & \multicolumn{5}{c}{Qwen3.8-27B} & \multicolumn{5}{c}{Gemma-4-26B-A4B} \\
 & ED$_{\text{party}}$ & corr$_F$ & Rel. & Cross & Gap & ED$_{\text{party}}$ & corr$_F$ & Rel. & Cross & Gap \\ \midrule
Human floor & -0.006 & 0.72 & 1.00 & 1.08 & 1.02 & -0.006 & 0.72 & 1.00 & 1.08 & 1.02 \\ \midrule
Aggregate sampler (no LLM) & 0.022 & 1.29 & 0.19$^\dagger$ & 0.66 & 1.02 & 0.022 & 1.29$^\dagger$ & 0.19$^\dagger$ & 0.66 & 1.02 \\
Demographic & 0.017 & 1.23 & 0.58$^\dagger$ & 1.33 & 1.02 & 0.019 & 1.39$^\dagger$ & 0.97 & 1.55 & 1.02 \\
\multicolumn{11}{l}{\emph{Selected baselines}} \\
\quad Demographic + reweighting & 0.025 & 1.22 & 0.58$^\dagger$ & 1.30 & 1.02 & 0.030 & 1.40$^\dagger$ & 0.97 & 1.55 & 1.03 \\
\quad Demographic + temperature & 0.018 & 1.12 & 0.41$^\dagger$ & 1.04 & 1.01 & 0.021 & 1.39$^\dagger$ & 0.91 & 1.53 & 1.01 \\
\quad Diversity instruction & 0.017 & 1.19 & 0.54$^\dagger$ & 1.24 & 1.02 & 0.018 & 1.23$^\dagger$ & 0.84 & 1.15 & 1.01 \\
\quad Diversity, best of 5 wordings & 0.015 & 1.15 & 0.51$^\dagger$ & 1.17 & 1.02 & 0.020 & 1.32$^\dagger$ & 0.85 & 1.22 & 1.02 \\
\quad Argyle backstory & 0.017 & 1.12 & 0.42$^\dagger$ & 1.00 & 1.03 & 0.019 & 1.23$^\dagger$ & 0.82 & 1.22 & 1.02 \\
\quad Life-story backstory & 0.017 & 1.19 & 0.52$^\dagger$ & 1.22 & 1.02 & 0.023 & 1.47$^\dagger$ & 1.02 & 1.72 & 1.02 \\
\quad Anthology & 0.020 & 1.24 & 0.53$^\dagger$ & 0.78 & 1.02 & 0.022 & 1.32$^\dagger$ & 0.65$^\dagger$ & 0.93 & 1.02 \\
\multicolumn{11}{l}{\emph{Sharpened baselines ($\tau{=}0.05$)}} \\
\quad Demographic & 0.024 & 1.36$^\dagger$ & 0.92 & 1.46 & 1.02 & 0.019 & 1.40$^\dagger$ & 1.02 & 1.56 & 1.02 \\
\quad Diversity instruction & 0.022 & 1.37$^\dagger$ & 0.88 & 1.46 & 1.01 & 0.017 & 1.23$^\dagger$ & 0.88 & 1.15 & 1.01 \\
\quad Argyle backstory & 0.030 & 1.39$^\dagger$ & 0.87 & 1.33 & 1.02 & 0.022 & 1.24$^\dagger$ & 0.89 & 1.20 & 1.01 \\
\quad Life-story backstory & 0.022 & 1.41$^\dagger$ & 0.87 & 1.52 & 1.02 & 0.023 & 1.48$^\dagger$ & 1.10 & 1.72 & 1.02 \\
\midrule
\lpp\ (ours) & 0.013 & 1.11 & 0.90 & 1.10 & 1.02 & 0.011 & 0.98 & 1.20 & 1.03 & 1.02 \\
\quad \lpp\ without coupling ($k{=}1.6$) & 0.011 & 1.10 & 0.79$^\dagger$ & 0.99 & 1.02 & 0.010 & 0.98 & 1.05 & 0.85 & 1.02 \\
\bottomrule
\end{tabular}}
\caption{GSS test half, party-crosstab calibration; persona configuration and post-processing frozen from CEK. ED$_{\text{party}}$: party-conditional energy distance (bias-corrected, can be $<0$); corr$_F$, Rel., Cross as in Table~\ref{tab:main-otp}; Gap: gap between Democrats and Republicans / human. $^\dagger$ as in Table~\ref{tab:main-otp}. Human reliability reference: GSS 2006--08 panel (two-year interval).}
\label{tab:gss-otp}
\end{table*}
\begin{table*}[t]\centering\small\setlength{\tabcolsep}{7pt}
\resizebox{\textwidth}{!}{\begin{tabular}{lcccccccccc}
\toprule
 & \multicolumn{5}{c}{Qwen3.8-27B} & \multicolumn{5}{c}{Gemma-4-26B-A4B} \\
 & ED$_{\text{party}}$ & corr$_F$ & Rel. & Cross & Gap & ED$_{\text{party}}$ & corr$_F$ & Rel. & Cross & Gap \\ \midrule
Human floor & -0.003 & 0.26 & 1.00 & 0.98 & 1.09 & -0.003 & 0.26 & 1.00 & 0.98 & 1.09 \\ \midrule
Aggregate sampler (no LLM) & 0.028 & 0.78$^\dagger$ & 0.21$^\dagger$ & 0.45 & 1.10 & 0.028 & 0.78$^\dagger$ & 0.21$^\dagger$ & 0.45 & 1.10 \\
Demographic & 0.019 & 0.62 & 0.56$^\dagger$ & 1.13 & 1.10 & 0.041$^\dagger$ & 0.91$^\dagger$ & 1.02 & 1.45 & 1.09 \\
\multicolumn{11}{l}{\emph{Selected baselines}} \\
\quad Demographic + reweighting & 0.021 & 0.61 & 0.56$^\dagger$ & 1.13 & 1.09 & 0.047$^\dagger$ & 0.92$^\dagger$ & 1.02 & 1.46 & 1.10 \\
\quad Demographic + temperature & 0.018 & 0.58 & 0.39$^\dagger$ & 0.79 & 1.09 & 0.040$^\dagger$ & 0.91$^\dagger$ & 0.96 & 1.44 & 1.09 \\
\quad Diversity instruction & 0.018 & 0.58 & 0.49$^\dagger$ & 0.99 & 1.10 & 0.033$^\dagger$ & 0.82$^\dagger$ & 1.00 & 1.38 & 1.09 \\
\quad Diversity, best of 5 wordings & 0.019 & 0.61 & 0.48$^\dagger$ & 1.00 & 1.09 & 0.031$^\dagger$ & 0.92$^\dagger$ & 0.97 & 1.62 & 1.09 \\
\quad Argyle backstory & 0.020 & 0.61 & 0.36$^\dagger$ & 0.72 & 1.09 & 0.025 & 0.62 & 0.84 & 1.20 & 1.09 \\
\quad Life-story backstory & 0.016 & 0.57 & 0.45$^\dagger$ & 0.88 & 1.09 & 0.030 & 0.77$^\dagger$ & 1.00 & 1.37 & 1.09 \\
\quad Anthology & 0.023 & 0.69 & 0.61$^\dagger$ & 0.70 & 1.09 & 0.028 & 0.78$^\dagger$ & 0.22$^\dagger$ & 0.46 & 1.09 \\
\multicolumn{11}{l}{\emph{Sharpened baselines ($\tau{=}0.05$)}} \\
\quad Demographic & 0.043$^\dagger$ & 0.97$^\dagger$ & 0.94 & 1.63 & 1.09 & 0.039$^\dagger$ & 0.90$^\dagger$ & 1.08 & 1.44 & 1.09 \\
\quad Diversity instruction & 0.037$^\dagger$ & 0.79$^\dagger$ & 0.83$^\dagger$ & 1.38 & 1.09 & 0.035$^\dagger$ & 0.84$^\dagger$ & 1.08 & 1.40 & 1.08 \\
\quad Argyle backstory & 0.043$^\dagger$ & 0.74 & 0.79$^\dagger$ & 1.25 & 1.09 & 0.025 & 0.60 & 0.92 & 1.18 & 1.09 \\
\quad Life-story backstory & 0.022 & 0.71 & 0.92 & 1.38 & 1.09 & 0.030 & 0.78$^\dagger$ & 1.10 & 1.39 & 1.09 \\
\midrule
\lpp\ (ours) & 0.011 & 0.57 & 1.06 & 0.91 & 1.09 & 0.010 & 0.54 & 1.16 & 0.97 & 1.09 \\
\quad \lpp\ without coupling ($k{=}1.6$) & 0.016 & 0.58 & 0.97 & 0.58 & 1.10 & 0.012 & 0.52 & 1.16 & 0.74 & 1.10 \\
\bottomrule
\end{tabular}}
\caption{HM test half, party-crosstab calibration; persona configuration and post-processing frozen from CEK. ED$_{\text{party}}$: party-conditional energy distance (bias-corrected, can be $<0$); corr$_F$, Rel., Cross as in Table~\ref{tab:main-otp}; Gap: gap between Democrats and Republicans / human. $^\dagger$ as in Table~\ref{tab:main-otp}. Human reliability reference: HM follow-up wave of the control arm.}
\label{tab:hm-otp}
\end{table*}
\begin{table*}[t]\centering\small\setlength{\tabcolsep}{7pt}
\resizebox{\textwidth}{!}{\begin{tabular}{lcccccccccc}
\toprule
 & \multicolumn{5}{c}{Qwen3.8-27B} & \multicolumn{5}{c}{Gemma-4-26B-A4B} \\
 & ED$_{\text{party}}$ & corr$_F$ & Rel. & Cross & Gap & ED$_{\text{party}}$ & corr$_F$ & Rel. & Cross & Gap \\ \midrule
Human floor & -0.006 & 0.72 & 1.00 & 1.08 & 1.02 & -0.006 & 0.72 & 1.00 & 1.08 & 1.02 \\ \midrule
Aggregate sampler (no LLM) & 0.347$^\dagger$ & 1.74$^\dagger$ & 0.00$^\dagger$ & 0.40 & 0.19 & 0.347$^\dagger$ & 1.74$^\dagger$ & 0.00$^\dagger$ & 0.40 & 0.19 \\
Demographic & 0.098 & 1.47$^\dagger$ & 0.62$^\dagger$ & 1.43 & 1.18 & 0.257$^\dagger$ & 2.16$^\dagger$ & 1.19 & 2.04 & 1.48 \\
\multicolumn{11}{l}{\emph{Selected baselines}} \\
\quad Demographic + reweighting & 0.103 & 1.47 & 0.62$^\dagger$ & 1.42 & 1.18 & 0.272$^\dagger$ & 2.18$^\dagger$ & 1.19 & 2.06 & 1.49 \\
\quad Demographic + temperature & 0.070 & 1.26 & 0.42$^\dagger$ & 1.01 & 0.99 & 0.248$^\dagger$ & 2.12$^\dagger$ & 1.12 & 2.01 & 1.46 \\
\quad Diversity instruction & 0.097 & 1.44 & 0.59$^\dagger$ & 1.35 & 1.16 & 0.210$^\dagger$ & 1.70$^\dagger$ & 0.97 & 1.38 & 1.25 \\
\quad Diversity, best of 5 wordings & 0.085 & 1.34 & 0.52$^\dagger$ & 1.16 & 1.03 & 0.209$^\dagger$ & 1.75$^\dagger$ & 0.93 & 1.30 & 1.16 \\
\quad Argyle backstory & 0.152 & 1.47 & 0.55$^\dagger$ & 1.37 & 1.37 & 0.476$^\dagger$ & 2.47$^\dagger$ & 1.22 & 2.37 & 1.91 \\
\quad Life-story backstory & 0.108 & 1.44 & 0.55$^\dagger$ & 1.33 & 1.12 & 0.234$^\dagger$ & 2.16$^\dagger$ & 1.19 & 2.20 & 1.52 \\
\quad Anthology & 0.157 & 1.57$^\dagger$ & 0.43$^\dagger$ & 0.52 & 0.47 & 0.192$^\dagger$ & 1.63$^\dagger$ & 0.55$^\dagger$ & 0.68 & 0.48 \\
\multicolumn{11}{l}{\emph{Sharpened baselines ($\tau{=}0.05$)}} \\
\quad Demographic & 0.140 & 1.68$^\dagger$ & 1.00 & 1.62 & 1.24 & 0.259$^\dagger$ & 2.17$^\dagger$ & 1.24 & 2.05 & 1.48 \\
\quad Diversity instruction & 0.166 & 1.77$^\dagger$ & 0.98 & 1.68 & 1.30 & 0.205$^\dagger$ & 1.68$^\dagger$ & 1.01 & 1.38 & 1.25 \\
\quad Argyle backstory & 0.468$^\dagger$ & 2.51$^\dagger$ & 1.21 & 2.51 & 1.85 & 0.494$^\dagger$ & 2.50$^\dagger$ & 1.30$^\dagger$ & 2.38 & 1.91 \\
\quad Life-story backstory & 0.229$^\dagger$ & 1.99$^\dagger$ & 0.99 & 1.86 & 1.34 & 0.239$^\dagger$ & 2.17$^\dagger$ & 1.27 & 2.23 & 1.53 \\
\midrule
\lpp\ (ours) & 0.093 & 1.26 & 0.84 & 0.88 & 0.65 & 0.069 & 1.08 & 1.24 & 0.83 & 0.76 \\
\quad \lpp\ without coupling ($k{=}1.6$) & 0.098 & 1.28 & 0.73$^\dagger$ & 0.78 & 0.66 & 0.080 & 1.15 & 1.03 & 0.67 & 0.70 \\
\bottomrule
\end{tabular}}
\caption{GSS test half, topline calibration; persona configuration and post-processing frozen from CEK. ED$_{\text{party}}$: party-conditional energy distance (bias-corrected, can be $<0$); corr$_F$, Rel., Cross as in Table~\ref{tab:main-otp}; Gap: gap between Democrats and Republicans / human. $^\dagger$ as in Table~\ref{tab:main-otp}. Human reliability reference: GSS 2006--08 panel (two-year interval).}
\label{tab:gss}
\end{table*}
\begin{table*}[t]\centering\small\setlength{\tabcolsep}{7pt}
\resizebox{\textwidth}{!}{\begin{tabular}{lcccccccccc}
\toprule
 & \multicolumn{5}{c}{Qwen3.8-27B} & \multicolumn{5}{c}{Gemma-4-26B-A4B} \\
 & ED$_{\text{party}}$ & corr$_F$ & Rel. & Cross & Gap & ED$_{\text{party}}$ & corr$_F$ & Rel. & Cross & Gap \\ \midrule
Human floor & -0.003 & 0.26 & 1.00 & 0.98 & 1.09 & -0.003 & 0.26 & 1.00 & 0.98 & 1.09 \\ \midrule
Aggregate sampler (no LLM) & 0.274$^\dagger$ & 1.11$^\dagger$ & 0.00$^\dagger$ & 0.16 & 0.14 & 0.274$^\dagger$ & 1.11$^\dagger$ & 0.00$^\dagger$ & 0.16 & 0.14 \\
Demographic & 0.080 & 0.77 & 0.67$^\dagger$ & 1.41 & 1.52 & 0.201$^\dagger$ & 1.19$^\dagger$ & 1.10 & 1.74 & 1.78 \\
\multicolumn{11}{l}{\emph{Selected baselines}} \\
\quad Demographic + reweighting & 0.082 & 0.76 & 0.68$^\dagger$ & 1.41 & 1.52 & 0.208$^\dagger$ & 1.19$^\dagger$ & 1.10 & 1.74 & 1.79 \\
\quad Demographic + temperature & 0.037 & 0.56 & 0.44$^\dagger$ & 0.86 & 1.22 & 0.194$^\dagger$ & 1.18$^\dagger$ & 1.05 & 1.73 & 1.77 \\
\quad Diversity instruction & 0.066 & 0.63 & 0.60$^\dagger$ & 1.18 & 1.41 & 0.157$^\dagger$ & 1.13$^\dagger$ & 1.11 & 1.71 & 1.72 \\
\quad Diversity, best of 5 wordings & 0.069 & 0.73 & 0.59$^\dagger$ & 1.30 & 1.46 & 0.252$^\dagger$ & 1.37$^\dagger$ & 1.10 & 2.19 & 1.95 \\
\quad Argyle backstory & 0.115 & 0.61 & 0.57$^\dagger$ & 1.15 & 1.64 & 0.336$^\dagger$ & 1.34$^\dagger$ & 1.06 & 2.14 & 2.11 \\
\quad Life-story backstory & 0.037 & 0.55 & 0.50$^\dagger$ & 0.97 & 1.26 & 0.171$^\dagger$ & 1.07$^\dagger$ & 1.10 & 1.65 & 1.71 \\
\quad Anthology & 0.104 & 0.85$^\dagger$ & 0.56$^\dagger$ & 0.46 & 0.53 & 0.272$^\dagger$ & 1.10$^\dagger$ & 0.02$^\dagger$ & 0.16 & 0.13 \\
\multicolumn{11}{l}{\emph{Sharpened baselines ($\tau{=}0.05$)}} \\
\quad Demographic & 0.205$^\dagger$ & 1.32$^\dagger$ & 1.06$^\dagger$ & 2.10 & 1.83 & 0.200$^\dagger$ & 1.18$^\dagger$ & 1.14 & 1.73 & 1.79 \\
\quad Diversity instruction & 0.183$^\dagger$ & 1.14$^\dagger$ & 0.97 & 1.88 & 1.77 & 0.157$^\dagger$ & 1.15$^\dagger$ & 1.19$^\dagger$ & 1.72 & 1.71 \\
\quad Argyle backstory & 0.323$^\dagger$ & 1.33$^\dagger$ & 1.06 & 2.14 & 2.13 & 0.326$^\dagger$ & 1.30$^\dagger$ & 1.11 & 2.09 & 2.09 \\
\quad Life-story backstory & 0.129 & 1.05$^\dagger$ & 1.04 & 1.86 & 1.71 & 0.177$^\dagger$ & 1.09$^\dagger$ & 1.19$^\dagger$ & 1.68 & 1.73 \\
\midrule
\lpp\ (ours) & 0.058 & 0.65 & 1.02 & 0.71 & 0.66 & 0.027 & 0.52 & 1.16 & 0.84 & 0.83 \\
\quad \lpp\ without coupling ($k{=}1.6$) & 0.116$^\dagger$ & 0.86$^\dagger$ & 0.91$^\dagger$ & 0.28 & 0.40 & 0.052 & 0.61 & 1.17 & 0.53 & 0.65 \\
\bottomrule
\end{tabular}}
\caption{HM test half, topline calibration; persona configuration and post-processing frozen from CEK. ED$_{\text{party}}$: party-conditional energy distance (bias-corrected, can be $<0$); corr$_F$, Rel., Cross as in Table~\ref{tab:main-otp}; Gap: gap between Democrats and Republicans / human. $^\dagger$ as in Table~\ref{tab:main-otp}. Human reliability reference: HM follow-up wave of the control arm.}
\label{tab:hm}
\end{table*}
\begin{table*}[t]\centering\small\setlength{\tabcolsep}{4pt}
\resizebox{\textwidth}{!}{\begin{tabular}{ll cccccc cccccc}
\toprule
 & & \multicolumn{6}{c}{Reliability 0.8 of human} & \multicolumn{6}{c}{Reliability 0.9 of human} \\
\cmidrule(lr){3-8}\cmidrule(lr){9-14}
Survey & Model & \lpp & Uncoupled & Best baseline & Reach & $\Delta$ [95\% CI] & P(lowest) & \lpp & Uncoupled & Best baseline & Reach & $\Delta$ [95\% CI] & P(lowest) \\ \midrule
CEK & Qwen & \textbf{1.59} & 1.58 & 2.05 (Best diversity) & 5/6 & $-$0.43 [$-$0.71, $-$0.13] & 99\% & \textbf{1.65} & 1.61 & 2.15 (Best diversity) & 5/6 & $-$0.48 [$-$0.76, $-$0.13] & 100\% \\
CEK & Gemma & \textbf{1.70} & 1.65 & 2.04 (Best diversity) & 5/6 & $-$0.31 [$-$0.59, +0.03] & 96\% & \textbf{1.70} & 1.65 & 2.16 (Best diversity) & 5/6 & $-$0.41 [$-$0.72, $-$0.07] & 100\% \\
GSS & Qwen & \textbf{1.09} & 1.10 & 1.28 (Best diversity) & 5/6 & $-$0.18 [$-$0.33, $-$0.00] & 98\% & \textbf{1.11} & not reached & 1.36 (Demographic) & 1/6 & $-$0.25 [$-$0.45, $-$0.05] & 100\% \\
GSS & Gemma & \textbf{0.95} & 0.98 & 1.22 (Diversity) & 5/6 & $-$0.22 [$-$0.40, $-$0.04] & 99\% & \textbf{0.95} & 0.98 & 1.39 (Demographic) & 2/6 & $-$0.42 [$-$0.62, $-$0.20] & 100\% \\
HM & Qwen & \textbf{0.57} & 0.60 & 0.69 (Life story) & 5/6 & $-$0.11 [$-$0.26, +0.05] & 94\% & \textbf{0.56} & 0.58 & 0.70 (Life story) & 2/6 & $-$0.15 [$-$0.30, +0.01] & 95\% \\
HM & Gemma & \textbf{0.54} & 0.52 & 0.65 (Argyle) & 5/6 & $-$0.09 [$-$0.27, +0.09] & 82\% & \textbf{0.54} & 0.52 & 0.60 (Argyle) & 5/6 & $-$0.06 [$-$0.26, +0.12] & 76\% \\
\bottomrule
\end{tabular}}
\caption{Correlation error at matched reliability, interpolated on the frontier of each persona type (Figure~\ref{fig:frontier}; party-crosstab calibration, post-processing down to $\tau=0.01$). Best baseline: lowest among the six baseline persona types that reach the target reliability; Reach: how many do (none: no baseline does). $\Delta$: \lpp\ minus the lowest reaching baseline per half-sample, mean and 2.5/97.5 percentiles. P(lowest): share of the 200 paired half-samples in which \lpp\ is below every baseline that reaches the target. Uncoupled: \lpp\ with one deviation scale for all engagement levels (not selected; first latent draw); not reached: its frontier ends below the target. A frontier lying entirely above the target is evaluated at its least reliable point. GSS reliability is relative to a two-year panel.}
\label{tab:matched}
\end{table*}
\begin{table*}[t]\centering\small
\begin{tabular}{@{}>{\raggedright\arraybackslash}p{3.6cm}>{\raggedright\arraybackslash}p{2.7cm}>{\raggedright\arraybackslash}p{8.85cm}@{}}
\toprule
Check & Quantity & Result \\ \midrule
GSS calibrated with 2022 tabulations & corr$_F$, \lpp\ vs.\ baselines & Qwen: selected 1.19 vs.\ 1.15--1.31; at Rel.\ 0.9 1.17 vs.\ 1.42; Gemma: selected 1.06 vs.\ 1.34--1.72; at Rel.\ 0.9 1.05 vs.\ 1.35 \\
Redrawn latents, 6 seeds (CEK) & corr$_F$ mean $\pm$ SD & Qwen: 1.67 $\pm$ 0.06, range 1.60--1.76 (best baseline at Rel.\ 0.9: 2.15); Gemma: 1.84 $\pm$ 0.10, range 1.70--2.00 (best baseline at Rel.\ 0.9: 2.16) \\
Polychoric instead of Pearson (CEK) & rank correlation of methods & Qwen: $\rho$ = 0.94; Gemma: $\rho$ = 0.94 \\
Human retest by partisan strength (CEK) & independent / leaner / weak / strong & 0.75 / 0.68 / 0.66 / 0.65 \\
Engagement-dependent sharpening (dev.) & selection criterion (lower is better) & best graded 0.90 vs.\ uniform 0.87 \\
One continuous ideology latent (CEK) & corr$_F$ and Rel. & Qwen: selected 1.60 at Rel.\ 0.48, 2.30 at Rel.\ 0.9 (\lpp: 1.68 at 0.95, 1.65 at 0.9); Gemma: selected 2.45 at Rel.\ 1.00, 2.38 at Rel.\ 0.9 (\lpp: 1.84 at 1.24, 1.70 at 0.9) \\
Latents as direct offsets, no prompting (CEK) & corr$_F$ and Rel. & Qwen: selected 1.71 at Rel.\ 0.87; at Rel.\ 0.9 1.74 (\lpp\ 1.65, demographic 2.51); Gemma: selected 2.61 at Rel.\ 0.98; at Rel.\ 0.9 2.46 (\lpp\ 1.70, demographic 2.51) \\
Uncoupled persistent offsets, no prompting (CEK) & corr$_F$ and Rel. & Qwen: selected 1.76 at Rel.\ 0.90; at Rel.\ 0.9 1.75 (\lpp\ 1.65, demographic 2.51); Gemma: selected 2.58 at Rel.\ 0.98; at Rel.\ 0.9 2.45 (\lpp\ 1.70, demographic 2.51) \\
Retest with reworded questions (CEK) & Rel., \lpp\ vs.\ baselines & Qwen: selected 0.83 vs.\ 0.32--0.44; highest reachable 0.90 vs.\ 0.37--0.86; Gemma: selected 1.01 vs.\ 0.37--0.89; highest reachable 1.01 vs.\ 0.37--0.91 \\
No calibration (CEK) & corr$_F$ at Rel.\ 0.9 & Qwen: \lpp\ 1.87 vs.\ Anthology 2.32, other baselines 2.88--3.39; Gemma: \lpp\ 2.13 vs.\ Anthology 2.12, other baselines 3.13--4.22 \\
\bottomrule
\end{tabular}
\caption{Robustness checks (party-crosstab calibration unless noted). Correlation errors are not comparable across rows that use different samples.}
\label{tab:robust}
\end{table*}

\paragraph{Example \lpp\ persona} (a CEK test respondent with low engagement, so deviations are frequent):

\begin{quote}\small
You are a 30-39 year-old White man living in the United States. Your highest level of education is college. Generally speaking, you think of yourself as a strong Democrat, and when it comes to politics you think of yourself as liberal. You rarely follow politics. You form opinions issue by issue, based on your own experiences and gut feelings, and they often do not line up with a single political ideology. In your life, respecting traditions and customs is somewhat important to you; thinking for yourself and making your own choices is very important to you; safety and stability for yourself and your family is not very important to you; helping the people close to you is somewhat important to you; personal success and ambition is not very important to you; fairness and equal opportunity for everyone is very important to you.\\
Your views on specific issues:\\
- On climate change and environmental policy, your views are somewhat more conservative than most people who share your general politics.\\
- On regulating banks and Wall Street, your views are noticeably more liberal than most people who share your general politics.\\
- On government spending on transportation and infrastructure, your views are noticeably more liberal than most people who share your general politics.\\
- On how the government provides veterans' care, your views are somewhat more conservative than most people who share your general politics.\\
- On taxes and the federal tax system, your views are noticeably more conservative than most people who share your general politics.
\end{quote}
The persona is followed by ``Now answer the survey question as this person would.'', the question, its labeled options, and ``Answer with a single number from 1 to $K$.''
The Demographic baseline is the first three sentences.

\paragraph{Engagement sentences.}
\emph{High:} ``You follow politics very closely, and your opinions on different issues tend to fit together into a consistent political outlook.''
\emph{Mid:} ``You follow politics somewhat. Your opinions mostly line up with your general political leanings, but not on every issue.''
\emph{Low:} as in the example.

\paragraph{Domain phrases.}
GSS: \emph{government help for the poor and reducing income inequality}; \emph{crime, guns and the death penalty}; \emph{abortion and sexual morality}; \emph{free speech and censorship}; \emph{trust in national institutions such as the government, the press and the military}.
HM: \emph{crime and policing}; \emph{national security and terrorism}; \emph{government help for the poor}; \emph{health care policy}; \emph{government spending to boost the economy}.
No other part of the configuration was changed for these surveys.

\paragraph{Baseline prompts.}
The diversity instruction appends ``Keep in mind that real people often hold mixed views that do not follow a single party line; answer as this specific individual, not as a typical member of a group.'' The four alternative wordings are ``Remember that this person's opinions are their own and may differ from what most people in their party think.''; ``People who share a party often disagree on specific issues. Answer based on this person's individual circumstances and values, not on party stereotypes.''; ``Avoid stereotypes: this person holds their own combination of views, which need not be consistent across issues.''; and ``Think about this person's own life and experiences when answering; real people's opinions are often inconsistent across issues and only loosely tied to party.'' (selected).
The life-story backstory is a short first-person life story (requested at about 120 words, without political issues) that the same model writes for the demographic profile (sampled at temperature 1, seeded per respondent). It is prepended to the profile.
Anthology follows \citet{moon2024virtual}. A pool of 400 backstories is generated at temperature 1 in response to ``Tell me about yourself'', and the model answers a demographic questionnaire on the six profile attributes as each backstory. Backstories are then assigned one-to-one to respondents by maximum-weight bipartite matching on the summed log-probabilities of the respondents' attributes, and the survey is answered conditioned on the matched backstory alone. The original method elicits backstories from base models, but no base checkpoint of the models we study is available. The instruction-tuned models answer the bare question with a description of themselves as an AI, even as a raw-text continuation without a chat template. We therefore ask the model to answer the question in the voice of ``a randomly chosen adult living in the United States'' (about 150 words). A filter for AI self-descriptions rejected none of the 400 backstories for either model.

On GSS and HM, the three backstory baselines are built in the same way from each survey's covariates (life stories sampled in batches of 12). Because the test halves (428 and 642 respondents) exceed the 400-story Anthology pool, each story may be matched to at most two respondents in these surveys.

\section{Implementation Details}
\label{app:impl}
\begin{table}[t]\centering\small\setlength{\tabcolsep}{3pt}
\resizebox{\columnwidth}{!}{\begin{tabular}{lcccccccc}
\toprule
 & \multicolumn{4}{c}{Qwen3.8-27B} & \multicolumn{4}{c}{Gemma-4-26B-A4B} \\
 & corr$_F$ & Rel. & Cross & Gap$^{\text{top}}$ & corr$_F$ & Rel. & Cross & Gap$^{\text{top}}$ \\ \midrule
\lpp\ (full) & 1.68 & 0.95 & 1.10 & 1.12 & 1.84 & 1.24 & 1.30 & 1.15 \\
\quad $-$ coupling ($k{=}1.6$) & 1.60 & 0.84 & 0.94 & 1.22 & 1.79 & 1.21 & 1.01 & 1.27 \\
\quad $-$ value priorities & 1.78 & 1.02 & 1.15 & 1.30 & 1.74 & 1.23 & 1.09 & 1.21 \\
\quad $-$ issue deviations & 1.85 & 0.83 & 1.44 & 1.42 & 2.36 & 1.09 & 1.82 & 1.83 \\
\quad $-$ sharpening & 1.50 & 0.54 & 0.83 & 0.93 & 1.83 & 1.18 & 1.29 & 1.15 \\
\quad party-relative $\to$ absolute dev. & 1.65 & 1.06 & 0.84 & 0.77 & 1.85 & 1.30 & 0.99 & 1.02 \\
\quad latents $\to$ placebo attributes & 2.10 & 0.80 & 1.52 & 1.67 & 2.51 & 1.05 & 1.93 & 1.97 \\
\quad second latent seed & 1.67 & 0.95 & 1.13 & 1.15 & 1.77 & 1.24 & 1.26 & 1.19 \\
\quad engagement prior .15/.35/.50 & 1.60 & 0.94 & 0.90 & 1.15 & 1.76 & 1.24 & 0.98 & 1.22 \\
\quad engagement prior .40/.40/.20 & 1.63 & 0.93 & 1.17 & 1.29 & 1.77 & 1.22 & 1.26 & 1.42 \\
Demographic + sharpening ($\tau{=}0.15$) & 2.46 & 0.86 & 1.63 & 1.75 & 2.71 & 1.06 & 2.08 & 1.95 \\
Demographic & 1.65 & 0.52 & 1.21 & 1.60 & 2.64 & 0.98 & 2.03 & 1.94 \\
\bottomrule
\end{tabular}}
\caption{Ablations on the CEK test half (party-crosstab calibration; every row is calibrated, and sharpened with $\tau{=}0.15$ as in \lpp\ unless it removes sharpening; the sharpened baselines of Table~\ref{tab:main-otp} use $\tau{=}0.05$). Rel., Cross, and Gap$^{\text{top}}$ (gap between Democrats and Republicans under topline calibration) are ratios to the human value (ideal 1; human floor for Cross 1.06 Qwen, 1.07 Gemma).}
\label{tab:abl}
\end{table}
\begin{table}[t]\centering\small\setlength{\tabcolsep}{3pt}
\resizebox{\columnwidth}{!}{\begin{tabular}{lccccccc}
\toprule
 & Qwen & Qwen & Qwen & Gemma & Gemma & Gemma & GPT \\
 & CEK & GSS & HM & CEK & GSS & HM & CEK \\ \midrule
Cross-domain $|r|$ & 1.53 & 2.24 & 1.34 & 2.69 & 2.81 & 1.90 & 4.10 \\
Party gap & 1.51 & 1.31 & 1.29 & 1.33 & 1.31 & 1.34 & 1.65 \\
Within-party SD & 0.84 & 0.81 & 0.78 & 0.47 & 0.53 & 0.56 & 0.33 \\
Party $R^2$ & 2.80 & 2.37 & 2.45 & 3.94 & 3.17 & 3.32 & 6.74 \\
\bottomrule
\end{tabular}}
\caption{Diversity-instruction personas, no post-processing: statistics of the simulated public relative to the human test half (1 = human). Human floor (two disjoint human samples): Cross-domain $|r|$ 0.98--1.08; Party gap 0.98--1.09; Within-party SD 1.00--1.04; Party $R^2$ 1.01--1.20. Gemma/GSS: one item (allowing an atheist to teach) receives the same answer from every diversity-instruction persona; correlations are computed over the remaining items.}
\label{tab:diag-div}
\end{table}

\paragraph{Readout.} Qwen runs in 4-bit NF4 quantization. Gemma uses its 8-bit FP8 checkpoint, with the experts' weights dequantized on the fly and all activations in bf16. We use the checkpoints \texttt{Qwen/Qwen3.8-27B} (revision \texttt{1d4bf0f}), quantized to NF4 at load time, for Qwen and \texttt{RedHatAI/gemma-4-26B-A4B-it-FP8-dynamic} (revision \texttt{ed35d7a}), an FP8 quantization of \texttt{google/gemma-4-26B-A4B-it}, for Gemma. Both use chat templates (Qwen with thinking disabled). We read the next-token probabilities of the option digits after the assistant prefix and renormalize them. The digits carry on average 99.7\% of the probability mass. GPT-6.1-Sol is called through the Codex CLI (\texttt{codex exec}) with reasoning effort set to low, in a fresh read-only session from an empty temporary directory and without user configuration. It returns a JSON object with all answers of a persona, and the second call for reliability uses the identical prompt.

\paragraph{Calibration map.} For item $j$, the monotone transport plan between the pooled simulated marginal and the reference marginal is computed by the north-west-corner rule on the ordered options and row-normalized to a stochastic matrix $M_j$. Party-conditional calibration does this separately for Democrats (party ID 1--3 on the 7-point scale), independents (4), and Republicans (5--7).

\paragraph{Hyperparameters.} Hyperparameters are selected on the CEK development half (100 simulated respondents, compared with all 311 development respondents). The 32 persona configurations include engagement-coupled thresholds $\{k_{\text{mid}},k_{\text{low}}\}\in\{(1.6,0.6),(1.0,0.4),(2.5,1.0)\}$ with $k_{\text{high}}\in\{\infty,2.5\}$ and uniform thresholds $k\in\{0.1,0.25,0.4,0.6,1,1.6,2.5\}$. For every method, the post-processing is sharpening with $\tau\in\{0.05,0.1,0.15,0.25,0.35,0.5,0.7\}$, none, or temperature $T\in\{1.5,2,3,4\}$. The criterion is the sum of correlation error and reliability error, each normalized by the calibrated Demographic baseline (correlation error relative to the human floor) and averaged over the two calibration regimes. For topline calibration, the normalized ED$_{\text{party}}$ is added. The same criterion selects each baseline's post-processing, and ablation variants use the selected $\tau{=}0.15$ without re-selection.

\paragraph{Evaluation.} Each of 200 resamples draws half of the test respondents without replacement (paired across methods), draws one answer per simulated person and item, and computes all metrics against the human answers of the same half. The human floor compares an equally sized sample of the development half with the same test half. Energy distances standardize each item by the human test-half SD and use the unbiased (U-statistic) estimator, which can be slightly negative.

\paragraph{Compute.} Qwen3.8-27B runs used one 24\,GB GPU and Gemma-4-26B-A4B runs used two. One full pass of 311 personas over 14 items takes about 25 minutes for Qwen3.8-27B and 6 minutes for Gemma-4-26B-A4B. GPT-6.1-Sol runs used about 6{,}000 Codex CLI calls in total.

\section{Additional Results}
\label{app:diag}
\begin{table}[t]\centering\small\setlength{\tabcolsep}{2.6pt}
\resizebox{\columnwidth}{!}{\begin{tabular}{lcccc@{\hspace{6pt}}ccc}
\toprule
 & \multicolumn{4}{c}{Toplines} & \multicolumn{3}{c}{+ party crosstabs} \\
\cmidrule(lr){2-5}\cmidrule(lr){6-8}
 & ED$_{\text{p}}$ & corr$_F$ & wpSD & Rel. & corr$_F$ & Cross & Within \\ \midrule
Human floor & -0.01 & 1.11 & 1.04 & 1.00 & 1.11 & 1.06 & 1.12 \\ \midrule
Demographic & 1.15 & 4.97 & 0.72 & 1.13 & 2.78 & 2.23 & 1.36 \\
Diversity instruction & 0.92 & 3.91 & 0.76 & 1.00 & 1.91 & 1.50 & 1.10 \\
\lpp & 0.23 & 2.47 & 0.98 & 1.25 & 2.22 & 0.74 & 2.05 \\
\bottomrule
\end{tabular}}
\caption{GPT-6.1-Sol, CEK test half, after calibration with toplines only (left) or toplines and party crosstabs (right). wpSD, Cross, Within: within-party SD, cross- and within-domain $|r|$ relative to humans. Rel.: empirical reliability from two independent calls per persona after topline calibration, relative to humans (1.38 for demographic personas before calibration).}
\label{tab:gpt}
\end{table}
\begin{table}[t]\centering\small\setlength{\tabcolsep}{2.5pt}
\resizebox{\columnwidth}{!}{\begin{tabular}{ll cccc@{\hspace{6pt}}cccc}
\toprule
 & & \multicolumn{4}{c}{Qwen3.8-27B} & \multicolumn{4}{c}{Gemma-4-26B-A4B} \\
\cmidrule(lr){3-6}\cmidrule(lr){7-10}
Persona & Post-proc. & On-topic & $\times$hum. & $r_{\text{cells}}$ & RMSE & On-topic & $\times$hum. & $r_{\text{cells}}$ & RMSE \\ \midrule
Demographic & none & 0.64 & 6.3 & 0.85 & 0.44 & 0.45 & 4.4 & 0.59 & 0.24 \\
 & cal. & 0.69 & 6.8 & 0.82 & 0.48 & 0.48 & 4.7 & 0.54 & 0.27 \\
 & temp.\ 2 + cal. & 0.55 & 5.4 & 0.84 & 0.34 & 0.49 & 4.8 & 0.54 & 0.27 \\
 & sharp.\ 0.15 + cal. & 0.73 & 7.2 & 0.79 & 0.70 & 0.53 & 5.2 & 0.49 & 0.31 \\
\midrule
\lpp & none & 0.39 & 3.8 & 0.84 & \textbf{0.21} & 0.31 & 3.1 & 0.70 & \textbf{0.15} \\
 & cal. & 0.47 & 4.6 & 0.83 & \textbf{0.26} & 0.38 & 3.7 & 0.66 & \textbf{0.19} \\
 & temp.\ 2 + cal. & 0.39 & 3.8 & 0.85 & \textbf{0.20} & 0.38 & 3.7 & 0.66 & \textbf{0.18} \\
 & sharp.\ 0.15 + cal. & 0.48 & 4.7 & 0.81 & \textbf{0.28} & 0.38 & 3.7 & 0.67 & \textbf{0.19} \\
\bottomrule
\end{tabular}}
\caption{HM framing experiment (Section~\ref{sec:dynamic}): mean on-topic effect (framed minus control, in human control-arm SD; human 0.10), its ratio to the human effect, cell-wise correlation with the human $4\times5$ effect matrix and its RMSE. Calibration is fit on the simulated control arm only. Bold: lower RMSE of the two persona types.}
\label{tab:hmdyn}
\end{table}

Table~\ref{tab:diag-div} repeats the diagnosis of Section~\ref{sec:diagnosis} for the diversity instruction. Relative to the demographic persona, it lowers cross-domain constraint by at most about 15\% (on Gemma/GSS one constant item is excluded), which leaves every statistic far from the human level. Within-party heterogeneity is even lower than with the demographic persona on every model and survey.

\paragraph{Unstated domains.} Here we state deviations only for climate, financial regulation, and infrastructure, using uniform thresholds ($k{=}1.6$) with the same sharpening and topline calibration as \lpp. On Qwen (CEK test half), the mean $|r|$ between the two unstated domains (veterans' care, taxes) and all other domains is then 1.53$\times$ the human value. This is no better than the demographic persona (1.55$\times$), whereas stating all five domains gives 1.21$\times$. The party gap on the unstated items is not reduced either (1.76$\times$ vs.\ 1.63$\times$).

Per-method results for CEK with topline calibration are in Table~\ref{tab:main}. GSS and HM results are in Tables~\ref{tab:gss-otp} and~\ref{tab:hm-otp} with party-crosstab calibration and in Tables~\ref{tab:gss} and~\ref{tab:hm} with topline calibration. Table~\ref{tab:transfer} summarizes topline calibration across settings, and Table~\ref{tab:gpt} reports GPT-6.1-Sol.

\paragraph{GPT-6.1-Sol.} GPT-6.1-Sol samples all answers of a persona in one call, so sharpening does not apply, and its deviation scale was chosen on the development half. \lpp\ is best under topline calibration (\N{cg-gpt-pool-lpp-corr} vs.\ \N{cg-gpt-pool-sel-div-corr}, difference \N{cg-gpt-pool-sel-div-ci}) but not with party crosstabs (\N{cg-gpt-otp-lpp-corr} vs.\ \N{cg-gpt-otp-sel-div-corr}). The reason is that GPT applies a stated deviation to every item of its domain (within-domain $|r|$ \N{cg-gpt-otp-lpp-within}$\times$ human, Table~\ref{tab:gpt}). Its sampled answers are at least as reliable as human ones for every persona type, so the reliability deficit concerns the open models' probability readouts.

\paragraph{Topline calibration.} Without party crosstabs (Table~\ref{tab:transfer}), \lpp\ has the lowest correlation error in \N{nlowest-pool} of six settings (all three with Gemma, and GSS with Qwen, where it ties). On HM with Qwen, its personas under-separate the parties (party gap \N{hm-qwen-pool-lpp-gap} of human), which party crosstabs correct.

\section{Robustness}
\label{app:robust}

Table~\ref{tab:robust} summarizes our additional robustness checks. Main results average six latent draws of \lpp, while the reworded retest, polychoric check, framing experiment, and opinion-switching analysis use the first draw.

\paragraph{Calibration transfer, seeds, and ordinal correlations.}
Calibrating GSS to published-style tabulations from an earlier wave (2022) instead of the 2024 development half tests calibration with an out-of-sample reference. Redrawing all latents across five additional seeds bounds the variance introduced by the fixed prior; every seed remains below the best baseline at human-level reliability. Replacing Pearson with polychoric correlations accounts for the ordinal response scales and yields nearly identical method rankings.

\paragraph{Reliability and calibration ceiling.}
For the reworded retest, we ask the 14 CEK items in a second, meaning-preserving wording using the same response scale (released with the code), keeping the persona text and latent draw fixed, calibrating each wave separately, and computing reliability as the between-wave response correlation; this reduces to our main reliability measure when the wordings coincide. Without calibration, every persona type can be sharpened to human-level reliability, showing that the baseline ceilings in Table~\ref{tab:matched} arise because calibration reintroduces spread as noise. Uncalibrated, baseline correlation error at human-level reliability remains far above \lpp's on Qwen, whereas Anthology matches uncalibrated \lpp\ on Gemma. For the partisan-strength breakdown, we use partisanship as a proxy for political engagement because none of the three surveys measures political interest directly.

\paragraph{Alternative latent controls.}
The direct-offset control tilts each Demographic response distribution,
$p'(k)\propto p(k)\exp(\beta\,\ell_d\,s_j\,(k-4)/3)$,
by the persona's domain deviation level $\ell_d\in\{-2,\dots,2\}$ (the same discrete bin that \lpp\ states in text for the same latent draw). Here $s_j$ is the sign of the difference in expected answers between Republican and Democratic Demographic personas, and both $\beta{=}3$ and the post-processing are selected on the development half. Its uncoupled variant assigns the same deviation scale to all personas, corresponding to persistent random persona$\times$domain shifts without engagement structure. The one-dimensional baseline instead assigns each persona a continuous ideology score drawn around its party's mean, with post-processing selected on the development half like every other method.

\paragraph{Deviation framing and engagement priors.}
Under topline calibration, stating deviations in absolute rather than party-relative terms under-separates the parties by roughly as much as \lpp\ over-separates them (party gap \N{cg-qwen-pool-abl-abs-gap} vs.\ \N{cg-qwen-pool-lpp-gap} of the human value), so the framing of deviations makes little difference. Finally, varying the engagement prior shares (Table~\ref{tab:abl}) shifts correlation error by less than 0.1 on either model and slightly lowers it in both cases (Qwen \N{cg-qwen-otp-abl-englow-corr} and \N{cg-qwen-otp-abl-enghigh-corr} vs.\ \N{cg-qwen-otp-lpp-corr}; Gemma \N{cg-gemma-otp-abl-englow-corr} and \N{cg-gemma-otp-abl-enghigh-corr} vs.\ \N{cg-gemma-otp-lpp-corr}), showing that our findings do not hinge on the chosen shares.

\end{document}

%% file: macros.tex
\newcommand{\DIAGwithin}{1.2--2.4$\times$}
\newcommand{\DIAGcross}{1.6--4.6$\times$}
\newcommand{\DIAGpid}{2.5--7.6$\times$}

\makeatletter
\newcommand{\N}[1]{\ifcsname N@#1\endcsname\csname N@#1\endcsname\else\textbf{\textcolor{red}{??#1}}\fi}
\expandafter\def\csname N@cg-qwen-otp-floor-corr\endcsname{1.10}
\expandafter\def\csname N@cg-qwen-otp-floor-cross\endcsname{1.06}
\expandafter\def\csname N@cg-qwen-otp-lpp-corr\endcsname{1.68}
\expandafter\def\csname N@cg-qwen-otp-lpp-rel\endcsname{0.95}
\expandafter\def\csname N@cg-qwen-otp-lpp-cross\endcsname{1.10}
\expandafter\def\csname N@cg-qwen-otp-lpp-within\endcsname{1.29}
\expandafter\def\csname N@cg-qwen-otp-lpp-gap\endcsname{1.01}
\expandafter\def\csname N@cg-qwen-otp-lpp-wpsd\endcsname{1.03}
\expandafter\def\csname N@cg-qwen-otp-lpp-edp\endcsname{0.021}
\expandafter\def\csname N@cg-qwen-otp-sel-corr\endcsname{1.46--1.87}
\expandafter\def\csname N@cg-qwen-otp-sel-rel\endcsname{0.33--0.59}
\expandafter\def\csname N@cg-qwen-otp-sel-n\endcsname{7}
\expandafter\def\csname N@cg-qwen-otp-sel-nworse\endcsname{1}
\expandafter\def\csname N@cg-qwen-otp-sel-divbest-corr\endcsname{1.49}
\expandafter\def\csname N@cg-qwen-otp-sel-divbest-rel\endcsname{0.42}
\expandafter\def\csname N@cg-qwen-otp-sel-divbest-cross\endcsname{1.00}
\expandafter\def\csname N@cg-qwen-otp-sel-divbest-d\endcsname{0.19}
\expandafter\def\csname N@cg-qwen-otp-sel-divbest-ci\endcsname{$[-0.13, 0.49]$}
\expandafter\def\csname N@cg-qwen-otp-sel-divbest-pwin\endcsname{12\%}
\expandafter\def\csname N@cg-qwen-otp-sel-divbest-edp\endcsname{0.030}
\expandafter\def\csname N@cg-qwen-otp-sel-divbest-edpci\endcsname{$[-0.087, 0.067]$}
\expandafter\def\csname N@cg-qwen-otp-sel-rw-corr\endcsname{1.67}
\expandafter\def\csname N@cg-qwen-otp-sel-rw-rel\endcsname{0.52}
\expandafter\def\csname N@cg-qwen-otp-sel-rw-cross\endcsname{1.23}
\expandafter\def\csname N@cg-qwen-otp-sel-rw-d\endcsname{0.00}
\expandafter\def\csname N@cg-qwen-otp-sel-rw-ci\endcsname{$[-0.31, 0.36]$}
\expandafter\def\csname N@cg-qwen-otp-sel-rw-pwin\endcsname{50\%}
\expandafter\def\csname N@cg-qwen-otp-sel-rw-edp\endcsname{0.037}
\expandafter\def\csname N@cg-qwen-otp-sel-rw-edpci\endcsname{$[-0.106, 0.053]$}
\expandafter\def\csname N@cg-qwen-otp-sel-base-corr\endcsname{1.48}
\expandafter\def\csname N@cg-qwen-otp-sel-base-rel\endcsname{0.34}
\expandafter\def\csname N@cg-qwen-otp-sel-base-cross\endcsname{0.86}
\expandafter\def\csname N@cg-qwen-otp-sel-base-d\endcsname{0.19}
\expandafter\def\csname N@cg-qwen-otp-sel-base-ci\endcsname{$[-0.18, 0.54]$}
\expandafter\def\csname N@cg-qwen-otp-sel-base-pwin\endcsname{12\%}
\expandafter\def\csname N@cg-qwen-otp-sel-base-edp\endcsname{0.025}
\expandafter\def\csname N@cg-qwen-otp-sel-base-edpci\endcsname{$[-0.084, 0.069]$}
\expandafter\def\csname N@cg-qwen-otp-sel-div-corr\endcsname{1.52}
\expandafter\def\csname N@cg-qwen-otp-sel-div-rel\endcsname{0.44}
\expandafter\def\csname N@cg-qwen-otp-sel-div-cross\endcsname{1.06}
\expandafter\def\csname N@cg-qwen-otp-sel-div-d\endcsname{0.16}
\expandafter\def\csname N@cg-qwen-otp-sel-div-ci\endcsname{$[-0.11, 0.46]$}
\expandafter\def\csname N@cg-qwen-otp-sel-div-pwin\endcsname{20\%}
\expandafter\def\csname N@cg-qwen-otp-sel-div-edp\endcsname{0.025}
\expandafter\def\csname N@cg-qwen-otp-sel-div-edpci\endcsname{$[-0.070, 0.066]$}
\expandafter\def\csname N@cg-qwen-otp-sel-argyle-corr\endcsname{1.51}
\expandafter\def\csname N@cg-qwen-otp-sel-argyle-rel\endcsname{0.33}
\expandafter\def\csname N@cg-qwen-otp-sel-argyle-cross\endcsname{0.86}
\expandafter\def\csname N@cg-qwen-otp-sel-argyle-d\endcsname{0.17}
\expandafter\def\csname N@cg-qwen-otp-sel-argyle-ci\endcsname{$[-0.16, 0.51]$}
\expandafter\def\csname N@cg-qwen-otp-sel-argyle-pwin\endcsname{16\%}
\expandafter\def\csname N@cg-qwen-otp-sel-argyle-edp\endcsname{0.028}
\expandafter\def\csname N@cg-qwen-otp-sel-argyle-edpci\endcsname{$[-0.077, 0.069]$}
\expandafter\def\csname N@cg-qwen-otp-sel-story-corr\endcsname{1.46}
\expandafter\def\csname N@cg-qwen-otp-sel-story-rel\endcsname{0.43}
\expandafter\def\csname N@cg-qwen-otp-sel-story-cross\endcsname{1.02}
\expandafter\def\csname N@cg-qwen-otp-sel-story-d\endcsname{0.22}
\expandafter\def\csname N@cg-qwen-otp-sel-story-ci\endcsname{$[-0.06, 0.52]$}
\expandafter\def\csname N@cg-qwen-otp-sel-story-pwin\endcsname{6\%}
\expandafter\def\csname N@cg-qwen-otp-sel-story-edp\endcsname{0.023}
\expandafter\def\csname N@cg-qwen-otp-sel-story-edpci\endcsname{$[-0.071, 0.077]$}
\expandafter\def\csname N@cg-qwen-otp-sel-anth-corr\endcsname{1.87}
\expandafter\def\csname N@cg-qwen-otp-sel-anth-rel\endcsname{0.59}
\expandafter\def\csname N@cg-qwen-otp-sel-anth-cross\endcsname{0.66}
\expandafter\def\csname N@cg-qwen-otp-sel-anth-d\endcsname{$-$0.20}
\expandafter\def\csname N@cg-qwen-otp-sel-anth-ci\endcsname{$[-0.61, 0.28]$}
\expandafter\def\csname N@cg-qwen-otp-sel-anth-pwin\endcsname{82\%}
\expandafter\def\csname N@cg-qwen-otp-sel-anth-edp\endcsname{0.031}
\expandafter\def\csname N@cg-qwen-otp-sel-anth-edpci\endcsname{$[-0.088, 0.058]$}
\expandafter\def\csname N@cg-qwen-otp-sel-drange\endcsname{$-$0.20 to +0.22}
\expandafter\def\csname N@cg-qwen-otp-sel-nsig\endcsname{0}
\expandafter\def\csname N@cg-qwen-otp-sel-relx\endcsname{1.6--2.9}
\expandafter\def\csname N@cg-qwen-otp-sel-pwinmin\endcsname{6\%}
\expandafter\def\csname N@cg-qwen-otp-sel-pwinphrase\endcsname{at least 6\% of}
\expandafter\def\csname N@cg-qwen-otp-shp-corr\endcsname{2.24--2.54}
\expandafter\def\csname N@cg-qwen-otp-shp-rel\endcsname{0.93--0.99}
\expandafter\def\csname N@cg-qwen-otp-shp-n\endcsname{4}
\expandafter\def\csname N@cg-qwen-otp-shp-nworse\endcsname{4}
\expandafter\def\csname N@cg-qwen-otp-shp-base-corr\endcsname{2.54}
\expandafter\def\csname N@cg-qwen-otp-shp-base-rel\endcsname{0.94}
\expandafter\def\csname N@cg-qwen-otp-shp-base-cross\endcsname{1.66}
\expandafter\def\csname N@cg-qwen-otp-shp-base-d\endcsname{$-$0.86}
\expandafter\def\csname N@cg-qwen-otp-shp-base-ci\endcsname{$[-1.16, -0.51]$}
\expandafter\def\csname N@cg-qwen-otp-shp-base-pwin\endcsname{100\%}
\expandafter\def\csname N@cg-qwen-otp-shp-base-edp\endcsname{0.062}
\expandafter\def\csname N@cg-qwen-otp-shp-base-edpci\endcsname{$[-0.107, 0.028]$}
\expandafter\def\csname N@cg-qwen-otp-shp-div-corr\endcsname{2.24}
\expandafter\def\csname N@cg-qwen-otp-shp-div-rel\endcsname{0.93}
\expandafter\def\csname N@cg-qwen-otp-shp-div-cross\endcsname{1.59}
\expandafter\def\csname N@cg-qwen-otp-shp-div-d\endcsname{$-$0.56}
\expandafter\def\csname N@cg-qwen-otp-shp-div-ci\endcsname{$[-0.86, -0.25]$}
\expandafter\def\csname N@cg-qwen-otp-shp-div-pwin\endcsname{100\%}
\expandafter\def\csname N@cg-qwen-otp-shp-div-edp\endcsname{0.061}
\expandafter\def\csname N@cg-qwen-otp-shp-div-edpci\endcsname{$[-0.107, 0.029]$}
\expandafter\def\csname N@cg-qwen-otp-shp-argyle-corr\endcsname{2.51}
\expandafter\def\csname N@cg-qwen-otp-shp-argyle-rel\endcsname{0.95}
\expandafter\def\csname N@cg-qwen-otp-shp-argyle-cross\endcsname{1.86}
\expandafter\def\csname N@cg-qwen-otp-shp-argyle-d\endcsname{$-$0.83}
\expandafter\def\csname N@cg-qwen-otp-shp-argyle-ci\endcsname{$[-1.13, -0.40]$}
\expandafter\def\csname N@cg-qwen-otp-shp-argyle-pwin\endcsname{100\%}
\expandafter\def\csname N@cg-qwen-otp-shp-argyle-edp\endcsname{0.066}
\expandafter\def\csname N@cg-qwen-otp-shp-argyle-edpci\endcsname{$[-0.109, 0.018]$}
\expandafter\def\csname N@cg-qwen-otp-shp-story-corr\endcsname{2.29}
\expandafter\def\csname N@cg-qwen-otp-shp-story-rel\endcsname{0.99}
\expandafter\def\csname N@cg-qwen-otp-shp-story-cross\endcsname{1.65}
\expandafter\def\csname N@cg-qwen-otp-shp-story-d\endcsname{$-$0.62}
\expandafter\def\csname N@cg-qwen-otp-shp-story-ci\endcsname{$[-0.94, -0.27]$}
\expandafter\def\csname N@cg-qwen-otp-shp-story-pwin\endcsname{100\%}
\expandafter\def\csname N@cg-qwen-otp-shp-story-edp\endcsname{0.047}
\expandafter\def\csname N@cg-qwen-otp-shp-story-edpci\endcsname{$[-0.093, 0.046]$}
\expandafter\def\csname N@cg-qwen-otp-shp-drange\endcsname{$-$0.86 to $-$0.56}
\expandafter\def\csname N@cg-qwen-otp-shp-nsig\endcsname{4}
\expandafter\def\csname N@cg-qwen-otp-shp-relx\endcsname{1.0}
\expandafter\def\csname N@cg-qwen-otp-shp-pwinmin\endcsname{100\%}
\expandafter\def\csname N@cg-qwen-otp-shp-pwinphrase\endcsname{more than 99\% of}
\expandafter\def\csname N@cg-qwen-otp-abl-k16-corr\endcsname{1.60}
\expandafter\def\csname N@cg-qwen-otp-abl-k16-cross\endcsname{0.94}
\expandafter\def\csname N@cg-qwen-otp-abl-k16-pwin\endcsname{30.0\%}
\expandafter\def\csname N@cg-qwen-otp-abl-k16-gap\endcsname{1.01}
\expandafter\def\csname N@cg-qwen-otp-abl-k16-dcross\endcsname{0.16}
\expandafter\def\csname N@cg-qwen-otp-abl-k16-dcross-ci\endcsname{$[-0.06, 0.41]$}
\expandafter\def\csname N@cg-qwen-otp-abl-k16-dgap\endcsname{$-$0.00}
\expandafter\def\csname N@cg-qwen-otp-abl-k16-dgap-ci\endcsname{$[-0.22, 0.27]$}
\expandafter\def\csname N@cg-qwen-otp-abl-k16-dcorr\endcsname{0.08}
\expandafter\def\csname N@cg-qwen-otp-abl-k16-dcorr-ci\endcsname{$[-0.22, 0.36]$}
\expandafter\def\csname N@cg-qwen-otp-abl-k16-rel\endcsname{0.84}
\expandafter\def\csname N@cg-qwen-otp-abl-novals-corr\endcsname{1.78}
\expandafter\def\csname N@cg-qwen-otp-abl-novals-cross\endcsname{1.15}
\expandafter\def\csname N@cg-qwen-otp-abl-novals-pwin\endcsname{78.0\%}
\expandafter\def\csname N@cg-qwen-otp-abl-novals-gap\endcsname{1.01}
\expandafter\def\csname N@cg-qwen-otp-abl-novals-dcross\endcsname{$-$0.05}
\expandafter\def\csname N@cg-qwen-otp-abl-novals-dcross-ci\endcsname{$[-0.30, 0.21]$}
\expandafter\def\csname N@cg-qwen-otp-abl-novals-dgap\endcsname{0.00}
\expandafter\def\csname N@cg-qwen-otp-abl-novals-dgap-ci\endcsname{$[-0.22, 0.26]$}
\expandafter\def\csname N@cg-qwen-otp-abl-novals-dcorr\endcsname{$-$0.10}
\expandafter\def\csname N@cg-qwen-otp-abl-novals-dcorr-ci\endcsname{$[-0.36, 0.19]$}
\expandafter\def\csname N@cg-qwen-otp-abl-novals-rel\endcsname{1.02}
\expandafter\def\csname N@cg-qwen-otp-abl-nodev-corr\endcsname{1.85}
\expandafter\def\csname N@cg-qwen-otp-abl-nodev-cross\endcsname{1.44}
\expandafter\def\csname N@cg-qwen-otp-abl-nodev-pwin\endcsname{87.5\%}
\expandafter\def\csname N@cg-qwen-otp-abl-nodev-gap\endcsname{1.01}
\expandafter\def\csname N@cg-qwen-otp-abl-nodev-dcross\endcsname{$-$0.34}
\expandafter\def\csname N@cg-qwen-otp-abl-nodev-dcross-ci\endcsname{$[-0.58, -0.09]$}
\expandafter\def\csname N@cg-qwen-otp-abl-nodev-dgap\endcsname{$-$0.00}
\expandafter\def\csname N@cg-qwen-otp-abl-nodev-dgap-ci\endcsname{$[-0.27, 0.28]$}
\expandafter\def\csname N@cg-qwen-otp-abl-nodev-dcorr\endcsname{$-$0.17}
\expandafter\def\csname N@cg-qwen-otp-abl-nodev-dcorr-ci\endcsname{$[-0.44, 0.17]$}
\expandafter\def\csname N@cg-qwen-otp-abl-nodev-rel\endcsname{0.83}
\expandafter\def\csname N@cg-qwen-otp-abl-nosharp-corr\endcsname{1.50}
\expandafter\def\csname N@cg-qwen-otp-abl-nosharp-cross\endcsname{0.83}
\expandafter\def\csname N@cg-qwen-otp-abl-nosharp-pwin\endcsname{12.0\%}
\expandafter\def\csname N@cg-qwen-otp-abl-nosharp-gap\endcsname{1.02}
\expandafter\def\csname N@cg-qwen-otp-abl-nosharp-dcross\endcsname{0.27}
\expandafter\def\csname N@cg-qwen-otp-abl-nosharp-dcross-ci\endcsname{$[0.09, 0.46]$}
\expandafter\def\csname N@cg-qwen-otp-abl-nosharp-dgap\endcsname{$-$0.01}
\expandafter\def\csname N@cg-qwen-otp-abl-nosharp-dgap-ci\endcsname{$[-0.19, 0.16]$}
\expandafter\def\csname N@cg-qwen-otp-abl-nosharp-dcorr\endcsname{0.18}
\expandafter\def\csname N@cg-qwen-otp-abl-nosharp-dcorr-ci\endcsname{$[-0.11, 0.50]$}
\expandafter\def\csname N@cg-qwen-otp-abl-nosharp-rel\endcsname{0.54}
\expandafter\def\csname N@cg-qwen-otp-agg-corr\endcsname{1.97}
\expandafter\def\csname N@cg-qwen-otp-agg-cross\endcsname{0.49}
\expandafter\def\csname N@cg-qwen-otp-agg-pwin\endcsname{91.0\%}
\expandafter\def\csname N@cg-qwen-otp-agg-gap\endcsname{1.02}
\expandafter\def\csname N@cg-qwen-otp-agg-dcross\endcsname{0.61}
\expandafter\def\csname N@cg-qwen-otp-agg-dcross-ci\endcsname{$[0.34, 0.88]$}
\expandafter\def\csname N@cg-qwen-otp-agg-dgap\endcsname{$-$0.01}
\expandafter\def\csname N@cg-qwen-otp-agg-dgap-ci\endcsname{$[-0.28, 0.27]$}
\expandafter\def\csname N@cg-qwen-otp-agg-dcorr\endcsname{$-$0.30}
\expandafter\def\csname N@cg-qwen-otp-agg-dcorr-ci\endcsname{$[-0.73, 0.15]$}
\expandafter\def\csname N@cg-qwen-otp-agg-rel\endcsname{0.13}
\expandafter\def\csname N@cg-qwen-otp-abl-abs-corr\endcsname{1.65}
\expandafter\def\csname N@cg-qwen-otp-abl-abs-cross\endcsname{0.84}
\expandafter\def\csname N@cg-qwen-otp-abl-abs-pwin\endcsname{41.5\%}
\expandafter\def\csname N@cg-qwen-otp-abl-abs-gap\endcsname{1.00}
\expandafter\def\csname N@cg-qwen-otp-abl-abs-dcross\endcsname{0.26}
\expandafter\def\csname N@cg-qwen-otp-abl-abs-dcross-ci\endcsname{$[0.03, 0.54]$}
\expandafter\def\csname N@cg-qwen-otp-abl-abs-dgap\endcsname{0.01}
\expandafter\def\csname N@cg-qwen-otp-abl-abs-dgap-ci\endcsname{$[-0.24, 0.24]$}
\expandafter\def\csname N@cg-qwen-otp-abl-abs-dcorr\endcsname{0.02}
\expandafter\def\csname N@cg-qwen-otp-abl-abs-dcorr-ci\endcsname{$[-0.29, 0.32]$}
\expandafter\def\csname N@cg-qwen-otp-abl-abs-rel\endcsname{1.06}
\expandafter\def\csname N@cg-qwen-otp-abl-placebo-corr\endcsname{2.10}
\expandafter\def\csname N@cg-qwen-otp-abl-placebo-cross\endcsname{1.52}
\expandafter\def\csname N@cg-qwen-otp-abl-placebo-pwin\endcsname{99.0\%}
\expandafter\def\csname N@cg-qwen-otp-abl-placebo-gap\endcsname{1.01}
\expandafter\def\csname N@cg-qwen-otp-abl-placebo-dcross\endcsname{$-$0.42}
\expandafter\def\csname N@cg-qwen-otp-abl-placebo-dcross-ci\endcsname{$[-0.67, -0.13]$}
\expandafter\def\csname N@cg-qwen-otp-abl-placebo-dgap\endcsname{$-$0.00}
\expandafter\def\csname N@cg-qwen-otp-abl-placebo-dgap-ci\endcsname{$[-0.24, 0.23]$}
\expandafter\def\csname N@cg-qwen-otp-abl-placebo-dcorr\endcsname{$-$0.42}
\expandafter\def\csname N@cg-qwen-otp-abl-placebo-dcorr-ci\endcsname{$[-0.78, -0.07]$}
\expandafter\def\csname N@cg-qwen-otp-abl-placebo-rel\endcsname{0.80}
\expandafter\def\csname N@cg-qwen-otp-abl-seed2-corr\endcsname{1.67}
\expandafter\def\csname N@cg-qwen-otp-abl-seed2-cross\endcsname{1.13}
\expandafter\def\csname N@cg-qwen-otp-abl-seed2-pwin\endcsname{49.0\%}
\expandafter\def\csname N@cg-qwen-otp-abl-seed2-gap\endcsname{1.02}
\expandafter\def\csname N@cg-qwen-otp-abl-seed2-dcross\endcsname{$-$0.02}
\expandafter\def\csname N@cg-qwen-otp-abl-seed2-dcross-ci\endcsname{$[-0.29, 0.21]$}
\expandafter\def\csname N@cg-qwen-otp-abl-seed2-dgap\endcsname{$-$0.01}
\expandafter\def\csname N@cg-qwen-otp-abl-seed2-dgap-ci\endcsname{$[-0.23, 0.24]$}
\expandafter\def\csname N@cg-qwen-otp-abl-seed2-dcorr\endcsname{0.00}
\expandafter\def\csname N@cg-qwen-otp-abl-seed2-dcorr-ci\endcsname{$[-0.24, 0.30]$}
\expandafter\def\csname N@cg-qwen-otp-abl-seed2-rel\endcsname{0.95}
\expandafter\def\csname N@cg-qwen-otp-abl-englow-corr\endcsname{1.60}
\expandafter\def\csname N@cg-qwen-otp-abl-englow-cross\endcsname{0.90}
\expandafter\def\csname N@cg-qwen-otp-abl-englow-pwin\endcsname{30.5\%}
\expandafter\def\csname N@cg-qwen-otp-abl-englow-gap\endcsname{1.00}
\expandafter\def\csname N@cg-qwen-otp-abl-englow-dcross\endcsname{0.20}
\expandafter\def\csname N@cg-qwen-otp-abl-englow-dcross-ci\endcsname{$[-0.03, 0.49]$}
\expandafter\def\csname N@cg-qwen-otp-abl-englow-dgap\endcsname{0.01}
\expandafter\def\csname N@cg-qwen-otp-abl-englow-dgap-ci\endcsname{$[-0.28, 0.28]$}
\expandafter\def\csname N@cg-qwen-otp-abl-englow-dcorr\endcsname{0.07}
\expandafter\def\csname N@cg-qwen-otp-abl-englow-dcorr-ci\endcsname{$[-0.21, 0.35]$}
\expandafter\def\csname N@cg-qwen-otp-abl-englow-rel\endcsname{0.94}
\expandafter\def\csname N@cg-qwen-otp-abl-enghigh-corr\endcsname{1.63}
\expandafter\def\csname N@cg-qwen-otp-abl-enghigh-cross\endcsname{1.17}
\expandafter\def\csname N@cg-qwen-otp-abl-enghigh-pwin\endcsname{39.0\%}
\expandafter\def\csname N@cg-qwen-otp-abl-enghigh-gap\endcsname{1.02}
\expandafter\def\csname N@cg-qwen-otp-abl-enghigh-dcross\endcsname{$-$0.07}
\expandafter\def\csname N@cg-qwen-otp-abl-enghigh-dcross-ci\endcsname{$[-0.32, 0.19]$}
\expandafter\def\csname N@cg-qwen-otp-abl-enghigh-dgap\endcsname{$-$0.00}
\expandafter\def\csname N@cg-qwen-otp-abl-enghigh-dgap-ci\endcsname{$[-0.21, 0.28]$}
\expandafter\def\csname N@cg-qwen-otp-abl-enghigh-dcorr\endcsname{0.04}
\expandafter\def\csname N@cg-qwen-otp-abl-enghigh-dcorr-ci\endcsname{$[-0.23, 0.40]$}
\expandafter\def\csname N@cg-qwen-otp-abl-enghigh-rel\endcsname{0.93}
\expandafter\def\csname N@cg-qwen-otp-abl-demsharp-corr\endcsname{2.46}
\expandafter\def\csname N@cg-qwen-otp-abl-demsharp-cross\endcsname{1.63}
\expandafter\def\csname N@cg-qwen-otp-abl-demsharp-pwin\endcsname{100.0\%}
\expandafter\def\csname N@cg-qwen-otp-abl-demsharp-gap\endcsname{1.01}
\expandafter\def\csname N@cg-qwen-otp-abl-demsharp-dcross\endcsname{$-$0.53}
\expandafter\def\csname N@cg-qwen-otp-abl-demsharp-dcross-ci\endcsname{$[-0.79, -0.28]$}
\expandafter\def\csname N@cg-qwen-otp-abl-demsharp-dgap\endcsname{$-$0.00}
\expandafter\def\csname N@cg-qwen-otp-abl-demsharp-dgap-ci\endcsname{$[-0.23, 0.26]$}
\expandafter\def\csname N@cg-qwen-otp-abl-demsharp-dcorr\endcsname{$-$0.78}
\expandafter\def\csname N@cg-qwen-otp-abl-demsharp-dcorr-ci\endcsname{$[-1.08, -0.40]$}
\expandafter\def\csname N@cg-qwen-otp-abl-demsharp-rel\endcsname{0.86}
\expandafter\def\csname N@cg-qwen-otp-demo-corr\endcsname{1.65}
\expandafter\def\csname N@cg-qwen-otp-demo-cross\endcsname{1.21}
\expandafter\def\csname N@cg-qwen-otp-demo-pwin\endcsname{42.0\%}
\expandafter\def\csname N@cg-qwen-otp-demo-gap\endcsname{1.02}
\expandafter\def\csname N@cg-qwen-otp-demo-dcross\endcsname{$-$0.11}
\expandafter\def\csname N@cg-qwen-otp-demo-dcross-ci\endcsname{$[-0.36, 0.12]$}
\expandafter\def\csname N@cg-qwen-otp-demo-dgap\endcsname{$-$0.01}
\expandafter\def\csname N@cg-qwen-otp-demo-dgap-ci\endcsname{$[-0.22, 0.26]$}
\expandafter\def\csname N@cg-qwen-otp-demo-dcorr\endcsname{0.03}
\expandafter\def\csname N@cg-qwen-otp-demo-dcorr-ci\endcsname{$[-0.27, 0.34]$}
\expandafter\def\csname N@cg-qwen-otp-demo-rel\endcsname{0.52}
\expandafter\def\csname N@cg-qwen-pool-floor-corr\endcsname{1.10}
\expandafter\def\csname N@cg-qwen-pool-floor-cross\endcsname{1.06}
\expandafter\def\csname N@cg-qwen-pool-lpp-corr\endcsname{1.85}
\expandafter\def\csname N@cg-qwen-pool-lpp-rel\endcsname{0.96}
\expandafter\def\csname N@cg-qwen-pool-lpp-cross\endcsname{1.17}
\expandafter\def\csname N@cg-qwen-pool-lpp-within\endcsname{1.33}
\expandafter\def\csname N@cg-qwen-pool-lpp-gap\endcsname{1.12}
\expandafter\def\csname N@cg-qwen-pool-lpp-wpsd\endcsname{1.02}
\expandafter\def\csname N@cg-qwen-pool-lpp-edp\endcsname{0.079}
\expandafter\def\csname N@cg-qwen-pool-sel-corr\endcsname{1.59--2.40}
\expandafter\def\csname N@cg-qwen-pool-sel-rel\endcsname{0.39--0.62}
\expandafter\def\csname N@cg-qwen-pool-sel-n\endcsname{7}
\expandafter\def\csname N@cg-qwen-pool-sel-nworse\endcsname{3}
\expandafter\def\csname N@cg-qwen-pool-sel-divbest-corr\endcsname{1.75}
\expandafter\def\csname N@cg-qwen-pool-sel-divbest-rel\endcsname{0.49}
\expandafter\def\csname N@cg-qwen-pool-sel-divbest-cross\endcsname{1.25}
\expandafter\def\csname N@cg-qwen-pool-sel-divbest-d\endcsname{0.11}
\expandafter\def\csname N@cg-qwen-pool-sel-divbest-ci\endcsname{$[-0.24, 0.45]$}
\expandafter\def\csname N@cg-qwen-pool-sel-divbest-pwin\endcsname{29\%}
\expandafter\def\csname N@cg-qwen-pool-sel-divbest-edp\endcsname{0.168}
\expandafter\def\csname N@cg-qwen-pool-sel-divbest-edpci\endcsname{$[-0.206, 0.022]$}
\expandafter\def\csname N@cg-qwen-pool-sel-rw-corr\endcsname{2.19}
\expandafter\def\csname N@cg-qwen-pool-sel-rw-rel\endcsname{0.62}
\expandafter\def\csname N@cg-qwen-pool-sel-rw-cross\endcsname{1.54}
\expandafter\def\csname N@cg-qwen-pool-sel-rw-d\endcsname{$-$0.34}
\expandafter\def\csname N@cg-qwen-pool-sel-rw-ci\endcsname{$[-0.74, 0.05]$}
\expandafter\def\csname N@cg-qwen-pool-sel-rw-pwin\endcsname{94\%}
\expandafter\def\csname N@cg-qwen-pool-sel-rw-edp\endcsname{0.246}
\expandafter\def\csname N@cg-qwen-pool-sel-rw-edpci\endcsname{$[-0.351, 0.002]$}
\expandafter\def\csname N@cg-qwen-pool-sel-base-corr\endcsname{1.59}
\expandafter\def\csname N@cg-qwen-pool-sel-base-rel\endcsname{0.39}
\expandafter\def\csname N@cg-qwen-pool-sel-base-cross\endcsname{1.04}
\expandafter\def\csname N@cg-qwen-pool-sel-base-d\endcsname{0.26}
\expandafter\def\csname N@cg-qwen-pool-sel-base-ci\endcsname{$[-0.12, 0.60]$}
\expandafter\def\csname N@cg-qwen-pool-sel-base-pwin\endcsname{7\%}
\expandafter\def\csname N@cg-qwen-pool-sel-base-edp\endcsname{0.128}
\expandafter\def\csname N@cg-qwen-pool-sel-base-edpci\endcsname{$[-0.161, 0.047]$}
\expandafter\def\csname N@cg-qwen-pool-sel-div-corr\endcsname{1.90}
\expandafter\def\csname N@cg-qwen-pool-sel-div-rel\endcsname{0.54}
\expandafter\def\csname N@cg-qwen-pool-sel-div-cross\endcsname{1.36}
\expandafter\def\csname N@cg-qwen-pool-sel-div-d\endcsname{$-$0.05}
\expandafter\def\csname N@cg-qwen-pool-sel-div-ci\endcsname{$[-0.40, 0.29]$}
\expandafter\def\csname N@cg-qwen-pool-sel-div-pwin\endcsname{62\%}
\expandafter\def\csname N@cg-qwen-pool-sel-div-edp\endcsname{0.203}
\expandafter\def\csname N@cg-qwen-pool-sel-div-edpci\endcsname{$[-0.250, -0.010]$}
\expandafter\def\csname N@cg-qwen-pool-sel-argyle-corr\endcsname{1.83}
\expandafter\def\csname N@cg-qwen-pool-sel-argyle-rel\endcsname{0.48}
\expandafter\def\csname N@cg-qwen-pool-sel-argyle-cross\endcsname{1.28}
\expandafter\def\csname N@cg-qwen-pool-sel-argyle-d\endcsname{0.03}
\expandafter\def\csname N@cg-qwen-pool-sel-argyle-ci\endcsname{$[-0.28, 0.36]$}
\expandafter\def\csname N@cg-qwen-pool-sel-argyle-pwin\endcsname{42\%}
\expandafter\def\csname N@cg-qwen-pool-sel-argyle-edp\endcsname{0.295}
\expandafter\def\csname N@cg-qwen-pool-sel-argyle-edpci\endcsname{$[-0.349, -0.080]$}
\expandafter\def\csname N@cg-qwen-pool-sel-story-corr\endcsname{1.67}
\expandafter\def\csname N@cg-qwen-pool-sel-story-rel\endcsname{0.49}
\expandafter\def\csname N@cg-qwen-pool-sel-story-cross\endcsname{1.25}
\expandafter\def\csname N@cg-qwen-pool-sel-story-d\endcsname{0.18}
\expandafter\def\csname N@cg-qwen-pool-sel-story-ci\endcsname{$[-0.13, 0.48]$}
\expandafter\def\csname N@cg-qwen-pool-sel-story-pwin\endcsname{12\%}
\expandafter\def\csname N@cg-qwen-pool-sel-story-edp\endcsname{0.112}
\expandafter\def\csname N@cg-qwen-pool-sel-story-edpci\endcsname{$[-0.148, 0.070]$}
\expandafter\def\csname N@cg-qwen-pool-sel-anth-corr\endcsname{2.40}
\expandafter\def\csname N@cg-qwen-pool-sel-anth-rel\endcsname{0.52}
\expandafter\def\csname N@cg-qwen-pool-sel-anth-cross\endcsname{0.55}
\expandafter\def\csname N@cg-qwen-pool-sel-anth-d\endcsname{$-$0.55}
\expandafter\def\csname N@cg-qwen-pool-sel-anth-ci\endcsname{$[-1.05, -0.06]$}
\expandafter\def\csname N@cg-qwen-pool-sel-anth-pwin\endcsname{98\%}
\expandafter\def\csname N@cg-qwen-pool-sel-anth-edp\endcsname{0.229}
\expandafter\def\csname N@cg-qwen-pool-sel-anth-edpci\endcsname{$[-0.316, 0.040]$}
\expandafter\def\csname N@cg-qwen-pool-sel-drange\endcsname{$-$0.55 to +0.26}
\expandafter\def\csname N@cg-qwen-pool-sel-nsig\endcsname{1}
\expandafter\def\csname N@cg-qwen-pool-sel-relx\endcsname{1.6--2.4}
\expandafter\def\csname N@cg-qwen-pool-sel-pwinmin\endcsname{7\%}
\expandafter\def\csname N@cg-qwen-pool-sel-pwinphrase\endcsname{at least 7\% of}
\expandafter\def\csname N@cg-qwen-pool-shp-corr\endcsname{2.88--3.77}
\expandafter\def\csname N@cg-qwen-pool-shp-rel\endcsname{1.01--1.12}
\expandafter\def\csname N@cg-qwen-pool-shp-n\endcsname{4}
\expandafter\def\csname N@cg-qwen-pool-shp-nworse\endcsname{4}
\expandafter\def\csname N@cg-qwen-pool-shp-base-corr\endcsname{3.09}
\expandafter\def\csname N@cg-qwen-pool-shp-base-rel\endcsname{1.01}
\expandafter\def\csname N@cg-qwen-pool-shp-base-cross\endcsname{1.93}
\expandafter\def\csname N@cg-qwen-pool-shp-base-d\endcsname{$-$1.23}
\expandafter\def\csname N@cg-qwen-pool-shp-base-ci\endcsname{$[-1.52, -0.96]$}
\expandafter\def\csname N@cg-qwen-pool-shp-base-pwin\endcsname{100\%}
\expandafter\def\csname N@cg-qwen-pool-shp-base-edp\endcsname{0.402}
\expandafter\def\csname N@cg-qwen-pool-shp-base-edpci\endcsname{$[-0.484, -0.184]$}
\expandafter\def\csname N@cg-qwen-pool-shp-div-corr\endcsname{2.88}
\expandafter\def\csname N@cg-qwen-pool-shp-div-rel\endcsname{1.01}
\expandafter\def\csname N@cg-qwen-pool-shp-div-cross\endcsname{1.91}
\expandafter\def\csname N@cg-qwen-pool-shp-div-d\endcsname{$-$1.03}
\expandafter\def\csname N@cg-qwen-pool-shp-div-ci\endcsname{$[-1.32, -0.76]$}
\expandafter\def\csname N@cg-qwen-pool-shp-div-pwin\endcsname{100\%}
\expandafter\def\csname N@cg-qwen-pool-shp-div-edp\endcsname{0.383}
\expandafter\def\csname N@cg-qwen-pool-shp-div-edpci\endcsname{$[-0.464, -0.165]$}
\expandafter\def\csname N@cg-qwen-pool-shp-argyle-corr\endcsname{3.77}
\expandafter\def\csname N@cg-qwen-pool-shp-argyle-rel\endcsname{1.12}
\expandafter\def\csname N@cg-qwen-pool-shp-argyle-cross\endcsname{2.54}
\expandafter\def\csname N@cg-qwen-pool-shp-argyle-d\endcsname{$-$1.92}
\expandafter\def\csname N@cg-qwen-pool-shp-argyle-ci\endcsname{$[-2.25, -1.61]$}
\expandafter\def\csname N@cg-qwen-pool-shp-argyle-pwin\endcsname{100\%}
\expandafter\def\csname N@cg-qwen-pool-shp-argyle-edp\endcsname{0.824}
\expandafter\def\csname N@cg-qwen-pool-shp-argyle-edpci\endcsname{$[-0.926, -0.564]$}
\expandafter\def\csname N@cg-qwen-pool-shp-story-corr\endcsname{2.98}
\expandafter\def\csname N@cg-qwen-pool-shp-story-rel\endcsname{1.07}
\expandafter\def\csname N@cg-qwen-pool-shp-story-cross\endcsname{2.04}
\expandafter\def\csname N@cg-qwen-pool-shp-story-d\endcsname{$-$1.13}
\expandafter\def\csname N@cg-qwen-pool-shp-story-ci\endcsname{$[-1.44, -0.83]$}
\expandafter\def\csname N@cg-qwen-pool-shp-story-pwin\endcsname{100\%}
\expandafter\def\csname N@cg-qwen-pool-shp-story-edp\endcsname{0.348}
\expandafter\def\csname N@cg-qwen-pool-shp-story-edpci\endcsname{$[-0.426, -0.133]$}
\expandafter\def\csname N@cg-qwen-pool-shp-drange\endcsname{$-$1.92 to $-$1.03}
\expandafter\def\csname N@cg-qwen-pool-shp-nsig\endcsname{4}
\expandafter\def\csname N@cg-qwen-pool-shp-relx\endcsname{0.9--1.0}
\expandafter\def\csname N@cg-qwen-pool-shp-pwinmin\endcsname{100\%}
\expandafter\def\csname N@cg-qwen-pool-shp-pwinphrase\endcsname{all}
\expandafter\def\csname N@cg-qwen-pool-abl-k16-corr\endcsname{1.76}
\expandafter\def\csname N@cg-qwen-pool-abl-k16-cross\endcsname{1.03}
\expandafter\def\csname N@cg-qwen-pool-abl-k16-pwin\endcsname{26.0\%}
\expandafter\def\csname N@cg-qwen-pool-abl-k16-gap\endcsname{1.22}
\expandafter\def\csname N@cg-qwen-pool-abl-k16-dcross\endcsname{0.14}
\expandafter\def\csname N@cg-qwen-pool-abl-k16-dcross-ci\endcsname{$[-0.08, 0.40]$}
\expandafter\def\csname N@cg-qwen-pool-abl-k16-dgap\endcsname{$-$0.10}
\expandafter\def\csname N@cg-qwen-pool-abl-k16-dgap-ci\endcsname{$[-0.33, 0.18]$}
\expandafter\def\csname N@cg-qwen-pool-abl-k16-dcorr\endcsname{0.09}
\expandafter\def\csname N@cg-qwen-pool-abl-k16-dcorr-ci\endcsname{$[-0.19, 0.38]$}
\expandafter\def\csname N@cg-qwen-pool-abl-k16-rel\endcsname{0.87}
\expandafter\def\csname N@cg-qwen-pool-abl-novals-corr\endcsname{2.06}
\expandafter\def\csname N@cg-qwen-pool-abl-novals-cross\endcsname{1.32}
\expandafter\def\csname N@cg-qwen-pool-abl-novals-pwin\endcsname{88.0\%}
\expandafter\def\csname N@cg-qwen-pool-abl-novals-gap\endcsname{1.30}
\expandafter\def\csname N@cg-qwen-pool-abl-novals-dcross\endcsname{$-$0.14}
\expandafter\def\csname N@cg-qwen-pool-abl-novals-dcross-ci\endcsname{$[-0.40, 0.09]$}
\expandafter\def\csname N@cg-qwen-pool-abl-novals-dgap\endcsname{$-$0.19}
\expandafter\def\csname N@cg-qwen-pool-abl-novals-dgap-ci\endcsname{$[-0.43, 0.05]$}
\expandafter\def\csname N@cg-qwen-pool-abl-novals-dcorr\endcsname{$-$0.21}
\expandafter\def\csname N@cg-qwen-pool-abl-novals-dcorr-ci\endcsname{$[-0.51, 0.10]$}
\expandafter\def\csname N@cg-qwen-pool-abl-novals-rel\endcsname{1.04}
\expandafter\def\csname N@cg-qwen-pool-abl-nodev-corr\endcsname{2.23}
\expandafter\def\csname N@cg-qwen-pool-abl-nodev-cross\endcsname{1.66}
\expandafter\def\csname N@cg-qwen-pool-abl-nodev-pwin\endcsname{98.5\%}
\expandafter\def\csname N@cg-qwen-pool-abl-nodev-gap\endcsname{1.42}
\expandafter\def\csname N@cg-qwen-pool-abl-nodev-dcross\endcsname{$-$0.49}
\expandafter\def\csname N@cg-qwen-pool-abl-nodev-dcross-ci\endcsname{$[-0.73, -0.25]$}
\expandafter\def\csname N@cg-qwen-pool-abl-nodev-dgap\endcsname{$-$0.30}
\expandafter\def\csname N@cg-qwen-pool-abl-nodev-dgap-ci\endcsname{$[-0.54, -0.05]$}
\expandafter\def\csname N@cg-qwen-pool-abl-nodev-dcorr\endcsname{$-$0.38}
\expandafter\def\csname N@cg-qwen-pool-abl-nodev-dcorr-ci\endcsname{$[-0.69, -0.04]$}
\expandafter\def\csname N@cg-qwen-pool-abl-nodev-rel\endcsname{0.88}
\expandafter\def\csname N@cg-qwen-pool-abl-nosharp-corr\endcsname{1.64}
\expandafter\def\csname N@cg-qwen-pool-abl-nosharp-cross\endcsname{0.82}
\expandafter\def\csname N@cg-qwen-pool-abl-nosharp-pwin\endcsname{10.5\%}
\expandafter\def\csname N@cg-qwen-pool-abl-nosharp-gap\endcsname{0.93}
\expandafter\def\csname N@cg-qwen-pool-abl-nosharp-dcross\endcsname{0.36}
\expandafter\def\csname N@cg-qwen-pool-abl-nosharp-dcross-ci\endcsname{$[0.19, 0.55]$}
\expandafter\def\csname N@cg-qwen-pool-abl-nosharp-dgap\endcsname{0.19}
\expandafter\def\csname N@cg-qwen-pool-abl-nosharp-dgap-ci\endcsname{$[0.01, 0.36]$}
\expandafter\def\csname N@cg-qwen-pool-abl-nosharp-dcorr\endcsname{0.22}
\expandafter\def\csname N@cg-qwen-pool-abl-nosharp-dcorr-ci\endcsname{$[-0.15, 0.56]$}
\expandafter\def\csname N@cg-qwen-pool-abl-nosharp-rel\endcsname{0.52}
\expandafter\def\csname N@cg-qwen-pool-agg-corr\endcsname{2.58}
\expandafter\def\csname N@cg-qwen-pool-agg-cross\endcsname{0.37}
\expandafter\def\csname N@cg-qwen-pool-agg-pwin\endcsname{99.0\%}
\expandafter\def\csname N@cg-qwen-pool-agg-gap\endcsname{0.25}
\expandafter\def\csname N@cg-qwen-pool-agg-dcross\endcsname{0.80}
\expandafter\def\csname N@cg-qwen-pool-agg-dcross-ci\endcsname{$[0.58, 1.10]$}
\expandafter\def\csname N@cg-qwen-pool-agg-dgap\endcsname{0.86}
\expandafter\def\csname N@cg-qwen-pool-agg-dgap-ci\endcsname{$[0.57, 1.22]$}
\expandafter\def\csname N@cg-qwen-pool-agg-dcorr\endcsname{$-$0.72}
\expandafter\def\csname N@cg-qwen-pool-agg-dcorr-ci\endcsname{$[-1.18, -0.21]$}
\expandafter\def\csname N@cg-qwen-pool-agg-rel\endcsname{0.00}
\expandafter\def\csname N@cg-qwen-pool-abl-abs-corr\endcsname{1.90}
\expandafter\def\csname N@cg-qwen-pool-abl-abs-cross\endcsname{0.81}
\expandafter\def\csname N@cg-qwen-pool-abl-abs-pwin\endcsname{60.0\%}
\expandafter\def\csname N@cg-qwen-pool-abl-abs-gap\endcsname{0.77}
\expandafter\def\csname N@cg-qwen-pool-abl-abs-dcross\endcsname{0.36}
\expandafter\def\csname N@cg-qwen-pool-abl-abs-dcross-ci\endcsname{$[0.10, 0.63]$}
\expandafter\def\csname N@cg-qwen-pool-abl-abs-dgap\endcsname{0.35}
\expandafter\def\csname N@cg-qwen-pool-abl-abs-dgap-ci\endcsname{$[0.06, 0.61]$}
\expandafter\def\csname N@cg-qwen-pool-abl-abs-dcorr\endcsname{$-$0.05}
\expandafter\def\csname N@cg-qwen-pool-abl-abs-dcorr-ci\endcsname{$[-0.46, 0.32]$}
\expandafter\def\csname N@cg-qwen-pool-abl-abs-rel\endcsname{1.08}
\expandafter\def\csname N@cg-qwen-pool-abl-placebo-corr\endcsname{2.73}
\expandafter\def\csname N@cg-qwen-pool-abl-placebo-cross\endcsname{1.90}
\expandafter\def\csname N@cg-qwen-pool-abl-placebo-pwin\endcsname{100.0\%}
\expandafter\def\csname N@cg-qwen-pool-abl-placebo-gap\endcsname{1.67}
\expandafter\def\csname N@cg-qwen-pool-abl-placebo-dcross\endcsname{$-$0.72}
\expandafter\def\csname N@cg-qwen-pool-abl-placebo-dcross-ci\endcsname{$[-1.01, -0.46]$}
\expandafter\def\csname N@cg-qwen-pool-abl-placebo-dgap\endcsname{$-$0.56}
\expandafter\def\csname N@cg-qwen-pool-abl-placebo-dgap-ci\endcsname{$[-0.82, -0.30]$}
\expandafter\def\csname N@cg-qwen-pool-abl-placebo-dcorr\endcsname{$-$0.88}
\expandafter\def\csname N@cg-qwen-pool-abl-placebo-dcorr-ci\endcsname{$[-1.24, -0.53]$}
\expandafter\def\csname N@cg-qwen-pool-abl-placebo-rel\endcsname{0.89}
\expandafter\def\csname N@cg-qwen-pool-abl-seed2-corr\endcsname{1.89}
\expandafter\def\csname N@cg-qwen-pool-abl-seed2-cross\endcsname{1.22}
\expandafter\def\csname N@cg-qwen-pool-abl-seed2-pwin\endcsname{58.5\%}
\expandafter\def\csname N@cg-qwen-pool-abl-seed2-gap\endcsname{1.15}
\expandafter\def\csname N@cg-qwen-pool-abl-seed2-dcross\endcsname{$-$0.04}
\expandafter\def\csname N@cg-qwen-pool-abl-seed2-dcross-ci\endcsname{$[-0.31, 0.21]$}
\expandafter\def\csname N@cg-qwen-pool-abl-seed2-dgap\endcsname{$-$0.04}
\expandafter\def\csname N@cg-qwen-pool-abl-seed2-dgap-ci\endcsname{$[-0.27, 0.22]$}
\expandafter\def\csname N@cg-qwen-pool-abl-seed2-dcorr\endcsname{$-$0.04}
\expandafter\def\csname N@cg-qwen-pool-abl-seed2-dcorr-ci\endcsname{$[-0.33, 0.26]$}
\expandafter\def\csname N@cg-qwen-pool-abl-seed2-rel\endcsname{0.97}
\expandafter\def\csname N@cg-qwen-pool-abl-englow-corr\endcsname{1.72}
\expandafter\def\csname N@cg-qwen-pool-abl-englow-cross\endcsname{1.01}
\expandafter\def\csname N@cg-qwen-pool-abl-englow-pwin\endcsname{17.5\%}
\expandafter\def\csname N@cg-qwen-pool-abl-englow-gap\endcsname{1.15}
\expandafter\def\csname N@cg-qwen-pool-abl-englow-dcross\endcsname{0.17}
\expandafter\def\csname N@cg-qwen-pool-abl-englow-dcross-ci\endcsname{$[-0.08, 0.45]$}
\expandafter\def\csname N@cg-qwen-pool-abl-englow-dgap\endcsname{$-$0.04}
\expandafter\def\csname N@cg-qwen-pool-abl-englow-dgap-ci\endcsname{$[-0.31, 0.24]$}
\expandafter\def\csname N@cg-qwen-pool-abl-englow-dcorr\endcsname{0.14}
\expandafter\def\csname N@cg-qwen-pool-abl-englow-dcorr-ci\endcsname{$[-0.15, 0.44]$}
\expandafter\def\csname N@cg-qwen-pool-abl-englow-rel\endcsname{0.96}
\expandafter\def\csname N@cg-qwen-pool-abl-enghigh-corr\endcsname{1.91}
\expandafter\def\csname N@cg-qwen-pool-abl-enghigh-cross\endcsname{1.31}
\expandafter\def\csname N@cg-qwen-pool-abl-enghigh-pwin\endcsname{65.5\%}
\expandafter\def\csname N@cg-qwen-pool-abl-enghigh-gap\endcsname{1.29}
\expandafter\def\csname N@cg-qwen-pool-abl-enghigh-dcross\endcsname{$-$0.14}
\expandafter\def\csname N@cg-qwen-pool-abl-enghigh-dcross-ci\endcsname{$[-0.40, 0.11]$}
\expandafter\def\csname N@cg-qwen-pool-abl-enghigh-dgap\endcsname{$-$0.17}
\expandafter\def\csname N@cg-qwen-pool-abl-enghigh-dgap-ci\endcsname{$[-0.41, 0.10]$}
\expandafter\def\csname N@cg-qwen-pool-abl-enghigh-dcorr\endcsname{$-$0.06}
\expandafter\def\csname N@cg-qwen-pool-abl-enghigh-dcorr-ci\endcsname{$[-0.33, 0.27]$}
\expandafter\def\csname N@cg-qwen-pool-abl-enghigh-rel\endcsname{0.97}
\expandafter\def\csname N@cg-qwen-pool-abl-demsharp-corr\endcsname{3.03}
\expandafter\def\csname N@cg-qwen-pool-abl-demsharp-cross\endcsname{1.92}
\expandafter\def\csname N@cg-qwen-pool-abl-demsharp-pwin\endcsname{100.0\%}
\expandafter\def\csname N@cg-qwen-pool-abl-demsharp-gap\endcsname{1.75}
\expandafter\def\csname N@cg-qwen-pool-abl-demsharp-dcross\endcsname{$-$0.74}
\expandafter\def\csname N@cg-qwen-pool-abl-demsharp-dcross-ci\endcsname{$[-1.00, -0.49]$}
\expandafter\def\csname N@cg-qwen-pool-abl-demsharp-dgap\endcsname{$-$0.63}
\expandafter\def\csname N@cg-qwen-pool-abl-demsharp-dgap-ci\endcsname{$[-0.87, -0.41]$}
\expandafter\def\csname N@cg-qwen-pool-abl-demsharp-dcorr\endcsname{$-$1.18}
\expandafter\def\csname N@cg-qwen-pool-abl-demsharp-dcorr-ci\endcsname{$[-1.47, -0.83]$}
\expandafter\def\csname N@cg-qwen-pool-abl-demsharp-rel\endcsname{0.93}
\expandafter\def\csname N@cg-qwen-pool-demo-corr\endcsname{2.19}
\expandafter\def\csname N@cg-qwen-pool-demo-cross\endcsname{1.54}
\expandafter\def\csname N@cg-qwen-pool-demo-pwin\endcsname{98.0\%}
\expandafter\def\csname N@cg-qwen-pool-demo-gap\endcsname{1.60}
\expandafter\def\csname N@cg-qwen-pool-demo-dcross\endcsname{$-$0.37}
\expandafter\def\csname N@cg-qwen-pool-demo-dcross-ci\endcsname{$[-0.65, -0.15]$}
\expandafter\def\csname N@cg-qwen-pool-demo-dgap\endcsname{$-$0.48}
\expandafter\def\csname N@cg-qwen-pool-demo-dgap-ci\endcsname{$[-0.72, -0.22]$}
\expandafter\def\csname N@cg-qwen-pool-demo-dcorr\endcsname{$-$0.34}
\expandafter\def\csname N@cg-qwen-pool-demo-dcorr-ci\endcsname{$[-0.67, -0.03]$}
\expandafter\def\csname N@cg-qwen-pool-demo-rel\endcsname{0.62}
\expandafter\def\csname N@cg-gemma-otp-floor-corr\endcsname{1.10}
\expandafter\def\csname N@cg-gemma-otp-floor-cross\endcsname{1.07}
\expandafter\def\csname N@cg-gemma-otp-lpp-corr\endcsname{1.84}
\expandafter\def\csname N@cg-gemma-otp-lpp-rel\endcsname{1.24}
\expandafter\def\csname N@cg-gemma-otp-lpp-cross\endcsname{1.30}
\expandafter\def\csname N@cg-gemma-otp-lpp-within\endcsname{1.60}
\expandafter\def\csname N@cg-gemma-otp-lpp-gap\endcsname{0.99}
\expandafter\def\csname N@cg-gemma-otp-lpp-wpsd\endcsname{1.03}
\expandafter\def\csname N@cg-gemma-otp-lpp-edp\endcsname{0.021}
\expandafter\def\csname N@cg-gemma-otp-sel-corr\endcsname{1.84--2.66}
\expandafter\def\csname N@cg-gemma-otp-sel-rel\endcsname{0.61--0.98}
\expandafter\def\csname N@cg-gemma-otp-sel-n\endcsname{7}
\expandafter\def\csname N@cg-gemma-otp-sel-nworse\endcsname{7}
\expandafter\def\csname N@cg-gemma-otp-sel-divbest-corr\endcsname{2.15}
\expandafter\def\csname N@cg-gemma-otp-sel-divbest-rel\endcsname{0.90}
\expandafter\def\csname N@cg-gemma-otp-sel-divbest-cross\endcsname{1.72}
\expandafter\def\csname N@cg-gemma-otp-sel-divbest-d\endcsname{$-$0.31}
\expandafter\def\csname N@cg-gemma-otp-sel-divbest-ci\endcsname{$[-0.70, 0.09]$}
\expandafter\def\csname N@cg-gemma-otp-sel-divbest-pwin\endcsname{94\%}
\expandafter\def\csname N@cg-gemma-otp-sel-divbest-edp\endcsname{0.040}
\expandafter\def\csname N@cg-gemma-otp-sel-divbest-edpci\endcsname{$[-0.085, 0.051]$}
\expandafter\def\csname N@cg-gemma-otp-sel-rw-corr\endcsname{2.66}
\expandafter\def\csname N@cg-gemma-otp-sel-rw-rel\endcsname{0.98}
\expandafter\def\csname N@cg-gemma-otp-sel-rw-cross\endcsname{2.04}
\expandafter\def\csname N@cg-gemma-otp-sel-rw-d\endcsname{$-$0.82}
\expandafter\def\csname N@cg-gemma-otp-sel-rw-ci\endcsname{$[-1.33, -0.38]$}
\expandafter\def\csname N@cg-gemma-otp-sel-rw-pwin\endcsname{100\%}
\expandafter\def\csname N@cg-gemma-otp-sel-rw-edp\endcsname{0.076}
\expandafter\def\csname N@cg-gemma-otp-sel-rw-edpci\endcsname{$[-0.173, 0.040]$}
\expandafter\def\csname N@cg-gemma-otp-sel-base-corr\endcsname{2.51}
\expandafter\def\csname N@cg-gemma-otp-sel-base-rel\endcsname{0.90}
\expandafter\def\csname N@cg-gemma-otp-sel-base-cross\endcsname{1.92}
\expandafter\def\csname N@cg-gemma-otp-sel-base-d\endcsname{$-$0.67}
\expandafter\def\csname N@cg-gemma-otp-sel-base-ci\endcsname{$[-1.06, -0.28]$}
\expandafter\def\csname N@cg-gemma-otp-sel-base-pwin\endcsname{100\%}
\expandafter\def\csname N@cg-gemma-otp-sel-base-edp\endcsname{0.052}
\expandafter\def\csname N@cg-gemma-otp-sel-base-edpci\endcsname{$[-0.101, 0.035]$}
\expandafter\def\csname N@cg-gemma-otp-sel-div-corr\endcsname{2.20}
\expandafter\def\csname N@cg-gemma-otp-sel-div-rel\endcsname{0.90}
\expandafter\def\csname N@cg-gemma-otp-sel-div-cross\endcsname{1.72}
\expandafter\def\csname N@cg-gemma-otp-sel-div-d\endcsname{$-$0.36}
\expandafter\def\csname N@cg-gemma-otp-sel-div-ci\endcsname{$[-0.70, -0.01]$}
\expandafter\def\csname N@cg-gemma-otp-sel-div-pwin\endcsname{98\%}
\expandafter\def\csname N@cg-gemma-otp-sel-div-edp\endcsname{0.041}
\expandafter\def\csname N@cg-gemma-otp-sel-div-edpci\endcsname{$[-0.077, 0.050]$}
\expandafter\def\csname N@cg-gemma-otp-sel-argyle-corr\endcsname{2.14}
\expandafter\def\csname N@cg-gemma-otp-sel-argyle-rel\endcsname{0.85}
\expandafter\def\csname N@cg-gemma-otp-sel-argyle-cross\endcsname{1.68}
\expandafter\def\csname N@cg-gemma-otp-sel-argyle-d\endcsname{$-$0.30}
\expandafter\def\csname N@cg-gemma-otp-sel-argyle-ci\endcsname{$[-0.65, 0.07]$}
\expandafter\def\csname N@cg-gemma-otp-sel-argyle-pwin\endcsname{94\%}
\expandafter\def\csname N@cg-gemma-otp-sel-argyle-edp\endcsname{0.046}
\expandafter\def\csname N@cg-gemma-otp-sel-argyle-edpci\endcsname{$[-0.097, 0.046]$}
\expandafter\def\csname N@cg-gemma-otp-sel-story-corr\endcsname{2.62}
\expandafter\def\csname N@cg-gemma-otp-sel-story-rel\endcsname{0.98}
\expandafter\def\csname N@cg-gemma-otp-sel-story-cross\endcsname{1.95}
\expandafter\def\csname N@cg-gemma-otp-sel-story-d\endcsname{$-$0.78}
\expandafter\def\csname N@cg-gemma-otp-sel-story-ci\endcsname{$[-1.21, -0.37]$}
\expandafter\def\csname N@cg-gemma-otp-sel-story-pwin\endcsname{100\%}
\expandafter\def\csname N@cg-gemma-otp-sel-story-edp\endcsname{0.056}
\expandafter\def\csname N@cg-gemma-otp-sel-story-edpci\endcsname{$[-0.100, 0.036]$}
\expandafter\def\csname N@cg-gemma-otp-sel-anth-corr\endcsname{1.84}
\expandafter\def\csname N@cg-gemma-otp-sel-anth-rel\endcsname{0.61}
\expandafter\def\csname N@cg-gemma-otp-sel-anth-cross\endcsname{0.69}
\expandafter\def\csname N@cg-gemma-otp-sel-anth-d\endcsname{$-$0.00}
\expandafter\def\csname N@cg-gemma-otp-sel-anth-ci\endcsname{$[-0.47, 0.47]$}
\expandafter\def\csname N@cg-gemma-otp-sel-anth-pwin\endcsname{52\%}
\expandafter\def\csname N@cg-gemma-otp-sel-anth-edp\endcsname{0.029}
\expandafter\def\csname N@cg-gemma-otp-sel-anth-edpci\endcsname{$[-0.088, 0.062]$}
\expandafter\def\csname N@cg-gemma-otp-sel-drange\endcsname{$-$0.82 to $-$0.00}
\expandafter\def\csname N@cg-gemma-otp-sel-nsig\endcsname{4}
\expandafter\def\csname N@cg-gemma-otp-sel-relx\endcsname{1.3--2.0}
\expandafter\def\csname N@cg-gemma-otp-sel-pwinmin\endcsname{52\%}
\expandafter\def\csname N@cg-gemma-otp-sel-pwinphrase\endcsname{at least 52\% of}
\expandafter\def\csname N@cg-gemma-otp-shp-corr\endcsname{2.20--2.71}
\expandafter\def\csname N@cg-gemma-otp-shp-rel\endcsname{1.00--1.08}
\expandafter\def\csname N@cg-gemma-otp-shp-n\endcsname{4}
\expandafter\def\csname N@cg-gemma-otp-shp-nworse\endcsname{4}
\expandafter\def\csname N@cg-gemma-otp-shp-base-corr\endcsname{2.71}
\expandafter\def\csname N@cg-gemma-otp-shp-base-rel\endcsname{1.07}
\expandafter\def\csname N@cg-gemma-otp-shp-base-cross\endcsname{2.08}
\expandafter\def\csname N@cg-gemma-otp-shp-base-d\endcsname{$-$0.87}
\expandafter\def\csname N@cg-gemma-otp-shp-base-ci\endcsname{$[-1.26, -0.50]$}
\expandafter\def\csname N@cg-gemma-otp-shp-base-pwin\endcsname{100\%}
\expandafter\def\csname N@cg-gemma-otp-shp-base-edp\endcsname{0.065}
\expandafter\def\csname N@cg-gemma-otp-shp-base-edpci\endcsname{$[-0.116, 0.023]$}
\expandafter\def\csname N@cg-gemma-otp-shp-div-corr\endcsname{2.23}
\expandafter\def\csname N@cg-gemma-otp-shp-div-rel\endcsname{1.00}
\expandafter\def\csname N@cg-gemma-otp-shp-div-cross\endcsname{1.73}
\expandafter\def\csname N@cg-gemma-otp-shp-div-d\endcsname{$-$0.39}
\expandafter\def\csname N@cg-gemma-otp-shp-div-ci\endcsname{$[-0.75, 0.01]$}
\expandafter\def\csname N@cg-gemma-otp-shp-div-pwin\endcsname{96\%}
\expandafter\def\csname N@cg-gemma-otp-shp-div-edp\endcsname{0.043}
\expandafter\def\csname N@cg-gemma-otp-shp-div-edpci\endcsname{$[-0.087, 0.045]$}
\expandafter\def\csname N@cg-gemma-otp-shp-argyle-corr\endcsname{2.20}
\expandafter\def\csname N@cg-gemma-otp-shp-argyle-rel\endcsname{1.00}
\expandafter\def\csname N@cg-gemma-otp-shp-argyle-cross\endcsname{1.69}
\expandafter\def\csname N@cg-gemma-otp-shp-argyle-d\endcsname{$-$0.36}
\expandafter\def\csname N@cg-gemma-otp-shp-argyle-ci\endcsname{$[-0.69, 0.01]$}
\expandafter\def\csname N@cg-gemma-otp-shp-argyle-pwin\endcsname{97\%}
\expandafter\def\csname N@cg-gemma-otp-shp-argyle-edp\endcsname{0.046}
\expandafter\def\csname N@cg-gemma-otp-shp-argyle-edpci\endcsname{$[-0.086, 0.052]$}
\expandafter\def\csname N@cg-gemma-otp-shp-story-corr\endcsname{2.63}
\expandafter\def\csname N@cg-gemma-otp-shp-story-rel\endcsname{1.08}
\expandafter\def\csname N@cg-gemma-otp-shp-story-cross\endcsname{1.95}
\expandafter\def\csname N@cg-gemma-otp-shp-story-d\endcsname{$-$0.79}
\expandafter\def\csname N@cg-gemma-otp-shp-story-ci\endcsname{$[-1.15, -0.44]$}
\expandafter\def\csname N@cg-gemma-otp-shp-story-pwin\endcsname{100\%}
\expandafter\def\csname N@cg-gemma-otp-shp-story-edp\endcsname{0.056}
\expandafter\def\csname N@cg-gemma-otp-shp-story-edpci\endcsname{$[-0.107, 0.032]$}
\expandafter\def\csname N@cg-gemma-otp-shp-drange\endcsname{$-$0.87 to $-$0.36}
\expandafter\def\csname N@cg-gemma-otp-shp-nsig\endcsname{2}
\expandafter\def\csname N@cg-gemma-otp-shp-relx\endcsname{1.1--1.2}
\expandafter\def\csname N@cg-gemma-otp-shp-pwinmin\endcsname{96\%}
\expandafter\def\csname N@cg-gemma-otp-shp-pwinphrase\endcsname{at least 96\% of}
\expandafter\def\csname N@cg-gemma-otp-abl-k16-corr\endcsname{1.79}
\expandafter\def\csname N@cg-gemma-otp-abl-k16-cross\endcsname{1.01}
\expandafter\def\csname N@cg-gemma-otp-abl-k16-pwin\endcsname{40.0\%}
\expandafter\def\csname N@cg-gemma-otp-abl-k16-gap\endcsname{0.99}
\expandafter\def\csname N@cg-gemma-otp-abl-k16-dcross\endcsname{0.29}
\expandafter\def\csname N@cg-gemma-otp-abl-k16-dcross-ci\endcsname{$[0.02, 0.62]$}
\expandafter\def\csname N@cg-gemma-otp-abl-k16-dgap\endcsname{$-$0.01}
\expandafter\def\csname N@cg-gemma-otp-abl-k16-dgap-ci\endcsname{$[-0.32, 0.25]$}
\expandafter\def\csname N@cg-gemma-otp-abl-k16-dcorr\endcsname{0.05}
\expandafter\def\csname N@cg-gemma-otp-abl-k16-dcorr-ci\endcsname{$[-0.22, 0.41]$}
\expandafter\def\csname N@cg-gemma-otp-abl-k16-rel\endcsname{1.21}
\expandafter\def\csname N@cg-gemma-otp-abl-novals-corr\endcsname{1.74}
\expandafter\def\csname N@cg-gemma-otp-abl-novals-cross\endcsname{1.09}
\expandafter\def\csname N@cg-gemma-otp-abl-novals-pwin\endcsname{30.0\%}
\expandafter\def\csname N@cg-gemma-otp-abl-novals-gap\endcsname{0.99}
\expandafter\def\csname N@cg-gemma-otp-abl-novals-dcross\endcsname{0.21}
\expandafter\def\csname N@cg-gemma-otp-abl-novals-dcross-ci\endcsname{$[-0.06, 0.59]$}
\expandafter\def\csname N@cg-gemma-otp-abl-novals-dgap\endcsname{$-$0.00}
\expandafter\def\csname N@cg-gemma-otp-abl-novals-dgap-ci\endcsname{$[-0.31, 0.25]$}
\expandafter\def\csname N@cg-gemma-otp-abl-novals-dcorr\endcsname{0.10}
\expandafter\def\csname N@cg-gemma-otp-abl-novals-dcorr-ci\endcsname{$[-0.16, 0.45]$}
\expandafter\def\csname N@cg-gemma-otp-abl-novals-rel\endcsname{1.23}
\expandafter\def\csname N@cg-gemma-otp-abl-nodev-corr\endcsname{2.36}
\expandafter\def\csname N@cg-gemma-otp-abl-nodev-cross\endcsname{1.82}
\expandafter\def\csname N@cg-gemma-otp-abl-nodev-pwin\endcsname{99.0\%}
\expandafter\def\csname N@cg-gemma-otp-abl-nodev-gap\endcsname{1.00}
\expandafter\def\csname N@cg-gemma-otp-abl-nodev-dcross\endcsname{$-$0.52}
\expandafter\def\csname N@cg-gemma-otp-abl-nodev-dcross-ci\endcsname{$[-0.83, -0.22]$}
\expandafter\def\csname N@cg-gemma-otp-abl-nodev-dgap\endcsname{$-$0.01}
\expandafter\def\csname N@cg-gemma-otp-abl-nodev-dgap-ci\endcsname{$[-0.39, 0.28]$}
\expandafter\def\csname N@cg-gemma-otp-abl-nodev-dcorr\endcsname{$-$0.51}
\expandafter\def\csname N@cg-gemma-otp-abl-nodev-dcorr-ci\endcsname{$[-0.86, -0.13]$}
\expandafter\def\csname N@cg-gemma-otp-abl-nodev-rel\endcsname{1.09}
\expandafter\def\csname N@cg-gemma-otp-abl-nosharp-corr\endcsname{1.83}
\expandafter\def\csname N@cg-gemma-otp-abl-nosharp-cross\endcsname{1.29}
\expandafter\def\csname N@cg-gemma-otp-abl-nosharp-pwin\endcsname{45.0\%}
\expandafter\def\csname N@cg-gemma-otp-abl-nosharp-gap\endcsname{0.99}
\expandafter\def\csname N@cg-gemma-otp-abl-nosharp-dcross\endcsname{0.01}
\expandafter\def\csname N@cg-gemma-otp-abl-nosharp-dcross-ci\endcsname{$[-0.13, 0.15]$}
\expandafter\def\csname N@cg-gemma-otp-abl-nosharp-dgap\endcsname{$-$0.00}
\expandafter\def\csname N@cg-gemma-otp-abl-nosharp-dgap-ci\endcsname{$[-0.10, 0.10]$}
\expandafter\def\csname N@cg-gemma-otp-abl-nosharp-dcorr\endcsname{0.01}
\expandafter\def\csname N@cg-gemma-otp-abl-nosharp-dcorr-ci\endcsname{$[-0.15, 0.16]$}
\expandafter\def\csname N@cg-gemma-otp-abl-nosharp-rel\endcsname{1.18}
\expandafter\def\csname N@cg-gemma-otp-agg-corr\endcsname{1.97}
\expandafter\def\csname N@cg-gemma-otp-agg-cross\endcsname{0.49}
\expandafter\def\csname N@cg-gemma-otp-agg-pwin\endcsname{70.0\%}
\expandafter\def\csname N@cg-gemma-otp-agg-gap\endcsname{1.02}
\expandafter\def\csname N@cg-gemma-otp-agg-dcross\endcsname{0.81}
\expandafter\def\csname N@cg-gemma-otp-agg-dcross-ci\endcsname{$[0.52, 1.16]$}
\expandafter\def\csname N@cg-gemma-otp-agg-dgap\endcsname{$-$0.03}
\expandafter\def\csname N@cg-gemma-otp-agg-dgap-ci\endcsname{$[-0.36, 0.26]$}
\expandafter\def\csname N@cg-gemma-otp-agg-dcorr\endcsname{$-$0.13}
\expandafter\def\csname N@cg-gemma-otp-agg-dcorr-ci\endcsname{$[-0.55, 0.36]$}
\expandafter\def\csname N@cg-gemma-otp-agg-rel\endcsname{0.13}
\expandafter\def\csname N@cg-gemma-otp-abl-abs-corr\endcsname{1.85}
\expandafter\def\csname N@cg-gemma-otp-abl-abs-cross\endcsname{0.99}
\expandafter\def\csname N@cg-gemma-otp-abl-abs-pwin\endcsname{56.5\%}
\expandafter\def\csname N@cg-gemma-otp-abl-abs-gap\endcsname{0.99}
\expandafter\def\csname N@cg-gemma-otp-abl-abs-dcross\endcsname{0.31}
\expandafter\def\csname N@cg-gemma-otp-abl-abs-dcross-ci\endcsname{$[-0.00, 0.67]$}
\expandafter\def\csname N@cg-gemma-otp-abl-abs-dgap\endcsname{$-$0.00}
\expandafter\def\csname N@cg-gemma-otp-abl-abs-dgap-ci\endcsname{$[-0.35, 0.23]$}
\expandafter\def\csname N@cg-gemma-otp-abl-abs-dcorr\endcsname{$-$0.01}
\expandafter\def\csname N@cg-gemma-otp-abl-abs-dcorr-ci\endcsname{$[-0.28, 0.37]$}
\expandafter\def\csname N@cg-gemma-otp-abl-abs-rel\endcsname{1.30}
\expandafter\def\csname N@cg-gemma-otp-abl-placebo-corr\endcsname{2.51}
\expandafter\def\csname N@cg-gemma-otp-abl-placebo-cross\endcsname{1.93}
\expandafter\def\csname N@cg-gemma-otp-abl-placebo-pwin\endcsname{100.0\%}
\expandafter\def\csname N@cg-gemma-otp-abl-placebo-gap\endcsname{1.01}
\expandafter\def\csname N@cg-gemma-otp-abl-placebo-dcross\endcsname{$-$0.63}
\expandafter\def\csname N@cg-gemma-otp-abl-placebo-dcross-ci\endcsname{$[-0.95, -0.31]$}
\expandafter\def\csname N@cg-gemma-otp-abl-placebo-dgap\endcsname{$-$0.02}
\expandafter\def\csname N@cg-gemma-otp-abl-placebo-dgap-ci\endcsname{$[-0.30, 0.23]$}
\expandafter\def\csname N@cg-gemma-otp-abl-placebo-dcorr\endcsname{$-$0.67}
\expandafter\def\csname N@cg-gemma-otp-abl-placebo-dcorr-ci\endcsname{$[-1.05, -0.26]$}
\expandafter\def\csname N@cg-gemma-otp-abl-placebo-rel\endcsname{1.05}
\expandafter\def\csname N@cg-gemma-otp-abl-seed2-corr\endcsname{1.77}
\expandafter\def\csname N@cg-gemma-otp-abl-seed2-cross\endcsname{1.26}
\expandafter\def\csname N@cg-gemma-otp-abl-seed2-pwin\endcsname{32.0\%}
\expandafter\def\csname N@cg-gemma-otp-abl-seed2-gap\endcsname{1.00}
\expandafter\def\csname N@cg-gemma-otp-abl-seed2-dcross\endcsname{0.04}
\expandafter\def\csname N@cg-gemma-otp-abl-seed2-dcross-ci\endcsname{$[-0.28, 0.34]$}
\expandafter\def\csname N@cg-gemma-otp-abl-seed2-dgap\endcsname{$-$0.01}
\expandafter\def\csname N@cg-gemma-otp-abl-seed2-dgap-ci\endcsname{$[-0.30, 0.22]$}
\expandafter\def\csname N@cg-gemma-otp-abl-seed2-dcorr\endcsname{0.07}
\expandafter\def\csname N@cg-gemma-otp-abl-seed2-dcorr-ci\endcsname{$[-0.21, 0.39]$}
\expandafter\def\csname N@cg-gemma-otp-abl-seed2-rel\endcsname{1.24}
\expandafter\def\csname N@cg-gemma-otp-abl-englow-corr\endcsname{1.76}
\expandafter\def\csname N@cg-gemma-otp-abl-englow-cross\endcsname{0.98}
\expandafter\def\csname N@cg-gemma-otp-abl-englow-pwin\endcsname{34.5\%}
\expandafter\def\csname N@cg-gemma-otp-abl-englow-gap\endcsname{0.99}
\expandafter\def\csname N@cg-gemma-otp-abl-englow-dcross\endcsname{0.32}
\expandafter\def\csname N@cg-gemma-otp-abl-englow-dcross-ci\endcsname{$[0.03, 0.64]$}
\expandafter\def\csname N@cg-gemma-otp-abl-englow-dgap\endcsname{$-$0.00}
\expandafter\def\csname N@cg-gemma-otp-abl-englow-dgap-ci\endcsname{$[-0.29, 0.24]$}
\expandafter\def\csname N@cg-gemma-otp-abl-englow-dcorr\endcsname{0.08}
\expandafter\def\csname N@cg-gemma-otp-abl-englow-dcorr-ci\endcsname{$[-0.19, 0.44]$}
\expandafter\def\csname N@cg-gemma-otp-abl-englow-rel\endcsname{1.24}
\expandafter\def\csname N@cg-gemma-otp-abl-enghigh-corr\endcsname{1.77}
\expandafter\def\csname N@cg-gemma-otp-abl-enghigh-cross\endcsname{1.26}
\expandafter\def\csname N@cg-gemma-otp-abl-enghigh-pwin\endcsname{35.5\%}
\expandafter\def\csname N@cg-gemma-otp-abl-enghigh-gap\endcsname{0.99}
\expandafter\def\csname N@cg-gemma-otp-abl-enghigh-dcross\endcsname{0.04}
\expandafter\def\csname N@cg-gemma-otp-abl-enghigh-dcross-ci\endcsname{$[-0.29, 0.35]$}
\expandafter\def\csname N@cg-gemma-otp-abl-enghigh-dgap\endcsname{0.00}
\expandafter\def\csname N@cg-gemma-otp-abl-enghigh-dgap-ci\endcsname{$[-0.30, 0.24]$}
\expandafter\def\csname N@cg-gemma-otp-abl-enghigh-dcorr\endcsname{0.07}
\expandafter\def\csname N@cg-gemma-otp-abl-enghigh-dcorr-ci\endcsname{$[-0.19, 0.38]$}
\expandafter\def\csname N@cg-gemma-otp-abl-enghigh-rel\endcsname{1.22}
\expandafter\def\csname N@cg-gemma-otp-abl-demsharp-corr\endcsname{2.71}
\expandafter\def\csname N@cg-gemma-otp-abl-demsharp-cross\endcsname{2.08}
\expandafter\def\csname N@cg-gemma-otp-abl-demsharp-pwin\endcsname{100.0\%}
\expandafter\def\csname N@cg-gemma-otp-abl-demsharp-gap\endcsname{1.00}
\expandafter\def\csname N@cg-gemma-otp-abl-demsharp-dcross\endcsname{$-$0.78}
\expandafter\def\csname N@cg-gemma-otp-abl-demsharp-dcross-ci\endcsname{$[-1.16, -0.45]$}
\expandafter\def\csname N@cg-gemma-otp-abl-demsharp-dgap\endcsname{$-$0.02}
\expandafter\def\csname N@cg-gemma-otp-abl-demsharp-dgap-ci\endcsname{$[-0.29, 0.25]$}
\expandafter\def\csname N@cg-gemma-otp-abl-demsharp-dcorr\endcsname{$-$0.87}
\expandafter\def\csname N@cg-gemma-otp-abl-demsharp-dcorr-ci\endcsname{$[-1.24, -0.47]$}
\expandafter\def\csname N@cg-gemma-otp-abl-demsharp-rel\endcsname{1.06}
\expandafter\def\csname N@cg-gemma-otp-demo-corr\endcsname{2.64}
\expandafter\def\csname N@cg-gemma-otp-demo-cross\endcsname{2.03}
\expandafter\def\csname N@cg-gemma-otp-demo-pwin\endcsname{100.0\%}
\expandafter\def\csname N@cg-gemma-otp-demo-gap\endcsname{1.00}
\expandafter\def\csname N@cg-gemma-otp-demo-dcross\endcsname{$-$0.73}
\expandafter\def\csname N@cg-gemma-otp-demo-dcross-ci\endcsname{$[-1.10, -0.38]$}
\expandafter\def\csname N@cg-gemma-otp-demo-dgap\endcsname{$-$0.01}
\expandafter\def\csname N@cg-gemma-otp-demo-dgap-ci\endcsname{$[-0.33, 0.26]$}
\expandafter\def\csname N@cg-gemma-otp-demo-dcorr\endcsname{$-$0.80}
\expandafter\def\csname N@cg-gemma-otp-demo-dcorr-ci\endcsname{$[-1.20, -0.39]$}
\expandafter\def\csname N@cg-gemma-otp-demo-rel\endcsname{0.98}
\expandafter\def\csname N@cg-gemma-pool-floor-corr\endcsname{1.10}
\expandafter\def\csname N@cg-gemma-pool-floor-cross\endcsname{1.07}
\expandafter\def\csname N@cg-gemma-pool-lpp-corr\endcsname{2.05}
\expandafter\def\csname N@cg-gemma-pool-lpp-rel\endcsname{1.25}
\expandafter\def\csname N@cg-gemma-pool-lpp-cross\endcsname{1.44}
\expandafter\def\csname N@cg-gemma-pool-lpp-within\endcsname{1.67}
\expandafter\def\csname N@cg-gemma-pool-lpp-gap\endcsname{1.15}
\expandafter\def\csname N@cg-gemma-pool-lpp-wpsd\endcsname{1.01}
\expandafter\def\csname N@cg-gemma-pool-lpp-edp\endcsname{0.071}
\expandafter\def\csname N@cg-gemma-pool-sel-corr\endcsname{2.26--3.65}
\expandafter\def\csname N@cg-gemma-pool-sel-rel\endcsname{0.57--1.09}
\expandafter\def\csname N@cg-gemma-pool-sel-n\endcsname{7}
\expandafter\def\csname N@cg-gemma-pool-sel-nworse\endcsname{7}
\expandafter\def\csname N@cg-gemma-pool-sel-divbest-corr\endcsname{2.92}
\expandafter\def\csname N@cg-gemma-pool-sel-divbest-rel\endcsname{1.01}
\expandafter\def\csname N@cg-gemma-pool-sel-divbest-cross\endcsname{2.22}
\expandafter\def\csname N@cg-gemma-pool-sel-divbest-d\endcsname{$-$0.86}
\expandafter\def\csname N@cg-gemma-pool-sel-divbest-ci\endcsname{$[-1.25, -0.47]$}
\expandafter\def\csname N@cg-gemma-pool-sel-divbest-pwin\endcsname{100\%}
\expandafter\def\csname N@cg-gemma-pool-sel-divbest-edp\endcsname{0.312}
\expandafter\def\csname N@cg-gemma-pool-sel-divbest-edpci\endcsname{$[-0.398, -0.091]$}
\expandafter\def\csname N@cg-gemma-pool-sel-rw-corr\endcsname{3.42}
\expandafter\def\csname N@cg-gemma-pool-sel-rw-rel\endcsname{1.09}
\expandafter\def\csname N@cg-gemma-pool-sel-rw-cross\endcsname{2.43}
\expandafter\def\csname N@cg-gemma-pool-sel-rw-d\endcsname{$-$1.37}
\expandafter\def\csname N@cg-gemma-pool-sel-rw-ci\endcsname{$[-1.76, -0.93]$}
\expandafter\def\csname N@cg-gemma-pool-sel-rw-pwin\endcsname{100\%}
\expandafter\def\csname N@cg-gemma-pool-sel-rw-edp\endcsname{0.452}
\expandafter\def\csname N@cg-gemma-pool-sel-rw-edpci\endcsname{$[-0.658, -0.182]$}
\expandafter\def\csname N@cg-gemma-pool-sel-base-corr\endcsname{3.28}
\expandafter\def\csname N@cg-gemma-pool-sel-base-rel\endcsname{1.01}
\expandafter\def\csname N@cg-gemma-pool-sel-base-cross\endcsname{2.34}
\expandafter\def\csname N@cg-gemma-pool-sel-base-d\endcsname{$-$1.23}
\expandafter\def\csname N@cg-gemma-pool-sel-base-ci\endcsname{$[-1.58, -0.86]$}
\expandafter\def\csname N@cg-gemma-pool-sel-base-pwin\endcsname{100\%}
\expandafter\def\csname N@cg-gemma-pool-sel-base-edp\endcsname{0.392}
\expandafter\def\csname N@cg-gemma-pool-sel-base-edpci\endcsname{$[-0.497, -0.159]$}
\expandafter\def\csname N@cg-gemma-pool-sel-div-corr\endcsname{2.95}
\expandafter\def\csname N@cg-gemma-pool-sel-div-rel\endcsname{1.00}
\expandafter\def\csname N@cg-gemma-pool-sel-div-cross\endcsname{2.18}
\expandafter\def\csname N@cg-gemma-pool-sel-div-d\endcsname{$-$0.90}
\expandafter\def\csname N@cg-gemma-pool-sel-div-ci\endcsname{$[-1.24, -0.52]$}
\expandafter\def\csname N@cg-gemma-pool-sel-div-pwin\endcsname{100\%}
\expandafter\def\csname N@cg-gemma-pool-sel-div-edp\endcsname{0.316}
\expandafter\def\csname N@cg-gemma-pool-sel-div-edpci\endcsname{$[-0.398, -0.099]$}
\expandafter\def\csname N@cg-gemma-pool-sel-argyle-corr\endcsname{3.65}
\expandafter\def\csname N@cg-gemma-pool-sel-argyle-rel\endcsname{1.07}
\expandafter\def\csname N@cg-gemma-pool-sel-argyle-cross\endcsname{2.64}
\expandafter\def\csname N@cg-gemma-pool-sel-argyle-d\endcsname{$-$1.59}
\expandafter\def\csname N@cg-gemma-pool-sel-argyle-ci\endcsname{$[-1.95, -1.20]$}
\expandafter\def\csname N@cg-gemma-pool-sel-argyle-pwin\endcsname{100\%}
\expandafter\def\csname N@cg-gemma-pool-sel-argyle-edp\endcsname{0.721}
\expandafter\def\csname N@cg-gemma-pool-sel-argyle-edpci\endcsname{$[-0.837, -0.485]$}
\expandafter\def\csname N@cg-gemma-pool-sel-story-corr\endcsname{3.40}
\expandafter\def\csname N@cg-gemma-pool-sel-story-rel\endcsname{1.06}
\expandafter\def\csname N@cg-gemma-pool-sel-story-cross\endcsname{2.44}
\expandafter\def\csname N@cg-gemma-pool-sel-story-d\endcsname{$-$1.35}
\expandafter\def\csname N@cg-gemma-pool-sel-story-ci\endcsname{$[-1.76, -0.96]$}
\expandafter\def\csname N@cg-gemma-pool-sel-story-pwin\endcsname{100\%}
\expandafter\def\csname N@cg-gemma-pool-sel-story-edp\endcsname{0.413}
\expandafter\def\csname N@cg-gemma-pool-sel-story-edpci\endcsname{$[-0.522, -0.167]$}
\expandafter\def\csname N@cg-gemma-pool-sel-anth-corr\endcsname{2.26}
\expandafter\def\csname N@cg-gemma-pool-sel-anth-rel\endcsname{0.57}
\expandafter\def\csname N@cg-gemma-pool-sel-anth-cross\endcsname{0.62}
\expandafter\def\csname N@cg-gemma-pool-sel-anth-d\endcsname{$-$0.21}
\expandafter\def\csname N@cg-gemma-pool-sel-anth-ci\endcsname{$[-0.68, 0.32]$}
\expandafter\def\csname N@cg-gemma-pool-sel-anth-pwin\endcsname{76\%}
\expandafter\def\csname N@cg-gemma-pool-sel-anth-edp\endcsname{0.205}
\expandafter\def\csname N@cg-gemma-pool-sel-anth-edpci\endcsname{$[-0.311, 0.010]$}
\expandafter\def\csname N@cg-gemma-pool-sel-drange\endcsname{$-$1.59 to $-$0.21}
\expandafter\def\csname N@cg-gemma-pool-sel-nsig\endcsname{6}
\expandafter\def\csname N@cg-gemma-pool-sel-relx\endcsname{1.1--2.2}
\expandafter\def\csname N@cg-gemma-pool-sel-pwinmin\endcsname{76\%}
\expandafter\def\csname N@cg-gemma-pool-sel-pwinphrase\endcsname{at least 76\% of}
\expandafter\def\csname N@cg-gemma-pool-shp-corr\endcsname{2.98--3.73}
\expandafter\def\csname N@cg-gemma-pool-shp-rel\endcsname{1.10--1.18}
\expandafter\def\csname N@cg-gemma-pool-shp-n\endcsname{4}
\expandafter\def\csname N@cg-gemma-pool-shp-nworse\endcsname{4}
\expandafter\def\csname N@cg-gemma-pool-shp-base-corr\endcsname{3.49}
\expandafter\def\csname N@cg-gemma-pool-shp-base-rel\endcsname{1.18}
\expandafter\def\csname N@cg-gemma-pool-shp-base-cross\endcsname{2.48}
\expandafter\def\csname N@cg-gemma-pool-shp-base-d\endcsname{$-$1.44}
\expandafter\def\csname N@cg-gemma-pool-shp-base-ci\endcsname{$[-1.83, -1.07]$}
\expandafter\def\csname N@cg-gemma-pool-shp-base-pwin\endcsname{100\%}
\expandafter\def\csname N@cg-gemma-pool-shp-base-edp\endcsname{0.441}
\expandafter\def\csname N@cg-gemma-pool-shp-base-edpci\endcsname{$[-0.563, -0.199]$}
\expandafter\def\csname N@cg-gemma-pool-shp-div-corr\endcsname{2.98}
\expandafter\def\csname N@cg-gemma-pool-shp-div-rel\endcsname{1.10}
\expandafter\def\csname N@cg-gemma-pool-shp-div-cross\endcsname{2.20}
\expandafter\def\csname N@cg-gemma-pool-shp-div-d\endcsname{$-$0.93}
\expandafter\def\csname N@cg-gemma-pool-shp-div-ci\endcsname{$[-1.24, -0.57]$}
\expandafter\def\csname N@cg-gemma-pool-shp-div-pwin\endcsname{100\%}
\expandafter\def\csname N@cg-gemma-pool-shp-div-edp\endcsname{0.338}
\expandafter\def\csname N@cg-gemma-pool-shp-div-edpci\endcsname{$[-0.443, -0.128]$}
\expandafter\def\csname N@cg-gemma-pool-shp-argyle-corr\endcsname{3.73}
\expandafter\def\csname N@cg-gemma-pool-shp-argyle-rel\endcsname{1.18}
\expandafter\def\csname N@cg-gemma-pool-shp-argyle-cross\endcsname{2.70}
\expandafter\def\csname N@cg-gemma-pool-shp-argyle-d\endcsname{$-$1.68}
\expandafter\def\csname N@cg-gemma-pool-shp-argyle-ci\endcsname{$[-2.00, -1.31]$}
\expandafter\def\csname N@cg-gemma-pool-shp-argyle-pwin\endcsname{100\%}
\expandafter\def\csname N@cg-gemma-pool-shp-argyle-edp\endcsname{0.741}
\expandafter\def\csname N@cg-gemma-pool-shp-argyle-edpci\endcsname{$[-0.850, -0.505]$}
\expandafter\def\csname N@cg-gemma-pool-shp-story-corr\endcsname{3.44}
\expandafter\def\csname N@cg-gemma-pool-shp-story-rel\endcsname{1.15}
\expandafter\def\csname N@cg-gemma-pool-shp-story-cross\endcsname{2.44}
\expandafter\def\csname N@cg-gemma-pool-shp-story-d\endcsname{$-$1.39}
\expandafter\def\csname N@cg-gemma-pool-shp-story-ci\endcsname{$[-1.77, -1.04]$}
\expandafter\def\csname N@cg-gemma-pool-shp-story-pwin\endcsname{100\%}
\expandafter\def\csname N@cg-gemma-pool-shp-story-edp\endcsname{0.431}
\expandafter\def\csname N@cg-gemma-pool-shp-story-edpci\endcsname{$[-0.537, -0.201]$}
\expandafter\def\csname N@cg-gemma-pool-shp-drange\endcsname{$-$1.68 to $-$0.93}
\expandafter\def\csname N@cg-gemma-pool-shp-nsig\endcsname{4}
\expandafter\def\csname N@cg-gemma-pool-shp-relx\endcsname{1.1}
\expandafter\def\csname N@cg-gemma-pool-shp-pwinmin\endcsname{100\%}
\expandafter\def\csname N@cg-gemma-pool-shp-pwinphrase\endcsname{all}
\expandafter\def\csname N@cg-gemma-pool-abl-k16-corr\endcsname{2.00}
\expandafter\def\csname N@cg-gemma-pool-abl-k16-cross\endcsname{1.21}
\expandafter\def\csname N@cg-gemma-pool-abl-k16-pwin\endcsname{43.0\%}
\expandafter\def\csname N@cg-gemma-pool-abl-k16-gap\endcsname{1.27}
\expandafter\def\csname N@cg-gemma-pool-abl-k16-dcross\endcsname{0.23}
\expandafter\def\csname N@cg-gemma-pool-abl-k16-dcross-ci\endcsname{$[-0.00, 0.52]$}
\expandafter\def\csname N@cg-gemma-pool-abl-k16-dgap\endcsname{$-$0.11}
\expandafter\def\csname N@cg-gemma-pool-abl-k16-dgap-ci\endcsname{$[-0.39, 0.14]$}
\expandafter\def\csname N@cg-gemma-pool-abl-k16-dcorr\endcsname{0.05}
\expandafter\def\csname N@cg-gemma-pool-abl-k16-dcorr-ci\endcsname{$[-0.24, 0.41]$}
\expandafter\def\csname N@cg-gemma-pool-abl-k16-rel\endcsname{1.25}
\expandafter\def\csname N@cg-gemma-pool-abl-novals-corr\endcsname{1.91}
\expandafter\def\csname N@cg-gemma-pool-abl-novals-cross\endcsname{1.23}
\expandafter\def\csname N@cg-gemma-pool-abl-novals-pwin\endcsname{18.5\%}
\expandafter\def\csname N@cg-gemma-pool-abl-novals-gap\endcsname{1.21}
\expandafter\def\csname N@cg-gemma-pool-abl-novals-dcross\endcsname{0.20}
\expandafter\def\csname N@cg-gemma-pool-abl-novals-dcross-ci\endcsname{$[-0.03, 0.56]$}
\expandafter\def\csname N@cg-gemma-pool-abl-novals-dgap\endcsname{$-$0.06}
\expandafter\def\csname N@cg-gemma-pool-abl-novals-dgap-ci\endcsname{$[-0.37, 0.19]$}
\expandafter\def\csname N@cg-gemma-pool-abl-novals-dcorr\endcsname{0.14}
\expandafter\def\csname N@cg-gemma-pool-abl-novals-dcorr-ci\endcsname{$[-0.10, 0.48]$}
\expandafter\def\csname N@cg-gemma-pool-abl-novals-rel\endcsname{1.24}
\expandafter\def\csname N@cg-gemma-pool-abl-nodev-corr\endcsname{3.10}
\expandafter\def\csname N@cg-gemma-pool-abl-nodev-cross\endcsname{2.27}
\expandafter\def\csname N@cg-gemma-pool-abl-nodev-pwin\endcsname{100.0\%}
\expandafter\def\csname N@cg-gemma-pool-abl-nodev-gap\endcsname{1.83}
\expandafter\def\csname N@cg-gemma-pool-abl-nodev-dcross\endcsname{$-$0.84}
\expandafter\def\csname N@cg-gemma-pool-abl-nodev-dcross-ci\endcsname{$[-1.15, -0.55]$}
\expandafter\def\csname N@cg-gemma-pool-abl-nodev-dgap\endcsname{$-$0.67}
\expandafter\def\csname N@cg-gemma-pool-abl-nodev-dgap-ci\endcsname{$[-1.04, -0.38]$}
\expandafter\def\csname N@cg-gemma-pool-abl-nodev-dcorr\endcsname{$-$1.05}
\expandafter\def\csname N@cg-gemma-pool-abl-nodev-dcorr-ci\endcsname{$[-1.37, -0.65]$}
\expandafter\def\csname N@cg-gemma-pool-abl-nodev-rel\endcsname{1.15}
\expandafter\def\csname N@cg-gemma-pool-abl-nosharp-corr\endcsname{2.03}
\expandafter\def\csname N@cg-gemma-pool-abl-nosharp-cross\endcsname{1.42}
\expandafter\def\csname N@cg-gemma-pool-abl-nosharp-pwin\endcsname{43.5\%}
\expandafter\def\csname N@cg-gemma-pool-abl-nosharp-gap\endcsname{1.15}
\expandafter\def\csname N@cg-gemma-pool-abl-nosharp-dcross\endcsname{0.01}
\expandafter\def\csname N@cg-gemma-pool-abl-nosharp-dcross-ci\endcsname{$[-0.13, 0.13]$}
\expandafter\def\csname N@cg-gemma-pool-abl-nosharp-dgap\endcsname{0.00}
\expandafter\def\csname N@cg-gemma-pool-abl-nosharp-dgap-ci\endcsname{$[-0.11, 0.11]$}
\expandafter\def\csname N@cg-gemma-pool-abl-nosharp-dcorr\endcsname{0.02}
\expandafter\def\csname N@cg-gemma-pool-abl-nosharp-dcorr-ci\endcsname{$[-0.15, 0.17]$}
\expandafter\def\csname N@cg-gemma-pool-abl-nosharp-rel\endcsname{1.20}
\expandafter\def\csname N@cg-gemma-pool-agg-corr\endcsname{2.57}
\expandafter\def\csname N@cg-gemma-pool-agg-cross\endcsname{0.37}
\expandafter\def\csname N@cg-gemma-pool-agg-pwin\endcsname{98.5\%}
\expandafter\def\csname N@cg-gemma-pool-agg-gap\endcsname{0.26}
\expandafter\def\csname N@cg-gemma-pool-agg-dcross\endcsname{1.07}
\expandafter\def\csname N@cg-gemma-pool-agg-dcross-ci\endcsname{$[0.81, 1.37]$}
\expandafter\def\csname N@cg-gemma-pool-agg-dgap\endcsname{0.89}
\expandafter\def\csname N@cg-gemma-pool-agg-dgap-ci\endcsname{$[0.59, 1.28]$}
\expandafter\def\csname N@cg-gemma-pool-agg-dcorr\endcsname{$-$0.52}
\expandafter\def\csname N@cg-gemma-pool-agg-dcorr-ci\endcsname{$[-0.96, -0.04]$}
\expandafter\def\csname N@cg-gemma-pool-agg-rel\endcsname{0.00}
\expandafter\def\csname N@cg-gemma-pool-abl-abs-corr\endcsname{2.01}
\expandafter\def\csname N@cg-gemma-pool-abl-abs-cross\endcsname{1.06}
\expandafter\def\csname N@cg-gemma-pool-abl-abs-pwin\endcsname{45.5\%}
\expandafter\def\csname N@cg-gemma-pool-abl-abs-gap\endcsname{1.02}
\expandafter\def\csname N@cg-gemma-pool-abl-abs-dcross\endcsname{0.38}
\expandafter\def\csname N@cg-gemma-pool-abl-abs-dcross-ci\endcsname{$[0.10, 0.71]$}
\expandafter\def\csname N@cg-gemma-pool-abl-abs-dgap\endcsname{0.13}
\expandafter\def\csname N@cg-gemma-pool-abl-abs-dgap-ci\endcsname{$[-0.23, 0.36]$}
\expandafter\def\csname N@cg-gemma-pool-abl-abs-dcorr\endcsname{0.04}
\expandafter\def\csname N@cg-gemma-pool-abl-abs-dcorr-ci\endcsname{$[-0.23, 0.40]$}
\expandafter\def\csname N@cg-gemma-pool-abl-abs-rel\endcsname{1.32}
\expandafter\def\csname N@cg-gemma-pool-abl-placebo-corr\endcsname{3.43}
\expandafter\def\csname N@cg-gemma-pool-abl-placebo-cross\endcsname{2.45}
\expandafter\def\csname N@cg-gemma-pool-abl-placebo-pwin\endcsname{100.0\%}
\expandafter\def\csname N@cg-gemma-pool-abl-placebo-gap\endcsname{1.97}
\expandafter\def\csname N@cg-gemma-pool-abl-placebo-dcross\endcsname{$-$1.01}
\expandafter\def\csname N@cg-gemma-pool-abl-placebo-dcross-ci\endcsname{$[-1.36, -0.69]$}
\expandafter\def\csname N@cg-gemma-pool-abl-placebo-dgap\endcsname{$-$0.82}
\expandafter\def\csname N@cg-gemma-pool-abl-placebo-dgap-ci\endcsname{$[-1.16, -0.56]$}
\expandafter\def\csname N@cg-gemma-pool-abl-placebo-dcorr\endcsname{$-$1.37}
\expandafter\def\csname N@cg-gemma-pool-abl-placebo-dcorr-ci\endcsname{$[-1.70, -0.96]$}
\expandafter\def\csname N@cg-gemma-pool-abl-placebo-rel\endcsname{1.14}
\expandafter\def\csname N@cg-gemma-pool-abl-seed2-corr\endcsname{1.98}
\expandafter\def\csname N@cg-gemma-pool-abl-seed2-cross\endcsname{1.40}
\expandafter\def\csname N@cg-gemma-pool-abl-seed2-pwin\endcsname{35.5\%}
\expandafter\def\csname N@cg-gemma-pool-abl-seed2-gap\endcsname{1.19}
\expandafter\def\csname N@cg-gemma-pool-abl-seed2-dcross\endcsname{0.04}
\expandafter\def\csname N@cg-gemma-pool-abl-seed2-dcross-ci\endcsname{$[-0.25, 0.35]$}
\expandafter\def\csname N@cg-gemma-pool-abl-seed2-dgap\endcsname{$-$0.04}
\expandafter\def\csname N@cg-gemma-pool-abl-seed2-dgap-ci\endcsname{$[-0.35, 0.23]$}
\expandafter\def\csname N@cg-gemma-pool-abl-seed2-dcorr\endcsname{0.07}
\expandafter\def\csname N@cg-gemma-pool-abl-seed2-dcorr-ci\endcsname{$[-0.22, 0.43]$}
\expandafter\def\csname N@cg-gemma-pool-abl-seed2-rel\endcsname{1.26}
\expandafter\def\csname N@cg-gemma-pool-abl-englow-corr\endcsname{1.94}
\expandafter\def\csname N@cg-gemma-pool-abl-englow-cross\endcsname{1.16}
\expandafter\def\csname N@cg-gemma-pool-abl-englow-pwin\endcsname{27.0\%}
\expandafter\def\csname N@cg-gemma-pool-abl-englow-gap\endcsname{1.22}
\expandafter\def\csname N@cg-gemma-pool-abl-englow-dcross\endcsname{0.28}
\expandafter\def\csname N@cg-gemma-pool-abl-englow-dcross-ci\endcsname{$[0.00, 0.61]$}
\expandafter\def\csname N@cg-gemma-pool-abl-englow-dgap\endcsname{$-$0.07}
\expandafter\def\csname N@cg-gemma-pool-abl-englow-dgap-ci\endcsname{$[-0.40, 0.16]$}
\expandafter\def\csname N@cg-gemma-pool-abl-englow-dcorr\endcsname{0.11}
\expandafter\def\csname N@cg-gemma-pool-abl-englow-dcorr-ci\endcsname{$[-0.14, 0.43]$}
\expandafter\def\csname N@cg-gemma-pool-abl-englow-rel\endcsname{1.25}
\expandafter\def\csname N@cg-gemma-pool-abl-enghigh-corr\endcsname{2.13}
\expandafter\def\csname N@cg-gemma-pool-abl-enghigh-cross\endcsname{1.54}
\expandafter\def\csname N@cg-gemma-pool-abl-enghigh-pwin\endcsname{70.0\%}
\expandafter\def\csname N@cg-gemma-pool-abl-enghigh-gap\endcsname{1.42}
\expandafter\def\csname N@cg-gemma-pool-abl-enghigh-dcross\endcsname{$-$0.10}
\expandafter\def\csname N@cg-gemma-pool-abl-enghigh-dcross-ci\endcsname{$[-0.36, 0.19]$}
\expandafter\def\csname N@cg-gemma-pool-abl-enghigh-dgap\endcsname{$-$0.26}
\expandafter\def\csname N@cg-gemma-pool-abl-enghigh-dgap-ci\endcsname{$[-0.59, -0.03]$}
\expandafter\def\csname N@cg-gemma-pool-abl-enghigh-dcorr\endcsname{$-$0.07}
\expandafter\def\csname N@cg-gemma-pool-abl-enghigh-dcorr-ci\endcsname{$[-0.34, 0.23]$}
\expandafter\def\csname N@cg-gemma-pool-abl-enghigh-rel\endcsname{1.26}
\expandafter\def\csname N@cg-gemma-pool-abl-demsharp-corr\endcsname{3.48}
\expandafter\def\csname N@cg-gemma-pool-abl-demsharp-cross\endcsname{2.47}
\expandafter\def\csname N@cg-gemma-pool-abl-demsharp-pwin\endcsname{100.0\%}
\expandafter\def\csname N@cg-gemma-pool-abl-demsharp-gap\endcsname{1.95}
\expandafter\def\csname N@cg-gemma-pool-abl-demsharp-dcross\endcsname{$-$1.03}
\expandafter\def\csname N@cg-gemma-pool-abl-demsharp-dcross-ci\endcsname{$[-1.41, -0.69]$}
\expandafter\def\csname N@cg-gemma-pool-abl-demsharp-dgap\endcsname{$-$0.80}
\expandafter\def\csname N@cg-gemma-pool-abl-demsharp-dgap-ci\endcsname{$[-1.13, -0.53]$}
\expandafter\def\csname N@cg-gemma-pool-abl-demsharp-dcorr\endcsname{$-$1.43}
\expandafter\def\csname N@cg-gemma-pool-abl-demsharp-dcorr-ci\endcsname{$[-1.80, -1.06]$}
\expandafter\def\csname N@cg-gemma-pool-abl-demsharp-rel\endcsname{1.17}
\expandafter\def\csname N@cg-gemma-pool-demo-corr\endcsname{3.42}
\expandafter\def\csname N@cg-gemma-pool-demo-cross\endcsname{2.43}
\expandafter\def\csname N@cg-gemma-pool-demo-pwin\endcsname{100.0\%}
\expandafter\def\csname N@cg-gemma-pool-demo-gap\endcsname{1.94}
\expandafter\def\csname N@cg-gemma-pool-demo-dcross\endcsname{$-$0.99}
\expandafter\def\csname N@cg-gemma-pool-demo-dcross-ci\endcsname{$[-1.37, -0.68]$}
\expandafter\def\csname N@cg-gemma-pool-demo-dgap\endcsname{$-$0.79}
\expandafter\def\csname N@cg-gemma-pool-demo-dgap-ci\endcsname{$[-1.12, -0.50]$}
\expandafter\def\csname N@cg-gemma-pool-demo-dcorr\endcsname{$-$1.37}
\expandafter\def\csname N@cg-gemma-pool-demo-dcorr-ci\endcsname{$[-1.75, -0.99]$}
\expandafter\def\csname N@cg-gemma-pool-demo-rel\endcsname{1.09}
\expandafter\def\csname N@cg-gpt-otp-floor-corr\endcsname{1.11}
\expandafter\def\csname N@cg-gpt-otp-floor-cross\endcsname{1.06}
\expandafter\def\csname N@cg-gpt-otp-lpp-corr\endcsname{2.22}
\expandafter\def\csname N@cg-gpt-otp-lpp-rel\endcsname{1.25}
\expandafter\def\csname N@cg-gpt-otp-lpp-cross\endcsname{0.74}
\expandafter\def\csname N@cg-gpt-otp-lpp-within\endcsname{2.05}
\expandafter\def\csname N@cg-gpt-otp-lpp-gap\endcsname{0.99}
\expandafter\def\csname N@cg-gpt-otp-lpp-wpsd\endcsname{1.03}
\expandafter\def\csname N@cg-gpt-otp-lpp-edp\endcsname{0.028}
\expandafter\def\csname N@cg-gpt-otp-sel-corr\endcsname{1.91--2.77}
\expandafter\def\csname N@cg-gpt-otp-sel-rel\endcsname{0.84--0.99}
\expandafter\def\csname N@cg-gpt-otp-sel-n\endcsname{3}
\expandafter\def\csname N@cg-gpt-otp-sel-nworse\endcsname{2}
\expandafter\def\csname N@cg-gpt-otp-sel-rw-corr\endcsname{2.77}
\expandafter\def\csname N@cg-gpt-otp-sel-rw-rel\endcsname{0.99}
\expandafter\def\csname N@cg-gpt-otp-sel-rw-cross\endcsname{2.22}
\expandafter\def\csname N@cg-gpt-otp-sel-rw-d\endcsname{$-$0.55}
\expandafter\def\csname N@cg-gpt-otp-sel-rw-ci\endcsname{$[-1.12, -0.06]$}
\expandafter\def\csname N@cg-gpt-otp-sel-rw-pwin\endcsname{98\%}
\expandafter\def\csname N@cg-gpt-otp-sel-rw-edp\endcsname{0.069}
\expandafter\def\csname N@cg-gpt-otp-sel-rw-edpci\endcsname{$[-0.135, 0.042]$}
\expandafter\def\csname N@cg-gpt-otp-sel-base-corr\endcsname{2.77}
\expandafter\def\csname N@cg-gpt-otp-sel-base-rel\endcsname{0.99}
\expandafter\def\csname N@cg-gpt-otp-sel-base-cross\endcsname{2.22}
\expandafter\def\csname N@cg-gpt-otp-sel-base-d\endcsname{$-$0.55}
\expandafter\def\csname N@cg-gpt-otp-sel-base-ci\endcsname{$[-0.94, -0.08]$}
\expandafter\def\csname N@cg-gpt-otp-sel-base-pwin\endcsname{99\%}
\expandafter\def\csname N@cg-gpt-otp-sel-base-edp\endcsname{0.039}
\expandafter\def\csname N@cg-gpt-otp-sel-base-edpci\endcsname{$[-0.072, 0.056]$}
\expandafter\def\csname N@cg-gpt-otp-sel-div-corr\endcsname{1.91}
\expandafter\def\csname N@cg-gpt-otp-sel-div-rel\endcsname{0.84}
\expandafter\def\csname N@cg-gpt-otp-sel-div-cross\endcsname{1.50}
\expandafter\def\csname N@cg-gpt-otp-sel-div-d\endcsname{0.31}
\expandafter\def\csname N@cg-gpt-otp-sel-div-ci\endcsname{$[0.01, 0.71]$}
\expandafter\def\csname N@cg-gpt-otp-sel-div-pwin\endcsname{2\%}
\expandafter\def\csname N@cg-gpt-otp-sel-div-edp\endcsname{0.019}
\expandafter\def\csname N@cg-gpt-otp-sel-div-edpci\endcsname{$[-0.064, 0.077]$}
\expandafter\def\csname N@cg-gpt-otp-sel-drange\endcsname{$-$0.55 to +0.31}
\expandafter\def\csname N@cg-gpt-otp-sel-nsig\endcsname{3}
\expandafter\def\csname N@cg-gpt-otp-sel-relx\endcsname{1.3--1.5}
\expandafter\def\csname N@cg-gpt-otp-sel-pwinmin\endcsname{2\%}
\expandafter\def\csname N@cg-gpt-otp-sel-pwinphrase\endcsname{at least 2\% of}
\expandafter\def\csname N@cg-gpt-otp-shp-corr\endcsname{1.90--2.77}
\expandafter\def\csname N@cg-gpt-otp-shp-rel\endcsname{0.84--0.99}
\expandafter\def\csname N@cg-gpt-otp-shp-n\endcsname{2}
\expandafter\def\csname N@cg-gpt-otp-shp-nworse\endcsname{1}
\expandafter\def\csname N@cg-gpt-otp-shp-base-corr\endcsname{2.77}
\expandafter\def\csname N@cg-gpt-otp-shp-base-rel\endcsname{0.99}
\expandafter\def\csname N@cg-gpt-otp-shp-base-cross\endcsname{2.22}
\expandafter\def\csname N@cg-gpt-otp-shp-base-d\endcsname{$-$0.55}
\expandafter\def\csname N@cg-gpt-otp-shp-base-ci\endcsname{$[-0.95, -0.04]$}
\expandafter\def\csname N@cg-gpt-otp-shp-base-pwin\endcsname{98\%}
\expandafter\def\csname N@cg-gpt-otp-shp-base-edp\endcsname{0.039}
\expandafter\def\csname N@cg-gpt-otp-shp-base-edpci\endcsname{$[-0.072, 0.054]$}
\expandafter\def\csname N@cg-gpt-otp-shp-div-corr\endcsname{1.90}
\expandafter\def\csname N@cg-gpt-otp-shp-div-rel\endcsname{0.84}
\expandafter\def\csname N@cg-gpt-otp-shp-div-cross\endcsname{1.48}
\expandafter\def\csname N@cg-gpt-otp-shp-div-d\endcsname{0.32}
\expandafter\def\csname N@cg-gpt-otp-shp-div-ci\endcsname{$[-0.04, 0.69]$}
\expandafter\def\csname N@cg-gpt-otp-shp-div-pwin\endcsname{4\%}
\expandafter\def\csname N@cg-gpt-otp-shp-div-edp\endcsname{0.018}
\expandafter\def\csname N@cg-gpt-otp-shp-div-edpci\endcsname{$[-0.055, 0.083]$}
\expandafter\def\csname N@cg-gpt-otp-shp-drange\endcsname{$-$0.55 to +0.32}
\expandafter\def\csname N@cg-gpt-otp-shp-nsig\endcsname{1}
\expandafter\def\csname N@cg-gpt-otp-shp-relx\endcsname{1.3--1.5}
\expandafter\def\csname N@cg-gpt-otp-shp-pwinmin\endcsname{4\%}
\expandafter\def\csname N@cg-gpt-otp-shp-pwinphrase\endcsname{at least 4\% of}
\expandafter\def\csname N@cg-gpt-otp-abl-k16-corr\endcsname{2.06}
\expandafter\def\csname N@cg-gpt-otp-abl-k16-cross\endcsname{1.24}
\expandafter\def\csname N@cg-gpt-otp-abl-k16-pwin\endcsname{13.5\%}
\expandafter\def\csname N@cg-gpt-otp-abl-k16-gap\endcsname{0.98}
\expandafter\def\csname N@cg-gpt-otp-abl-k16-dcross\endcsname{$-$0.50}
\expandafter\def\csname N@cg-gpt-otp-abl-k16-dcross-ci\endcsname{$[-0.76, -0.26]$}
\expandafter\def\csname N@cg-gpt-otp-abl-k16-dgap\endcsname{0.01}
\expandafter\def\csname N@cg-gpt-otp-abl-k16-dgap-ci\endcsname{$[-0.18, 0.21]$}
\expandafter\def\csname N@cg-gpt-otp-abl-k16-dcorr\endcsname{0.16}
\expandafter\def\csname N@cg-gpt-otp-abl-k16-dcorr-ci\endcsname{$[-0.13, 0.45]$}
\expandafter\def\csname N@cg-gpt-otp-abl-k16-rel\endcsname{1.22}
\expandafter\def\csname N@cg-gpt-otp-abl-nodev-corr\endcsname{2.68}
\expandafter\def\csname N@cg-gpt-otp-abl-nodev-cross\endcsname{2.18}
\expandafter\def\csname N@cg-gpt-otp-abl-nodev-pwin\endcsname{94.0\%}
\expandafter\def\csname N@cg-gpt-otp-abl-nodev-gap\endcsname{0.99}
\expandafter\def\csname N@cg-gpt-otp-abl-nodev-dcross\endcsname{$-$1.45}
\expandafter\def\csname N@cg-gpt-otp-abl-nodev-dcross-ci\endcsname{$[-1.85, -1.12]$}
\expandafter\def\csname N@cg-gpt-otp-abl-nodev-dgap\endcsname{0.01}
\expandafter\def\csname N@cg-gpt-otp-abl-nodev-dgap-ci\endcsname{$[-0.25, 0.29]$}
\expandafter\def\csname N@cg-gpt-otp-abl-nodev-dcorr\endcsname{$-$0.46}
\expandafter\def\csname N@cg-gpt-otp-abl-nodev-dcorr-ci\endcsname{$[-0.97, 0.14]$}
\expandafter\def\csname N@cg-gpt-otp-abl-nodev-rel\endcsname{1.03}
\expandafter\def\csname N@cg-gpt-otp-agg-corr\endcsname{1.99}
\expandafter\def\csname N@cg-gpt-otp-agg-cross\endcsname{0.48}
\expandafter\def\csname N@cg-gpt-otp-agg-pwin\endcsname{10.0\%}
\expandafter\def\csname N@cg-gpt-otp-agg-gap\endcsname{1.00}
\expandafter\def\csname N@cg-gpt-otp-agg-dcross\endcsname{0.25}
\expandafter\def\csname N@cg-gpt-otp-agg-dcross-ci\endcsname{$[0.07, 0.47]$}
\expandafter\def\csname N@cg-gpt-otp-agg-dgap\endcsname{$-$0.00}
\expandafter\def\csname N@cg-gpt-otp-agg-dgap-ci\endcsname{$[-0.27, 0.22]$}
\expandafter\def\csname N@cg-gpt-otp-agg-dcorr\endcsname{0.23}
\expandafter\def\csname N@cg-gpt-otp-agg-dcorr-ci\endcsname{$[-0.12, 0.58]$}
\expandafter\def\csname N@cg-gpt-otp-agg-rel\endcsname{0.13}
\expandafter\def\csname N@cg-gpt-otp-abl-demsharp-corr\endcsname{2.78}
\expandafter\def\csname N@cg-gpt-otp-abl-demsharp-cross\endcsname{2.22}
\expandafter\def\csname N@cg-gpt-otp-abl-demsharp-pwin\endcsname{97.0\%}
\expandafter\def\csname N@cg-gpt-otp-abl-demsharp-gap\endcsname{0.98}
\expandafter\def\csname N@cg-gpt-otp-abl-demsharp-dcross\endcsname{$-$1.49}
\expandafter\def\csname N@cg-gpt-otp-abl-demsharp-dcross-ci\endcsname{$[-1.93, -1.15]$}
\expandafter\def\csname N@cg-gpt-otp-abl-demsharp-dgap\endcsname{0.01}
\expandafter\def\csname N@cg-gpt-otp-abl-demsharp-dgap-ci\endcsname{$[-0.22, 0.29]$}
\expandafter\def\csname N@cg-gpt-otp-abl-demsharp-dcorr\endcsname{$-$0.56}
\expandafter\def\csname N@cg-gpt-otp-abl-demsharp-dcorr-ci\endcsname{$[-1.04, 0.02]$}
\expandafter\def\csname N@cg-gpt-otp-abl-demsharp-rel\endcsname{0.99}
\expandafter\def\csname N@cg-gpt-otp-demo-corr\endcsname{2.78}
\expandafter\def\csname N@cg-gpt-otp-demo-cross\endcsname{2.23}
\expandafter\def\csname N@cg-gpt-otp-demo-pwin\endcsname{98.5\%}
\expandafter\def\csname N@cg-gpt-otp-demo-gap\endcsname{0.98}
\expandafter\def\csname N@cg-gpt-otp-demo-dcross\endcsname{$-$1.49}
\expandafter\def\csname N@cg-gpt-otp-demo-dcross-ci\endcsname{$[-1.90, -1.13]$}
\expandafter\def\csname N@cg-gpt-otp-demo-dgap\endcsname{0.01}
\expandafter\def\csname N@cg-gpt-otp-demo-dgap-ci\endcsname{$[-0.25, 0.30]$}
\expandafter\def\csname N@cg-gpt-otp-demo-dcorr\endcsname{$-$0.56}
\expandafter\def\csname N@cg-gpt-otp-demo-dcorr-ci\endcsname{$[-0.97, -0.03]$}
\expandafter\def\csname N@cg-gpt-otp-demo-rel\endcsname{0.99}
\expandafter\def\csname N@cg-gpt-pool-floor-corr\endcsname{1.11}
\expandafter\def\csname N@cg-gpt-pool-floor-cross\endcsname{1.06}
\expandafter\def\csname N@cg-gpt-pool-lpp-corr\endcsname{2.47}
\expandafter\def\csname N@cg-gpt-pool-lpp-rel\endcsname{1.30}
\expandafter\def\csname N@cg-gpt-pool-lpp-cross\endcsname{1.40}
\expandafter\def\csname N@cg-gpt-pool-lpp-within\endcsname{2.23}
\expandafter\def\csname N@cg-gpt-pool-lpp-gap\endcsname{1.78}
\expandafter\def\csname N@cg-gpt-pool-lpp-wpsd\endcsname{0.98}
\expandafter\def\csname N@cg-gpt-pool-lpp-edp\endcsname{0.228}
\expandafter\def\csname N@cg-gpt-pool-sel-corr\endcsname{3.91--4.96}
\expandafter\def\csname N@cg-gpt-pool-sel-rel\endcsname{1.08--1.19}
\expandafter\def\csname N@cg-gpt-pool-sel-n\endcsname{3}
\expandafter\def\csname N@cg-gpt-pool-sel-nworse\endcsname{3}
\expandafter\def\csname N@cg-gpt-pool-sel-rw-corr\endcsname{4.95}
\expandafter\def\csname N@cg-gpt-pool-sel-rw-rel\endcsname{1.19}
\expandafter\def\csname N@cg-gpt-pool-sel-rw-cross\endcsname{3.74}
\expandafter\def\csname N@cg-gpt-pool-sel-rw-d\endcsname{$-$2.48}
\expandafter\def\csname N@cg-gpt-pool-sel-rw-ci\endcsname{$[-2.90, -2.02]$}
\expandafter\def\csname N@cg-gpt-pool-sel-rw-pwin\endcsname{100\%}
\expandafter\def\csname N@cg-gpt-pool-sel-rw-edp\endcsname{1.189}
\expandafter\def\csname N@cg-gpt-pool-sel-rw-edpci\endcsname{$[-1.301, -0.603]$}
\expandafter\def\csname N@cg-gpt-pool-sel-base-corr\endcsname{4.96}
\expandafter\def\csname N@cg-gpt-pool-sel-base-rel\endcsname{1.19}
\expandafter\def\csname N@cg-gpt-pool-sel-base-cross\endcsname{3.75}
\expandafter\def\csname N@cg-gpt-pool-sel-base-d\endcsname{$-$2.49}
\expandafter\def\csname N@cg-gpt-pool-sel-base-ci\endcsname{$[-2.77, -2.17]$}
\expandafter\def\csname N@cg-gpt-pool-sel-base-pwin\endcsname{100\%}
\expandafter\def\csname N@cg-gpt-pool-sel-base-edp\endcsname{1.157}
\expandafter\def\csname N@cg-gpt-pool-sel-base-edpci\endcsname{$[-1.112, -0.721]$}
\expandafter\def\csname N@cg-gpt-pool-sel-div-corr\endcsname{3.91}
\expandafter\def\csname N@cg-gpt-pool-sel-div-rel\endcsname{1.08}
\expandafter\def\csname N@cg-gpt-pool-sel-div-cross\endcsname{3.00}
\expandafter\def\csname N@cg-gpt-pool-sel-div-d\endcsname{$-$1.44}
\expandafter\def\csname N@cg-gpt-pool-sel-div-ci\endcsname{$[-1.77, -1.10]$}
\expandafter\def\csname N@cg-gpt-pool-sel-div-pwin\endcsname{100\%}
\expandafter\def\csname N@cg-gpt-pool-sel-div-edp\endcsname{0.919}
\expandafter\def\csname N@cg-gpt-pool-sel-div-edpci\endcsname{$[-0.864, -0.502]$}
\expandafter\def\csname N@cg-gpt-pool-sel-drange\endcsname{$-$2.49 to $-$1.44}
\expandafter\def\csname N@cg-gpt-pool-sel-nsig\endcsname{3}
\expandafter\def\csname N@cg-gpt-pool-sel-relx\endcsname{1.1--1.2}
\expandafter\def\csname N@cg-gpt-pool-sel-pwinmin\endcsname{100\%}
\expandafter\def\csname N@cg-gpt-pool-sel-pwinphrase\endcsname{all}
\expandafter\def\csname N@cg-gpt-pool-shp-corr\endcsname{3.90--4.96}
\expandafter\def\csname N@cg-gpt-pool-shp-rel\endcsname{1.08--1.19}
\expandafter\def\csname N@cg-gpt-pool-shp-n\endcsname{2}
\expandafter\def\csname N@cg-gpt-pool-shp-nworse\endcsname{2}
\expandafter\def\csname N@cg-gpt-pool-shp-base-corr\endcsname{4.96}
\expandafter\def\csname N@cg-gpt-pool-shp-base-rel\endcsname{1.19}
\expandafter\def\csname N@cg-gpt-pool-shp-base-cross\endcsname{3.74}
\expandafter\def\csname N@cg-gpt-pool-shp-base-d\endcsname{$-$2.49}
\expandafter\def\csname N@cg-gpt-pool-shp-base-ci\endcsname{$[-2.75, -2.12]$}
\expandafter\def\csname N@cg-gpt-pool-shp-base-pwin\endcsname{100\%}
\expandafter\def\csname N@cg-gpt-pool-shp-base-edp\endcsname{1.160}
\expandafter\def\csname N@cg-gpt-pool-shp-base-edpci\endcsname{$[-1.103, -0.711]$}
\expandafter\def\csname N@cg-gpt-pool-shp-div-corr\endcsname{3.90}
\expandafter\def\csname N@cg-gpt-pool-shp-div-rel\endcsname{1.08}
\expandafter\def\csname N@cg-gpt-pool-shp-div-cross\endcsname{3.00}
\expandafter\def\csname N@cg-gpt-pool-shp-div-d\endcsname{$-$1.43}
\expandafter\def\csname N@cg-gpt-pool-shp-div-ci\endcsname{$[-1.70, -1.13]$}
\expandafter\def\csname N@cg-gpt-pool-shp-div-pwin\endcsname{100\%}
\expandafter\def\csname N@cg-gpt-pool-shp-div-edp\endcsname{0.910}
\expandafter\def\csname N@cg-gpt-pool-shp-div-edpci\endcsname{$[-0.849, -0.491]$}
\expandafter\def\csname N@cg-gpt-pool-shp-drange\endcsname{$-$2.49 to $-$1.43}
\expandafter\def\csname N@cg-gpt-pool-shp-nsig\endcsname{2}
\expandafter\def\csname N@cg-gpt-pool-shp-relx\endcsname{1.1--1.2}
\expandafter\def\csname N@cg-gpt-pool-shp-pwinmin\endcsname{100\%}
\expandafter\def\csname N@cg-gpt-pool-shp-pwinphrase\endcsname{all}
\expandafter\def\csname N@cg-gpt-pool-abl-k16-corr\endcsname{3.45}
\expandafter\def\csname N@cg-gpt-pool-abl-k16-cross\endcsname{2.55}
\expandafter\def\csname N@cg-gpt-pool-abl-k16-pwin\endcsname{100.0\%}
\expandafter\def\csname N@cg-gpt-pool-abl-k16-gap\endcsname{2.31}
\expandafter\def\csname N@cg-gpt-pool-abl-k16-dcross\endcsname{$-$1.15}
\expandafter\def\csname N@cg-gpt-pool-abl-k16-dcross-ci\endcsname{$[-1.51, -0.89]$}
\expandafter\def\csname N@cg-gpt-pool-abl-k16-dgap\endcsname{$-$0.53}
\expandafter\def\csname N@cg-gpt-pool-abl-k16-dgap-ci\endcsname{$[-0.72, -0.36]$}
\expandafter\def\csname N@cg-gpt-pool-abl-k16-dcorr\endcsname{$-$0.98}
\expandafter\def\csname N@cg-gpt-pool-abl-k16-dcorr-ci\endcsname{$[-1.29, -0.68]$}
\expandafter\def\csname N@cg-gpt-pool-abl-k16-rel\endcsname{1.29}
\expandafter\def\csname N@cg-gpt-pool-abl-nodev-corr\endcsname{4.62}
\expandafter\def\csname N@cg-gpt-pool-abl-nodev-cross\endcsname{3.48}
\expandafter\def\csname N@cg-gpt-pool-abl-nodev-pwin\endcsname{100.0\%}
\expandafter\def\csname N@cg-gpt-pool-abl-nodev-gap\endcsname{2.52}
\expandafter\def\csname N@cg-gpt-pool-abl-nodev-dcross\endcsname{$-$2.08}
\expandafter\def\csname N@cg-gpt-pool-abl-nodev-dcross-ci\endcsname{$[-2.45, -1.74]$}
\expandafter\def\csname N@cg-gpt-pool-abl-nodev-dgap\endcsname{$-$0.74}
\expandafter\def\csname N@cg-gpt-pool-abl-nodev-dgap-ci\endcsname{$[-0.98, -0.47]$}
\expandafter\def\csname N@cg-gpt-pool-abl-nodev-dcorr\endcsname{$-$2.15}
\expandafter\def\csname N@cg-gpt-pool-abl-nodev-dcorr-ci\endcsname{$[-2.42, -1.86]$}
\expandafter\def\csname N@cg-gpt-pool-abl-nodev-rel\endcsname{1.20}
\expandafter\def\csname N@cg-gpt-pool-agg-corr\endcsname{2.57}
\expandafter\def\csname N@cg-gpt-pool-agg-cross\endcsname{0.37}
\expandafter\def\csname N@cg-gpt-pool-agg-pwin\endcsname{69.5\%}
\expandafter\def\csname N@cg-gpt-pool-agg-gap\endcsname{0.26}
\expandafter\def\csname N@cg-gpt-pool-agg-dcross\endcsname{1.03}
\expandafter\def\csname N@cg-gpt-pool-agg-dcross-ci\endcsname{$[0.74, 1.37]$}
\expandafter\def\csname N@cg-gpt-pool-agg-dgap\endcsname{1.52}
\expandafter\def\csname N@cg-gpt-pool-agg-dgap-ci\endcsname{$[1.20, 1.91]$}
\expandafter\def\csname N@cg-gpt-pool-agg-dcorr\endcsname{$-$0.10}
\expandafter\def\csname N@cg-gpt-pool-agg-dcorr-ci\endcsname{$[-0.55, 0.42]$}
\expandafter\def\csname N@cg-gpt-pool-agg-rel\endcsname{0.00}
\expandafter\def\csname N@cg-gpt-pool-abl-demsharp-corr\endcsname{4.96}
\expandafter\def\csname N@cg-gpt-pool-abl-demsharp-cross\endcsname{3.75}
\expandafter\def\csname N@cg-gpt-pool-abl-demsharp-pwin\endcsname{100.0\%}
\expandafter\def\csname N@cg-gpt-pool-abl-demsharp-gap\endcsname{2.72}
\expandafter\def\csname N@cg-gpt-pool-abl-demsharp-dcross\endcsname{$-$2.34}
\expandafter\def\csname N@cg-gpt-pool-abl-demsharp-dcross-ci\endcsname{$[-2.80, -1.95]$}
\expandafter\def\csname N@cg-gpt-pool-abl-demsharp-dgap\endcsname{$-$0.94}
\expandafter\def\csname N@cg-gpt-pool-abl-demsharp-dgap-ci\endcsname{$[-1.19, -0.66]$}
\expandafter\def\csname N@cg-gpt-pool-abl-demsharp-dcorr\endcsname{$-$2.49}
\expandafter\def\csname N@cg-gpt-pool-abl-demsharp-dcorr-ci\endcsname{$[-2.78, -2.16]$}
\expandafter\def\csname N@cg-gpt-pool-abl-demsharp-rel\endcsname{1.19}
\expandafter\def\csname N@cg-gpt-pool-demo-corr\endcsname{4.97}
\expandafter\def\csname N@cg-gpt-pool-demo-cross\endcsname{3.75}
\expandafter\def\csname N@cg-gpt-pool-demo-pwin\endcsname{100.0\%}
\expandafter\def\csname N@cg-gpt-pool-demo-gap\endcsname{2.73}
\expandafter\def\csname N@cg-gpt-pool-demo-dcross\endcsname{$-$2.35}
\expandafter\def\csname N@cg-gpt-pool-demo-dcross-ci\endcsname{$[-2.82, -1.98]$}
\expandafter\def\csname N@cg-gpt-pool-demo-dgap\endcsname{$-$0.94}
\expandafter\def\csname N@cg-gpt-pool-demo-dgap-ci\endcsname{$[-1.18, -0.68]$}
\expandafter\def\csname N@cg-gpt-pool-demo-dcorr\endcsname{$-$2.50}
\expandafter\def\csname N@cg-gpt-pool-demo-dcorr-ci\endcsname{$[-2.78, -2.15]$}
\expandafter\def\csname N@cg-gpt-pool-demo-rel\endcsname{1.19}
\expandafter\def\csname N@gss-qwen-otp-floor-corr\endcsname{0.72}
\expandafter\def\csname N@gss-qwen-otp-floor-cross\endcsname{1.08}
\expandafter\def\csname N@gss-qwen-otp-lpp-corr\endcsname{1.11}
\expandafter\def\csname N@gss-qwen-otp-lpp-rel\endcsname{0.90}
\expandafter\def\csname N@gss-qwen-otp-lpp-cross\endcsname{1.10}
\expandafter\def\csname N@gss-qwen-otp-lpp-within\endcsname{0.84}
\expandafter\def\csname N@gss-qwen-otp-lpp-gap\endcsname{1.02}
\expandafter\def\csname N@gss-qwen-otp-lpp-wpsd\endcsname{1.02}
\expandafter\def\csname N@gss-qwen-otp-lpp-edp\endcsname{0.013}
\expandafter\def\csname N@gss-qwen-otp-sel-corr\endcsname{1.12--1.24}
\expandafter\def\csname N@gss-qwen-otp-sel-rel\endcsname{0.41--0.58}
\expandafter\def\csname N@gss-qwen-otp-sel-n\endcsname{7}
\expandafter\def\csname N@gss-qwen-otp-sel-nworse\endcsname{7}
\expandafter\def\csname N@gss-qwen-otp-sel-divbest-corr\endcsname{1.15}
\expandafter\def\csname N@gss-qwen-otp-sel-divbest-rel\endcsname{0.51}
\expandafter\def\csname N@gss-qwen-otp-sel-divbest-cross\endcsname{1.17}
\expandafter\def\csname N@gss-qwen-otp-sel-divbest-d\endcsname{$-$0.04}
\expandafter\def\csname N@gss-qwen-otp-sel-divbest-ci\endcsname{$[-0.28, 0.15]$}
\expandafter\def\csname N@gss-qwen-otp-sel-divbest-pwin\endcsname{64\%}
\expandafter\def\csname N@gss-qwen-otp-sel-divbest-edp\endcsname{0.015}
\expandafter\def\csname N@gss-qwen-otp-sel-divbest-edpci\endcsname{$[-0.043, 0.047]$}
\expandafter\def\csname N@gss-qwen-otp-sel-rw-corr\endcsname{1.22}
\expandafter\def\csname N@gss-qwen-otp-sel-rw-rel\endcsname{0.58}
\expandafter\def\csname N@gss-qwen-otp-sel-rw-cross\endcsname{1.30}
\expandafter\def\csname N@gss-qwen-otp-sel-rw-d\endcsname{$-$0.11}
\expandafter\def\csname N@gss-qwen-otp-sel-rw-ci\endcsname{$[-0.32, 0.11]$}
\expandafter\def\csname N@gss-qwen-otp-sel-rw-pwin\endcsname{82\%}
\expandafter\def\csname N@gss-qwen-otp-sel-rw-edp\endcsname{0.025}
\expandafter\def\csname N@gss-qwen-otp-sel-rw-edpci\endcsname{$[-0.059, 0.038]$}
\expandafter\def\csname N@gss-qwen-otp-sel-base-corr\endcsname{1.12}
\expandafter\def\csname N@gss-qwen-otp-sel-base-rel\endcsname{0.41}
\expandafter\def\csname N@gss-qwen-otp-sel-base-cross\endcsname{1.04}
\expandafter\def\csname N@gss-qwen-otp-sel-base-d\endcsname{$-$0.01}
\expandafter\def\csname N@gss-qwen-otp-sel-base-ci\endcsname{$[-0.20, 0.17]$}
\expandafter\def\csname N@gss-qwen-otp-sel-base-pwin\endcsname{58\%}
\expandafter\def\csname N@gss-qwen-otp-sel-base-edp\endcsname{0.018}
\expandafter\def\csname N@gss-qwen-otp-sel-base-edpci\endcsname{$[-0.047, 0.037]$}
\expandafter\def\csname N@gss-qwen-otp-sel-div-corr\endcsname{1.19}
\expandafter\def\csname N@gss-qwen-otp-sel-div-rel\endcsname{0.54}
\expandafter\def\csname N@gss-qwen-otp-sel-div-cross\endcsname{1.24}
\expandafter\def\csname N@gss-qwen-otp-sel-div-d\endcsname{$-$0.08}
\expandafter\def\csname N@gss-qwen-otp-sel-div-ci\endcsname{$[-0.28, 0.13]$}
\expandafter\def\csname N@gss-qwen-otp-sel-div-pwin\endcsname{78\%}
\expandafter\def\csname N@gss-qwen-otp-sel-div-edp\endcsname{0.017}
\expandafter\def\csname N@gss-qwen-otp-sel-div-edpci\endcsname{$[-0.043, 0.037]$}
\expandafter\def\csname N@gss-qwen-otp-sel-argyle-corr\endcsname{1.12}
\expandafter\def\csname N@gss-qwen-otp-sel-argyle-rel\endcsname{0.42}
\expandafter\def\csname N@gss-qwen-otp-sel-argyle-cross\endcsname{1.00}
\expandafter\def\csname N@gss-qwen-otp-sel-argyle-d\endcsname{$-$0.01}
\expandafter\def\csname N@gss-qwen-otp-sel-argyle-ci\endcsname{$[-0.21, 0.23]$}
\expandafter\def\csname N@gss-qwen-otp-sel-argyle-pwin\endcsname{54\%}
\expandafter\def\csname N@gss-qwen-otp-sel-argyle-edp\endcsname{0.017}
\expandafter\def\csname N@gss-qwen-otp-sel-argyle-edpci\endcsname{$[-0.040, 0.040]$}
\expandafter\def\csname N@gss-qwen-otp-sel-story-corr\endcsname{1.19}
\expandafter\def\csname N@gss-qwen-otp-sel-story-rel\endcsname{0.52}
\expandafter\def\csname N@gss-qwen-otp-sel-story-cross\endcsname{1.22}
\expandafter\def\csname N@gss-qwen-otp-sel-story-d\endcsname{$-$0.08}
\expandafter\def\csname N@gss-qwen-otp-sel-story-ci\endcsname{$[-0.27, 0.11]$}
\expandafter\def\csname N@gss-qwen-otp-sel-story-pwin\endcsname{79\%}
\expandafter\def\csname N@gss-qwen-otp-sel-story-edp\endcsname{0.017}
\expandafter\def\csname N@gss-qwen-otp-sel-story-edpci\endcsname{$[-0.053, 0.037]$}
\expandafter\def\csname N@gss-qwen-otp-sel-anth-corr\endcsname{1.24}
\expandafter\def\csname N@gss-qwen-otp-sel-anth-rel\endcsname{0.53}
\expandafter\def\csname N@gss-qwen-otp-sel-anth-cross\endcsname{0.78}
\expandafter\def\csname N@gss-qwen-otp-sel-anth-d\endcsname{$-$0.12}
\expandafter\def\csname N@gss-qwen-otp-sel-anth-ci\endcsname{$[-0.34, 0.11]$}
\expandafter\def\csname N@gss-qwen-otp-sel-anth-pwin\endcsname{88\%}
\expandafter\def\csname N@gss-qwen-otp-sel-anth-edp\endcsname{0.020}
\expandafter\def\csname N@gss-qwen-otp-sel-anth-edpci\endcsname{$[-0.049, 0.043]$}
\expandafter\def\csname N@gss-qwen-otp-sel-drange\endcsname{$-$0.12 to $-$0.01}
\expandafter\def\csname N@gss-qwen-otp-sel-nsig\endcsname{0}
\expandafter\def\csname N@gss-qwen-otp-sel-relx\endcsname{1.5--2.2}
\expandafter\def\csname N@gss-qwen-otp-sel-pwinmin\endcsname{54\%}
\expandafter\def\csname N@gss-qwen-otp-sel-pwinphrase\endcsname{at least 54\% of}
\expandafter\def\csname N@gss-qwen-otp-shp-corr\endcsname{1.36--1.41}
\expandafter\def\csname N@gss-qwen-otp-shp-rel\endcsname{0.87--0.92}
\expandafter\def\csname N@gss-qwen-otp-shp-n\endcsname{4}
\expandafter\def\csname N@gss-qwen-otp-shp-nworse\endcsname{4}
\expandafter\def\csname N@gss-qwen-otp-shp-base-corr\endcsname{1.36}
\expandafter\def\csname N@gss-qwen-otp-shp-base-rel\endcsname{0.92}
\expandafter\def\csname N@gss-qwen-otp-shp-base-cross\endcsname{1.46}
\expandafter\def\csname N@gss-qwen-otp-shp-base-d\endcsname{$-$0.25}
\expandafter\def\csname N@gss-qwen-otp-shp-base-ci\endcsname{$[-0.50, -0.03]$}
\expandafter\def\csname N@gss-qwen-otp-shp-base-pwin\endcsname{99\%}
\expandafter\def\csname N@gss-qwen-otp-shp-base-edp\endcsname{0.024}
\expandafter\def\csname N@gss-qwen-otp-shp-base-edpci\endcsname{$[-0.052, 0.029]$}
\expandafter\def\csname N@gss-qwen-otp-shp-div-corr\endcsname{1.37}
\expandafter\def\csname N@gss-qwen-otp-shp-div-rel\endcsname{0.88}
\expandafter\def\csname N@gss-qwen-otp-shp-div-cross\endcsname{1.46}
\expandafter\def\csname N@gss-qwen-otp-shp-div-d\endcsname{$-$0.25}
\expandafter\def\csname N@gss-qwen-otp-shp-div-ci\endcsname{$[-0.43, -0.05]$}
\expandafter\def\csname N@gss-qwen-otp-shp-div-pwin\endcsname{100\%}
\expandafter\def\csname N@gss-qwen-otp-shp-div-edp\endcsname{0.022}
\expandafter\def\csname N@gss-qwen-otp-shp-div-edpci\endcsname{$[-0.052, 0.039]$}
\expandafter\def\csname N@gss-qwen-otp-shp-argyle-corr\endcsname{1.39}
\expandafter\def\csname N@gss-qwen-otp-shp-argyle-rel\endcsname{0.87}
\expandafter\def\csname N@gss-qwen-otp-shp-argyle-cross\endcsname{1.33}
\expandafter\def\csname N@gss-qwen-otp-shp-argyle-d\endcsname{$-$0.28}
\expandafter\def\csname N@gss-qwen-otp-shp-argyle-ci\endcsname{$[-0.49, -0.06]$}
\expandafter\def\csname N@gss-qwen-otp-shp-argyle-pwin\endcsname{98\%}
\expandafter\def\csname N@gss-qwen-otp-shp-argyle-edp\endcsname{0.030}
\expandafter\def\csname N@gss-qwen-otp-shp-argyle-edpci\endcsname{$[-0.060, 0.027]$}
\expandafter\def\csname N@gss-qwen-otp-shp-story-corr\endcsname{1.41}
\expandafter\def\csname N@gss-qwen-otp-shp-story-rel\endcsname{0.87}
\expandafter\def\csname N@gss-qwen-otp-shp-story-cross\endcsname{1.52}
\expandafter\def\csname N@gss-qwen-otp-shp-story-d\endcsname{$-$0.30}
\expandafter\def\csname N@gss-qwen-otp-shp-story-ci\endcsname{$[-0.54, -0.11]$}
\expandafter\def\csname N@gss-qwen-otp-shp-story-pwin\endcsname{100\%}
\expandafter\def\csname N@gss-qwen-otp-shp-story-edp\endcsname{0.022}
\expandafter\def\csname N@gss-qwen-otp-shp-story-edpci\endcsname{$[-0.053, 0.032]$}
\expandafter\def\csname N@gss-qwen-otp-shp-drange\endcsname{$-$0.30 to $-$0.25}
\expandafter\def\csname N@gss-qwen-otp-shp-nsig\endcsname{4}
\expandafter\def\csname N@gss-qwen-otp-shp-relx\endcsname{1.0}
\expandafter\def\csname N@gss-qwen-otp-shp-pwinmin\endcsname{98\%}
\expandafter\def\csname N@gss-qwen-otp-shp-pwinphrase\endcsname{at least 98\% of}
\expandafter\def\csname N@gss-qwen-otp-abl-k16-corr\endcsname{1.10}
\expandafter\def\csname N@gss-qwen-otp-abl-k16-cross\endcsname{0.99}
\expandafter\def\csname N@gss-qwen-otp-abl-k16-pwin\endcsname{42.0\%}
\expandafter\def\csname N@gss-qwen-otp-abl-k16-gap\endcsname{1.02}
\expandafter\def\csname N@gss-qwen-otp-abl-k16-dcross\endcsname{0.10}
\expandafter\def\csname N@gss-qwen-otp-abl-k16-dcross-ci\endcsname{$[-0.19, 0.39]$}
\expandafter\def\csname N@gss-qwen-otp-abl-k16-dgap\endcsname{0.00}
\expandafter\def\csname N@gss-qwen-otp-abl-k16-dgap-ci\endcsname{$[-0.17, 0.20]$}
\expandafter\def\csname N@gss-qwen-otp-abl-k16-dcorr\endcsname{0.02}
\expandafter\def\csname N@gss-qwen-otp-abl-k16-dcorr-ci\endcsname{$[-0.18, 0.21]$}
\expandafter\def\csname N@gss-qwen-otp-abl-k16-rel\endcsname{0.79}
\expandafter\def\csname N@gss-qwen-otp-abl-nosharp-corr\endcsname{1.03}
\expandafter\def\csname N@gss-qwen-otp-abl-nosharp-cross\endcsname{0.98}
\expandafter\def\csname N@gss-qwen-otp-abl-nosharp-pwin\endcsname{20.5\%}
\expandafter\def\csname N@gss-qwen-otp-abl-nosharp-gap\endcsname{1.02}
\expandafter\def\csname N@gss-qwen-otp-abl-nosharp-dcross\endcsname{0.11}
\expandafter\def\csname N@gss-qwen-otp-abl-nosharp-dcross-ci\endcsname{$[-0.17, 0.39]$}
\expandafter\def\csname N@gss-qwen-otp-abl-nosharp-dgap\endcsname{0.00}
\expandafter\def\csname N@gss-qwen-otp-abl-nosharp-dgap-ci\endcsname{$[-0.14, 0.16]$}
\expandafter\def\csname N@gss-qwen-otp-abl-nosharp-dcorr\endcsname{0.08}
\expandafter\def\csname N@gss-qwen-otp-abl-nosharp-dcorr-ci\endcsname{$[-0.11, 0.32]$}
\expandafter\def\csname N@gss-qwen-otp-abl-nosharp-rel\endcsname{0.61}
\expandafter\def\csname N@gss-qwen-otp-agg-corr\endcsname{1.29}
\expandafter\def\csname N@gss-qwen-otp-agg-cross\endcsname{0.66}
\expandafter\def\csname N@gss-qwen-otp-agg-pwin\endcsname{92.5\%}
\expandafter\def\csname N@gss-qwen-otp-agg-gap\endcsname{1.02}
\expandafter\def\csname N@gss-qwen-otp-agg-dcross\endcsname{0.44}
\expandafter\def\csname N@gss-qwen-otp-agg-dcross-ci\endcsname{$[0.18, 0.70]$}
\expandafter\def\csname N@gss-qwen-otp-agg-dgap\endcsname{0.01}
\expandafter\def\csname N@gss-qwen-otp-agg-dgap-ci\endcsname{$[-0.17, 0.18]$}
\expandafter\def\csname N@gss-qwen-otp-agg-dcorr\endcsname{$-$0.18}
\expandafter\def\csname N@gss-qwen-otp-agg-dcorr-ci\endcsname{$[-0.40, 0.06]$}
\expandafter\def\csname N@gss-qwen-otp-agg-rel\endcsname{0.19}
\expandafter\def\csname N@gss-qwen-otp-abl-seed2-corr\endcsname{1.12}
\expandafter\def\csname N@gss-qwen-otp-abl-seed2-cross\endcsname{1.10}
\expandafter\def\csname N@gss-qwen-otp-abl-seed2-pwin\endcsname{52.5\%}
\expandafter\def\csname N@gss-qwen-otp-abl-seed2-gap\endcsname{1.02}
\expandafter\def\csname N@gss-qwen-otp-abl-seed2-dcross\endcsname{0.00}
\expandafter\def\csname N@gss-qwen-otp-abl-seed2-dcross-ci\endcsname{$[-0.31, 0.32]$}
\expandafter\def\csname N@gss-qwen-otp-abl-seed2-dgap\endcsname{$-$0.00}
\expandafter\def\csname N@gss-qwen-otp-abl-seed2-dgap-ci\endcsname{$[-0.19, 0.19]$}
\expandafter\def\csname N@gss-qwen-otp-abl-seed2-dcorr\endcsname{$-$0.01}
\expandafter\def\csname N@gss-qwen-otp-abl-seed2-dcorr-ci\endcsname{$[-0.23, 0.20]$}
\expandafter\def\csname N@gss-qwen-otp-abl-seed2-rel\endcsname{0.89}
\expandafter\def\csname N@gss-qwen-otp-abl-demsharp-corr\endcsname{1.37}
\expandafter\def\csname N@gss-qwen-otp-abl-demsharp-cross\endcsname{1.46}
\expandafter\def\csname N@gss-qwen-otp-abl-demsharp-pwin\endcsname{100.0\%}
\expandafter\def\csname N@gss-qwen-otp-abl-demsharp-gap\endcsname{1.02}
\expandafter\def\csname N@gss-qwen-otp-abl-demsharp-dcross\endcsname{$-$0.37}
\expandafter\def\csname N@gss-qwen-otp-abl-demsharp-dcross-ci\endcsname{$[-0.67, -0.11]$}
\expandafter\def\csname N@gss-qwen-otp-abl-demsharp-dgap\endcsname{0.01}
\expandafter\def\csname N@gss-qwen-otp-abl-demsharp-dgap-ci\endcsname{$[-0.16, 0.20]$}
\expandafter\def\csname N@gss-qwen-otp-abl-demsharp-dcorr\endcsname{$-$0.25}
\expandafter\def\csname N@gss-qwen-otp-abl-demsharp-dcorr-ci\endcsname{$[-0.48, -0.04]$}
\expandafter\def\csname N@gss-qwen-otp-abl-demsharp-rel\endcsname{0.86}
\expandafter\def\csname N@gss-qwen-otp-demo-corr\endcsname{1.23}
\expandafter\def\csname N@gss-qwen-otp-demo-cross\endcsname{1.33}
\expandafter\def\csname N@gss-qwen-otp-demo-pwin\endcsname{85.0\%}
\expandafter\def\csname N@gss-qwen-otp-demo-gap\endcsname{1.02}
\expandafter\def\csname N@gss-qwen-otp-demo-dcross\endcsname{$-$0.23}
\expandafter\def\csname N@gss-qwen-otp-demo-dcross-ci\endcsname{$[-0.57, 0.06]$}
\expandafter\def\csname N@gss-qwen-otp-demo-dgap\endcsname{$-$0.00}
\expandafter\def\csname N@gss-qwen-otp-demo-dgap-ci\endcsname{$[-0.18, 0.18]$}
\expandafter\def\csname N@gss-qwen-otp-demo-dcorr\endcsname{$-$0.11}
\expandafter\def\csname N@gss-qwen-otp-demo-dcorr-ci\endcsname{$[-0.33, 0.08]$}
\expandafter\def\csname N@gss-qwen-otp-demo-rel\endcsname{0.58}
\expandafter\def\csname N@gss-qwen-pool-floor-corr\endcsname{0.72}
\expandafter\def\csname N@gss-qwen-pool-floor-cross\endcsname{1.08}
\expandafter\def\csname N@gss-qwen-pool-lpp-corr\endcsname{1.26}
\expandafter\def\csname N@gss-qwen-pool-lpp-rel\endcsname{0.84}
\expandafter\def\csname N@gss-qwen-pool-lpp-cross\endcsname{0.88}
\expandafter\def\csname N@gss-qwen-pool-lpp-within\endcsname{0.77}
\expandafter\def\csname N@gss-qwen-pool-lpp-gap\endcsname{0.65}
\expandafter\def\csname N@gss-qwen-pool-lpp-wpsd\endcsname{1.05}
\expandafter\def\csname N@gss-qwen-pool-lpp-edp\endcsname{0.093}
\expandafter\def\csname N@gss-qwen-pool-sel-corr\endcsname{1.26--1.57}
\expandafter\def\csname N@gss-qwen-pool-sel-rel\endcsname{0.42--0.62}
\expandafter\def\csname N@gss-qwen-pool-sel-n\endcsname{7}
\expandafter\def\csname N@gss-qwen-pool-sel-nworse\endcsname{7}
\expandafter\def\csname N@gss-qwen-pool-sel-divbest-corr\endcsname{1.34}
\expandafter\def\csname N@gss-qwen-pool-sel-divbest-rel\endcsname{0.52}
\expandafter\def\csname N@gss-qwen-pool-sel-divbest-cross\endcsname{1.16}
\expandafter\def\csname N@gss-qwen-pool-sel-divbest-d\endcsname{$-$0.08}
\expandafter\def\csname N@gss-qwen-pool-sel-divbest-ci\endcsname{$[-0.31, 0.10]$}
\expandafter\def\csname N@gss-qwen-pool-sel-divbest-pwin\endcsname{76\%}
\expandafter\def\csname N@gss-qwen-pool-sel-divbest-edp\endcsname{0.085}
\expandafter\def\csname N@gss-qwen-pool-sel-divbest-edpci\endcsname{$[-0.081, 0.082]$}
\expandafter\def\csname N@gss-qwen-pool-sel-rw-corr\endcsname{1.47}
\expandafter\def\csname N@gss-qwen-pool-sel-rw-rel\endcsname{0.62}
\expandafter\def\csname N@gss-qwen-pool-sel-rw-cross\endcsname{1.42}
\expandafter\def\csname N@gss-qwen-pool-sel-rw-d\endcsname{$-$0.22}
\expandafter\def\csname N@gss-qwen-pool-sel-rw-ci\endcsname{$[-0.44, 0.03]$}
\expandafter\def\csname N@gss-qwen-pool-sel-rw-pwin\endcsname{96\%}
\expandafter\def\csname N@gss-qwen-pool-sel-rw-edp\endcsname{0.103}
\expandafter\def\csname N@gss-qwen-pool-sel-rw-edpci\endcsname{$[-0.105, 0.081]$}
\expandafter\def\csname N@gss-qwen-pool-sel-base-corr\endcsname{1.26}
\expandafter\def\csname N@gss-qwen-pool-sel-base-rel\endcsname{0.42}
\expandafter\def\csname N@gss-qwen-pool-sel-base-cross\endcsname{1.01}
\expandafter\def\csname N@gss-qwen-pool-sel-base-d\endcsname{$-$0.00}
\expandafter\def\csname N@gss-qwen-pool-sel-base-ci\endcsname{$[-0.20, 0.21]$}
\expandafter\def\csname N@gss-qwen-pool-sel-base-pwin\endcsname{55\%}
\expandafter\def\csname N@gss-qwen-pool-sel-base-edp\endcsname{0.070}
\expandafter\def\csname N@gss-qwen-pool-sel-base-edpci\endcsname{$[-0.062, 0.098]$}
\expandafter\def\csname N@gss-qwen-pool-sel-div-corr\endcsname{1.44}
\expandafter\def\csname N@gss-qwen-pool-sel-div-rel\endcsname{0.59}
\expandafter\def\csname N@gss-qwen-pool-sel-div-cross\endcsname{1.35}
\expandafter\def\csname N@gss-qwen-pool-sel-div-d\endcsname{$-$0.18}
\expandafter\def\csname N@gss-qwen-pool-sel-div-ci\endcsname{$[-0.40, 0.06]$}
\expandafter\def\csname N@gss-qwen-pool-sel-div-pwin\endcsname{92\%}
\expandafter\def\csname N@gss-qwen-pool-sel-div-edp\endcsname{0.097}
\expandafter\def\csname N@gss-qwen-pool-sel-div-edpci\endcsname{$[-0.088, 0.075]$}
\expandafter\def\csname N@gss-qwen-pool-sel-argyle-corr\endcsname{1.47}
\expandafter\def\csname N@gss-qwen-pool-sel-argyle-rel\endcsname{0.55}
\expandafter\def\csname N@gss-qwen-pool-sel-argyle-cross\endcsname{1.37}
\expandafter\def\csname N@gss-qwen-pool-sel-argyle-d\endcsname{$-$0.21}
\expandafter\def\csname N@gss-qwen-pool-sel-argyle-ci\endcsname{$[-0.46, 0.03]$}
\expandafter\def\csname N@gss-qwen-pool-sel-argyle-pwin\endcsname{96\%}
\expandafter\def\csname N@gss-qwen-pool-sel-argyle-edp\endcsname{0.152}
\expandafter\def\csname N@gss-qwen-pool-sel-argyle-edpci\endcsname{$[-0.164, 0.044]$}
\expandafter\def\csname N@gss-qwen-pool-sel-story-corr\endcsname{1.44}
\expandafter\def\csname N@gss-qwen-pool-sel-story-rel\endcsname{0.55}
\expandafter\def\csname N@gss-qwen-pool-sel-story-cross\endcsname{1.33}
\expandafter\def\csname N@gss-qwen-pool-sel-story-d\endcsname{$-$0.18}
\expandafter\def\csname N@gss-qwen-pool-sel-story-ci\endcsname{$[-0.38, 0.05]$}
\expandafter\def\csname N@gss-qwen-pool-sel-story-pwin\endcsname{92\%}
\expandafter\def\csname N@gss-qwen-pool-sel-story-edp\endcsname{0.108}
\expandafter\def\csname N@gss-qwen-pool-sel-story-edpci\endcsname{$[-0.095, 0.069]$}
\expandafter\def\csname N@gss-qwen-pool-sel-anth-corr\endcsname{1.57}
\expandafter\def\csname N@gss-qwen-pool-sel-anth-rel\endcsname{0.43}
\expandafter\def\csname N@gss-qwen-pool-sel-anth-cross\endcsname{0.52}
\expandafter\def\csname N@gss-qwen-pool-sel-anth-d\endcsname{$-$0.31}
\expandafter\def\csname N@gss-qwen-pool-sel-anth-ci\endcsname{$[-0.53, -0.07]$}
\expandafter\def\csname N@gss-qwen-pool-sel-anth-pwin\endcsname{100\%}
\expandafter\def\csname N@gss-qwen-pool-sel-anth-edp\endcsname{0.157}
\expandafter\def\csname N@gss-qwen-pool-sel-anth-edpci\endcsname{$[-0.142, 0.021]$}
\expandafter\def\csname N@gss-qwen-pool-sel-drange\endcsname{$-$0.31 to $-$0.00}
\expandafter\def\csname N@gss-qwen-pool-sel-nsig\endcsname{1}
\expandafter\def\csname N@gss-qwen-pool-sel-relx\endcsname{1.3--2.0}
\expandafter\def\csname N@gss-qwen-pool-sel-pwinmin\endcsname{55\%}
\expandafter\def\csname N@gss-qwen-pool-sel-pwinphrase\endcsname{at least 55\% of}
\expandafter\def\csname N@gss-qwen-pool-shp-corr\endcsname{1.68--2.51}
\expandafter\def\csname N@gss-qwen-pool-shp-rel\endcsname{0.98--1.21}
\expandafter\def\csname N@gss-qwen-pool-shp-n\endcsname{4}
\expandafter\def\csname N@gss-qwen-pool-shp-nworse\endcsname{4}
\expandafter\def\csname N@gss-qwen-pool-shp-base-corr\endcsname{1.68}
\expandafter\def\csname N@gss-qwen-pool-shp-base-rel\endcsname{1.00}
\expandafter\def\csname N@gss-qwen-pool-shp-base-cross\endcsname{1.62}
\expandafter\def\csname N@gss-qwen-pool-shp-base-d\endcsname{$-$0.43}
\expandafter\def\csname N@gss-qwen-pool-shp-base-ci\endcsname{$[-0.66, -0.15]$}
\expandafter\def\csname N@gss-qwen-pool-shp-base-pwin\endcsname{100\%}
\expandafter\def\csname N@gss-qwen-pool-shp-base-edp\endcsname{0.140}
\expandafter\def\csname N@gss-qwen-pool-shp-base-edpci\endcsname{$[-0.140, 0.052]$}
\expandafter\def\csname N@gss-qwen-pool-shp-div-corr\endcsname{1.77}
\expandafter\def\csname N@gss-qwen-pool-shp-div-rel\endcsname{0.98}
\expandafter\def\csname N@gss-qwen-pool-shp-div-cross\endcsname{1.68}
\expandafter\def\csname N@gss-qwen-pool-shp-div-d\endcsname{$-$0.51}
\expandafter\def\csname N@gss-qwen-pool-shp-div-ci\endcsname{$[-0.75, -0.29]$}
\expandafter\def\csname N@gss-qwen-pool-shp-div-pwin\endcsname{100\%}
\expandafter\def\csname N@gss-qwen-pool-shp-div-edp\endcsname{0.166}
\expandafter\def\csname N@gss-qwen-pool-shp-div-edpci\endcsname{$[-0.162, 0.011]$}
\expandafter\def\csname N@gss-qwen-pool-shp-argyle-corr\endcsname{2.51}
\expandafter\def\csname N@gss-qwen-pool-shp-argyle-rel\endcsname{1.21}
\expandafter\def\csname N@gss-qwen-pool-shp-argyle-cross\endcsname{2.51}
\expandafter\def\csname N@gss-qwen-pool-shp-argyle-d\endcsname{$-$1.25}
\expandafter\def\csname N@gss-qwen-pool-shp-argyle-ci\endcsname{$[-1.50, -0.99]$}
\expandafter\def\csname N@gss-qwen-pool-shp-argyle-pwin\endcsname{100\%}
\expandafter\def\csname N@gss-qwen-pool-shp-argyle-edp\endcsname{0.468}
\expandafter\def\csname N@gss-qwen-pool-shp-argyle-edpci\endcsname{$[-0.524, -0.254]$}
\expandafter\def\csname N@gss-qwen-pool-shp-story-corr\endcsname{1.99}
\expandafter\def\csname N@gss-qwen-pool-shp-story-rel\endcsname{0.99}
\expandafter\def\csname N@gss-qwen-pool-shp-story-cross\endcsname{1.86}
\expandafter\def\csname N@gss-qwen-pool-shp-story-d\endcsname{$-$0.73}
\expandafter\def\csname N@gss-qwen-pool-shp-story-ci\endcsname{$[-0.98, -0.49]$}
\expandafter\def\csname N@gss-qwen-pool-shp-story-pwin\endcsname{100\%}
\expandafter\def\csname N@gss-qwen-pool-shp-story-edp\endcsname{0.229}
\expandafter\def\csname N@gss-qwen-pool-shp-story-edpci\endcsname{$[-0.240, -0.037]$}
\expandafter\def\csname N@gss-qwen-pool-shp-drange\endcsname{$-$1.25 to $-$0.43}
\expandafter\def\csname N@gss-qwen-pool-shp-nsig\endcsname{4}
\expandafter\def\csname N@gss-qwen-pool-shp-relx\endcsname{0.7--0.9}
\expandafter\def\csname N@gss-qwen-pool-shp-pwinmin\endcsname{100\%}
\expandafter\def\csname N@gss-qwen-pool-shp-pwinphrase\endcsname{all}
\expandafter\def\csname N@gss-qwen-pool-abl-k16-corr\endcsname{1.28}
\expandafter\def\csname N@gss-qwen-pool-abl-k16-cross\endcsname{0.78}
\expandafter\def\csname N@gss-qwen-pool-abl-k16-pwin\endcsname{57.0\%}
\expandafter\def\csname N@gss-qwen-pool-abl-k16-gap\endcsname{0.66}
\expandafter\def\csname N@gss-qwen-pool-abl-k16-dcross\endcsname{0.10}
\expandafter\def\csname N@gss-qwen-pool-abl-k16-dcross-ci\endcsname{$[-0.13, 0.32]$}
\expandafter\def\csname N@gss-qwen-pool-abl-k16-dgap\endcsname{$-$0.01}
\expandafter\def\csname N@gss-qwen-pool-abl-k16-dgap-ci\endcsname{$[-0.20, 0.20]$}
\expandafter\def\csname N@gss-qwen-pool-abl-k16-dcorr\endcsname{$-$0.02}
\expandafter\def\csname N@gss-qwen-pool-abl-k16-dcorr-ci\endcsname{$[-0.24, 0.19]$}
\expandafter\def\csname N@gss-qwen-pool-abl-k16-rel\endcsname{0.73}
\expandafter\def\csname N@gss-qwen-pool-abl-nosharp-corr\endcsname{1.21}
\expandafter\def\csname N@gss-qwen-pool-abl-nosharp-cross\endcsname{0.72}
\expandafter\def\csname N@gss-qwen-pool-abl-nosharp-pwin\endcsname{37.5\%}
\expandafter\def\csname N@gss-qwen-pool-abl-nosharp-gap\endcsname{0.60}
\expandafter\def\csname N@gss-qwen-pool-abl-nosharp-dcross\endcsname{0.16}
\expandafter\def\csname N@gss-qwen-pool-abl-nosharp-dcross-ci\endcsname{$[-0.06, 0.36]$}
\expandafter\def\csname N@gss-qwen-pool-abl-nosharp-dgap\endcsname{0.05}
\expandafter\def\csname N@gss-qwen-pool-abl-nosharp-dgap-ci\endcsname{$[-0.08, 0.20]$}
\expandafter\def\csname N@gss-qwen-pool-abl-nosharp-dcorr\endcsname{0.05}
\expandafter\def\csname N@gss-qwen-pool-abl-nosharp-dcorr-ci\endcsname{$[-0.16, 0.25]$}
\expandafter\def\csname N@gss-qwen-pool-abl-nosharp-rel\endcsname{0.52}
\expandafter\def\csname N@gss-qwen-pool-agg-corr\endcsname{1.74}
\expandafter\def\csname N@gss-qwen-pool-agg-cross\endcsname{0.40}
\expandafter\def\csname N@gss-qwen-pool-agg-pwin\endcsname{100.0\%}
\expandafter\def\csname N@gss-qwen-pool-agg-gap\endcsname{0.19}
\expandafter\def\csname N@gss-qwen-pool-agg-dcross\endcsname{0.48}
\expandafter\def\csname N@gss-qwen-pool-agg-dcross-ci\endcsname{$[0.27, 0.66]$}
\expandafter\def\csname N@gss-qwen-pool-agg-dgap\endcsname{0.46}
\expandafter\def\csname N@gss-qwen-pool-agg-dgap-ci\endcsname{$[0.26, 0.64]$}
\expandafter\def\csname N@gss-qwen-pool-agg-dcorr\endcsname{$-$0.48}
\expandafter\def\csname N@gss-qwen-pool-agg-dcorr-ci\endcsname{$[-0.70, -0.26]$}
\expandafter\def\csname N@gss-qwen-pool-agg-rel\endcsname{0.00}
\expandafter\def\csname N@gss-qwen-pool-abl-seed2-corr\endcsname{1.27}
\expandafter\def\csname N@gss-qwen-pool-abl-seed2-cross\endcsname{0.89}
\expandafter\def\csname N@gss-qwen-pool-abl-seed2-pwin\endcsname{54.0\%}
\expandafter\def\csname N@gss-qwen-pool-abl-seed2-gap\endcsname{0.70}
\expandafter\def\csname N@gss-qwen-pool-abl-seed2-dcross\endcsname{$-$0.01}
\expandafter\def\csname N@gss-qwen-pool-abl-seed2-dcross-ci\endcsname{$[-0.25, 0.25]$}
\expandafter\def\csname N@gss-qwen-pool-abl-seed2-dgap\endcsname{$-$0.05}
\expandafter\def\csname N@gss-qwen-pool-abl-seed2-dgap-ci\endcsname{$[-0.28, 0.14]$}
\expandafter\def\csname N@gss-qwen-pool-abl-seed2-dcorr\endcsname{$-$0.01}
\expandafter\def\csname N@gss-qwen-pool-abl-seed2-dcorr-ci\endcsname{$[-0.22, 0.22]$}
\expandafter\def\csname N@gss-qwen-pool-abl-seed2-rel\endcsname{0.83}
\expandafter\def\csname N@gss-qwen-pool-abl-demsharp-corr\endcsname{1.69}
\expandafter\def\csname N@gss-qwen-pool-abl-demsharp-cross\endcsname{1.64}
\expandafter\def\csname N@gss-qwen-pool-abl-demsharp-pwin\endcsname{100.0\%}
\expandafter\def\csname N@gss-qwen-pool-abl-demsharp-gap\endcsname{1.24}
\expandafter\def\csname N@gss-qwen-pool-abl-demsharp-dcross\endcsname{$-$0.76}
\expandafter\def\csname N@gss-qwen-pool-abl-demsharp-dcross-ci\endcsname{$[-1.01, -0.50]$}
\expandafter\def\csname N@gss-qwen-pool-abl-demsharp-dgap\endcsname{$-$0.59}
\expandafter\def\csname N@gss-qwen-pool-abl-demsharp-dgap-ci\endcsname{$[-0.78, -0.39]$}
\expandafter\def\csname N@gss-qwen-pool-abl-demsharp-dcorr\endcsname{$-$0.43}
\expandafter\def\csname N@gss-qwen-pool-abl-demsharp-dcorr-ci\endcsname{$[-0.67, -0.16]$}
\expandafter\def\csname N@gss-qwen-pool-abl-demsharp-rel\endcsname{0.94}
\expandafter\def\csname N@gss-qwen-pool-demo-corr\endcsname{1.47}
\expandafter\def\csname N@gss-qwen-pool-demo-cross\endcsname{1.43}
\expandafter\def\csname N@gss-qwen-pool-demo-pwin\endcsname{98.0\%}
\expandafter\def\csname N@gss-qwen-pool-demo-gap\endcsname{1.18}
\expandafter\def\csname N@gss-qwen-pool-demo-dcross\endcsname{$-$0.55}
\expandafter\def\csname N@gss-qwen-pool-demo-dcross-ci\endcsname{$[-0.82, -0.28]$}
\expandafter\def\csname N@gss-qwen-pool-demo-dgap\endcsname{$-$0.53}
\expandafter\def\csname N@gss-qwen-pool-demo-dgap-ci\endcsname{$[-0.74, -0.34]$}
\expandafter\def\csname N@gss-qwen-pool-demo-dcorr\endcsname{$-$0.22}
\expandafter\def\csname N@gss-qwen-pool-demo-dcorr-ci\endcsname{$[-0.46, -0.01]$}
\expandafter\def\csname N@gss-qwen-pool-demo-rel\endcsname{0.62}
\expandafter\def\csname N@gss-gemma-otp-floor-corr\endcsname{0.72}
\expandafter\def\csname N@gss-gemma-otp-floor-cross\endcsname{1.08}
\expandafter\def\csname N@gss-gemma-otp-lpp-corr\endcsname{0.98}
\expandafter\def\csname N@gss-gemma-otp-lpp-rel\endcsname{1.20}
\expandafter\def\csname N@gss-gemma-otp-lpp-cross\endcsname{1.03}
\expandafter\def\csname N@gss-gemma-otp-lpp-within\endcsname{1.03}
\expandafter\def\csname N@gss-gemma-otp-lpp-gap\endcsname{1.02}
\expandafter\def\csname N@gss-gemma-otp-lpp-wpsd\endcsname{1.02}
\expandafter\def\csname N@gss-gemma-otp-lpp-edp\endcsname{0.011}
\expandafter\def\csname N@gss-gemma-otp-sel-corr\endcsname{1.23--1.47}
\expandafter\def\csname N@gss-gemma-otp-sel-rel\endcsname{0.65--1.02}
\expandafter\def\csname N@gss-gemma-otp-sel-n\endcsname{7}
\expandafter\def\csname N@gss-gemma-otp-sel-nworse\endcsname{7}
\expandafter\def\csname N@gss-gemma-otp-sel-divbest-corr\endcsname{1.32}
\expandafter\def\csname N@gss-gemma-otp-sel-divbest-rel\endcsname{0.85}
\expandafter\def\csname N@gss-gemma-otp-sel-divbest-cross\endcsname{1.22}
\expandafter\def\csname N@gss-gemma-otp-sel-divbest-d\endcsname{$-$0.35}
\expandafter\def\csname N@gss-gemma-otp-sel-divbest-ci\endcsname{$[-0.54, -0.11]$}
\expandafter\def\csname N@gss-gemma-otp-sel-divbest-pwin\endcsname{100\%}
\expandafter\def\csname N@gss-gemma-otp-sel-divbest-edp\endcsname{0.020}
\expandafter\def\csname N@gss-gemma-otp-sel-divbest-edpci\endcsname{$[-0.051, 0.029]$}
\expandafter\def\csname N@gss-gemma-otp-sel-rw-corr\endcsname{1.40}
\expandafter\def\csname N@gss-gemma-otp-sel-rw-rel\endcsname{0.97}
\expandafter\def\csname N@gss-gemma-otp-sel-rw-cross\endcsname{1.55}
\expandafter\def\csname N@gss-gemma-otp-sel-rw-d\endcsname{$-$0.42}
\expandafter\def\csname N@gss-gemma-otp-sel-rw-ci\endcsname{$[-0.68, -0.14]$}
\expandafter\def\csname N@gss-gemma-otp-sel-rw-pwin\endcsname{100\%}
\expandafter\def\csname N@gss-gemma-otp-sel-rw-edp\endcsname{0.030}
\expandafter\def\csname N@gss-gemma-otp-sel-rw-edpci\endcsname{$[-0.066, 0.024]$}
\expandafter\def\csname N@gss-gemma-otp-sel-base-corr\endcsname{1.39}
\expandafter\def\csname N@gss-gemma-otp-sel-base-rel\endcsname{0.91}
\expandafter\def\csname N@gss-gemma-otp-sel-base-cross\endcsname{1.53}
\expandafter\def\csname N@gss-gemma-otp-sel-base-d\endcsname{$-$0.41}
\expandafter\def\csname N@gss-gemma-otp-sel-base-ci\endcsname{$[-0.62, -0.19]$}
\expandafter\def\csname N@gss-gemma-otp-sel-base-pwin\endcsname{100\%}
\expandafter\def\csname N@gss-gemma-otp-sel-base-edp\endcsname{0.021}
\expandafter\def\csname N@gss-gemma-otp-sel-base-edpci\endcsname{$[-0.050, 0.025]$}
\expandafter\def\csname N@gss-gemma-otp-sel-div-corr\endcsname{1.23}
\expandafter\def\csname N@gss-gemma-otp-sel-div-rel\endcsname{0.84}
\expandafter\def\csname N@gss-gemma-otp-sel-div-cross\endcsname{1.15}
\expandafter\def\csname N@gss-gemma-otp-sel-div-d\endcsname{$-$0.25}
\expandafter\def\csname N@gss-gemma-otp-sel-div-ci\endcsname{$[-0.45, -0.01]$}
\expandafter\def\csname N@gss-gemma-otp-sel-div-pwin\endcsname{98\%}
\expandafter\def\csname N@gss-gemma-otp-sel-div-edp\endcsname{0.018}
\expandafter\def\csname N@gss-gemma-otp-sel-div-edpci\endcsname{$[-0.054, 0.029]$}
\expandafter\def\csname N@gss-gemma-otp-sel-argyle-corr\endcsname{1.23}
\expandafter\def\csname N@gss-gemma-otp-sel-argyle-rel\endcsname{0.82}
\expandafter\def\csname N@gss-gemma-otp-sel-argyle-cross\endcsname{1.22}
\expandafter\def\csname N@gss-gemma-otp-sel-argyle-d\endcsname{$-$0.25}
\expandafter\def\csname N@gss-gemma-otp-sel-argyle-ci\endcsname{$[-0.46, -0.03]$}
\expandafter\def\csname N@gss-gemma-otp-sel-argyle-pwin\endcsname{99\%}
\expandafter\def\csname N@gss-gemma-otp-sel-argyle-edp\endcsname{0.019}
\expandafter\def\csname N@gss-gemma-otp-sel-argyle-edpci\endcsname{$[-0.052, 0.032]$}
\expandafter\def\csname N@gss-gemma-otp-sel-story-corr\endcsname{1.47}
\expandafter\def\csname N@gss-gemma-otp-sel-story-rel\endcsname{1.02}
\expandafter\def\csname N@gss-gemma-otp-sel-story-cross\endcsname{1.72}
\expandafter\def\csname N@gss-gemma-otp-sel-story-d\endcsname{$-$0.50}
\expandafter\def\csname N@gss-gemma-otp-sel-story-ci\endcsname{$[-0.77, -0.23]$}
\expandafter\def\csname N@gss-gemma-otp-sel-story-pwin\endcsname{100\%}
\expandafter\def\csname N@gss-gemma-otp-sel-story-edp\endcsname{0.023}
\expandafter\def\csname N@gss-gemma-otp-sel-story-edpci\endcsname{$[-0.053, 0.023]$}
\expandafter\def\csname N@gss-gemma-otp-sel-anth-corr\endcsname{1.32}
\expandafter\def\csname N@gss-gemma-otp-sel-anth-rel\endcsname{0.65}
\expandafter\def\csname N@gss-gemma-otp-sel-anth-cross\endcsname{0.93}
\expandafter\def\csname N@gss-gemma-otp-sel-anth-d\endcsname{$-$0.35}
\expandafter\def\csname N@gss-gemma-otp-sel-anth-ci\endcsname{$[-0.58, -0.12]$}
\expandafter\def\csname N@gss-gemma-otp-sel-anth-pwin\endcsname{100\%}
\expandafter\def\csname N@gss-gemma-otp-sel-anth-edp\endcsname{0.022}
\expandafter\def\csname N@gss-gemma-otp-sel-anth-edpci\endcsname{$[-0.054, 0.032]$}
\expandafter\def\csname N@gss-gemma-otp-sel-drange\endcsname{$-$0.50 to $-$0.25}
\expandafter\def\csname N@gss-gemma-otp-sel-nsig\endcsname{7}
\expandafter\def\csname N@gss-gemma-otp-sel-relx\endcsname{1.2--1.8}
\expandafter\def\csname N@gss-gemma-otp-sel-pwinmin\endcsname{98\%}
\expandafter\def\csname N@gss-gemma-otp-sel-pwinphrase\endcsname{at least 98\% of}
\expandafter\def\csname N@gss-gemma-otp-shp-corr\endcsname{1.23--1.48}
\expandafter\def\csname N@gss-gemma-otp-shp-rel\endcsname{0.88--1.10}
\expandafter\def\csname N@gss-gemma-otp-shp-n\endcsname{4}
\expandafter\def\csname N@gss-gemma-otp-shp-nworse\endcsname{4}
\expandafter\def\csname N@gss-gemma-otp-shp-base-corr\endcsname{1.40}
\expandafter\def\csname N@gss-gemma-otp-shp-base-rel\endcsname{1.02}
\expandafter\def\csname N@gss-gemma-otp-shp-base-cross\endcsname{1.56}
\expandafter\def\csname N@gss-gemma-otp-shp-base-d\endcsname{$-$0.42}
\expandafter\def\csname N@gss-gemma-otp-shp-base-ci\endcsname{$[-0.67, -0.18]$}
\expandafter\def\csname N@gss-gemma-otp-shp-base-pwin\endcsname{100\%}
\expandafter\def\csname N@gss-gemma-otp-shp-base-edp\endcsname{0.019}
\expandafter\def\csname N@gss-gemma-otp-shp-base-edpci\endcsname{$[-0.045, 0.028]$}
\expandafter\def\csname N@gss-gemma-otp-shp-div-corr\endcsname{1.23}
\expandafter\def\csname N@gss-gemma-otp-shp-div-rel\endcsname{0.88}
\expandafter\def\csname N@gss-gemma-otp-shp-div-cross\endcsname{1.15}
\expandafter\def\csname N@gss-gemma-otp-shp-div-d\endcsname{$-$0.25}
\expandafter\def\csname N@gss-gemma-otp-shp-div-ci\endcsname{$[-0.46, -0.07]$}
\expandafter\def\csname N@gss-gemma-otp-shp-div-pwin\endcsname{100\%}
\expandafter\def\csname N@gss-gemma-otp-shp-div-edp\endcsname{0.017}
\expandafter\def\csname N@gss-gemma-otp-shp-div-edpci\endcsname{$[-0.047, 0.032]$}
\expandafter\def\csname N@gss-gemma-otp-shp-argyle-corr\endcsname{1.24}
\expandafter\def\csname N@gss-gemma-otp-shp-argyle-rel\endcsname{0.89}
\expandafter\def\csname N@gss-gemma-otp-shp-argyle-cross\endcsname{1.20}
\expandafter\def\csname N@gss-gemma-otp-shp-argyle-d\endcsname{$-$0.26}
\expandafter\def\csname N@gss-gemma-otp-shp-argyle-ci\endcsname{$[-0.44, -0.07]$}
\expandafter\def\csname N@gss-gemma-otp-shp-argyle-pwin\endcsname{99\%}
\expandafter\def\csname N@gss-gemma-otp-shp-argyle-edp\endcsname{0.022}
\expandafter\def\csname N@gss-gemma-otp-shp-argyle-edpci\endcsname{$[-0.059, 0.031]$}
\expandafter\def\csname N@gss-gemma-otp-shp-story-corr\endcsname{1.48}
\expandafter\def\csname N@gss-gemma-otp-shp-story-rel\endcsname{1.10}
\expandafter\def\csname N@gss-gemma-otp-shp-story-cross\endcsname{1.72}
\expandafter\def\csname N@gss-gemma-otp-shp-story-d\endcsname{$-$0.51}
\expandafter\def\csname N@gss-gemma-otp-shp-story-ci\endcsname{$[-0.76, -0.28]$}
\expandafter\def\csname N@gss-gemma-otp-shp-story-pwin\endcsname{100\%}
\expandafter\def\csname N@gss-gemma-otp-shp-story-edp\endcsname{0.023}
\expandafter\def\csname N@gss-gemma-otp-shp-story-edpci\endcsname{$[-0.053, 0.025]$}
\expandafter\def\csname N@gss-gemma-otp-shp-drange\endcsname{$-$0.51 to $-$0.25}
\expandafter\def\csname N@gss-gemma-otp-shp-nsig\endcsname{4}
\expandafter\def\csname N@gss-gemma-otp-shp-relx\endcsname{1.1--1.4}
\expandafter\def\csname N@gss-gemma-otp-shp-pwinmin\endcsname{99\%}
\expandafter\def\csname N@gss-gemma-otp-shp-pwinphrase\endcsname{at least 99\% of}
\expandafter\def\csname N@gss-gemma-otp-abl-k16-corr\endcsname{0.98}
\expandafter\def\csname N@gss-gemma-otp-abl-k16-cross\endcsname{0.85}
\expandafter\def\csname N@gss-gemma-otp-abl-k16-pwin\endcsname{52.0\%}
\expandafter\def\csname N@gss-gemma-otp-abl-k16-gap\endcsname{1.02}
\expandafter\def\csname N@gss-gemma-otp-abl-k16-dcross\endcsname{0.17}
\expandafter\def\csname N@gss-gemma-otp-abl-k16-dcross-ci\endcsname{$[-0.11, 0.43]$}
\expandafter\def\csname N@gss-gemma-otp-abl-k16-dgap\endcsname{0.01}
\expandafter\def\csname N@gss-gemma-otp-abl-k16-dgap-ci\endcsname{$[-0.17, 0.21]$}
\expandafter\def\csname N@gss-gemma-otp-abl-k16-dcorr\endcsname{$-$0.00}
\expandafter\def\csname N@gss-gemma-otp-abl-k16-dcorr-ci\endcsname{$[-0.18, 0.17]$}
\expandafter\def\csname N@gss-gemma-otp-abl-k16-rel\endcsname{1.05}
\expandafter\def\csname N@gss-gemma-otp-abl-nosharp-corr\endcsname{0.97}
\expandafter\def\csname N@gss-gemma-otp-abl-nosharp-cross\endcsname{1.03}
\expandafter\def\csname N@gss-gemma-otp-abl-nosharp-pwin\endcsname{47.0\%}
\expandafter\def\csname N@gss-gemma-otp-abl-nosharp-gap\endcsname{1.02}
\expandafter\def\csname N@gss-gemma-otp-abl-nosharp-dcross\endcsname{$-$0.00}
\expandafter\def\csname N@gss-gemma-otp-abl-nosharp-dcross-ci\endcsname{$[-0.21, 0.23]$}
\expandafter\def\csname N@gss-gemma-otp-abl-nosharp-dgap\endcsname{0.01}
\expandafter\def\csname N@gss-gemma-otp-abl-nosharp-dgap-ci\endcsname{$[-0.12, 0.12]$}
\expandafter\def\csname N@gss-gemma-otp-abl-nosharp-dcorr\endcsname{0.01}
\expandafter\def\csname N@gss-gemma-otp-abl-nosharp-dcorr-ci\endcsname{$[-0.15, 0.23]$}
\expandafter\def\csname N@gss-gemma-otp-abl-nosharp-rel\endcsname{1.17}
\expandafter\def\csname N@gss-gemma-otp-agg-corr\endcsname{1.29}
\expandafter\def\csname N@gss-gemma-otp-agg-cross\endcsname{0.66}
\expandafter\def\csname N@gss-gemma-otp-agg-pwin\endcsname{99.5\%}
\expandafter\def\csname N@gss-gemma-otp-agg-gap\endcsname{1.02}
\expandafter\def\csname N@gss-gemma-otp-agg-dcross\endcsname{0.37}
\expandafter\def\csname N@gss-gemma-otp-agg-dcross-ci\endcsname{$[0.11, 0.68]$}
\expandafter\def\csname N@gss-gemma-otp-agg-dgap\endcsname{0.01}
\expandafter\def\csname N@gss-gemma-otp-agg-dgap-ci\endcsname{$[-0.16, 0.20]$}
\expandafter\def\csname N@gss-gemma-otp-agg-dcorr\endcsname{$-$0.31}
\expandafter\def\csname N@gss-gemma-otp-agg-dcorr-ci\endcsname{$[-0.54, -0.11]$}
\expandafter\def\csname N@gss-gemma-otp-agg-rel\endcsname{0.19}
\expandafter\def\csname N@gss-gemma-otp-abl-seed2-corr\endcsname{0.95}
\expandafter\def\csname N@gss-gemma-otp-abl-seed2-cross\endcsname{1.00}
\expandafter\def\csname N@gss-gemma-otp-abl-seed2-pwin\endcsname{44.0\%}
\expandafter\def\csname N@gss-gemma-otp-abl-seed2-gap\endcsname{1.02}
\expandafter\def\csname N@gss-gemma-otp-abl-seed2-dcross\endcsname{0.03}
\expandafter\def\csname N@gss-gemma-otp-abl-seed2-dcross-ci\endcsname{$[-0.24, 0.32]$}
\expandafter\def\csname N@gss-gemma-otp-abl-seed2-dgap\endcsname{0.00}
\expandafter\def\csname N@gss-gemma-otp-abl-seed2-dgap-ci\endcsname{$[-0.18, 0.22]$}
\expandafter\def\csname N@gss-gemma-otp-abl-seed2-dcorr\endcsname{0.02}
\expandafter\def\csname N@gss-gemma-otp-abl-seed2-dcorr-ci\endcsname{$[-0.15, 0.21]$}
\expandafter\def\csname N@gss-gemma-otp-abl-seed2-rel\endcsname{1.18}
\expandafter\def\csname N@gss-gemma-otp-abl-demsharp-corr\endcsname{1.40}
\expandafter\def\csname N@gss-gemma-otp-abl-demsharp-cross\endcsname{1.55}
\expandafter\def\csname N@gss-gemma-otp-abl-demsharp-pwin\endcsname{100.0\%}
\expandafter\def\csname N@gss-gemma-otp-abl-demsharp-gap\endcsname{1.01}
\expandafter\def\csname N@gss-gemma-otp-abl-demsharp-dcross\endcsname{$-$0.52}
\expandafter\def\csname N@gss-gemma-otp-abl-demsharp-dcross-ci\endcsname{$[-0.79, -0.24]$}
\expandafter\def\csname N@gss-gemma-otp-abl-demsharp-dgap\endcsname{0.01}
\expandafter\def\csname N@gss-gemma-otp-abl-demsharp-dgap-ci\endcsname{$[-0.16, 0.21]$}
\expandafter\def\csname N@gss-gemma-otp-abl-demsharp-dcorr\endcsname{$-$0.42}
\expandafter\def\csname N@gss-gemma-otp-abl-demsharp-dcorr-ci\endcsname{$[-0.66, -0.19]$}
\expandafter\def\csname N@gss-gemma-otp-abl-demsharp-rel\endcsname{1.01}
\expandafter\def\csname N@gss-gemma-otp-demo-corr\endcsname{1.39}
\expandafter\def\csname N@gss-gemma-otp-demo-cross\endcsname{1.55}
\expandafter\def\csname N@gss-gemma-otp-demo-pwin\endcsname{100.0\%}
\expandafter\def\csname N@gss-gemma-otp-demo-gap\endcsname{1.02}
\expandafter\def\csname N@gss-gemma-otp-demo-dcross\endcsname{$-$0.53}
\expandafter\def\csname N@gss-gemma-otp-demo-dcross-ci\endcsname{$[-0.82, -0.23]$}
\expandafter\def\csname N@gss-gemma-otp-demo-dgap\endcsname{0.00}
\expandafter\def\csname N@gss-gemma-otp-demo-dgap-ci\endcsname{$[-0.18, 0.17]$}
\expandafter\def\csname N@gss-gemma-otp-demo-dcorr\endcsname{$-$0.42}
\expandafter\def\csname N@gss-gemma-otp-demo-dcorr-ci\endcsname{$[-0.66, -0.18]$}
\expandafter\def\csname N@gss-gemma-otp-demo-rel\endcsname{0.97}
\expandafter\def\csname N@gss-gemma-pool-floor-corr\endcsname{0.72}
\expandafter\def\csname N@gss-gemma-pool-floor-cross\endcsname{1.08}
\expandafter\def\csname N@gss-gemma-pool-lpp-corr\endcsname{1.08}
\expandafter\def\csname N@gss-gemma-pool-lpp-rel\endcsname{1.24}
\expandafter\def\csname N@gss-gemma-pool-lpp-cross\endcsname{0.83}
\expandafter\def\csname N@gss-gemma-pool-lpp-within\endcsname{1.05}
\expandafter\def\csname N@gss-gemma-pool-lpp-gap\endcsname{0.76}
\expandafter\def\csname N@gss-gemma-pool-lpp-wpsd\endcsname{1.04}
\expandafter\def\csname N@gss-gemma-pool-lpp-edp\endcsname{0.069}
\expandafter\def\csname N@gss-gemma-pool-sel-corr\endcsname{1.63--2.47}
\expandafter\def\csname N@gss-gemma-pool-sel-rel\endcsname{0.55--1.22}
\expandafter\def\csname N@gss-gemma-pool-sel-n\endcsname{7}
\expandafter\def\csname N@gss-gemma-pool-sel-nworse\endcsname{7}
\expandafter\def\csname N@gss-gemma-pool-sel-divbest-corr\endcsname{1.75}
\expandafter\def\csname N@gss-gemma-pool-sel-divbest-rel\endcsname{0.93}
\expandafter\def\csname N@gss-gemma-pool-sel-divbest-cross\endcsname{1.30}
\expandafter\def\csname N@gss-gemma-pool-sel-divbest-d\endcsname{$-$0.67}
\expandafter\def\csname N@gss-gemma-pool-sel-divbest-ci\endcsname{$[-0.88, -0.47]$}
\expandafter\def\csname N@gss-gemma-pool-sel-divbest-pwin\endcsname{100\%}
\expandafter\def\csname N@gss-gemma-pool-sel-divbest-edp\endcsname{0.209}
\expandafter\def\csname N@gss-gemma-pool-sel-divbest-edpci\endcsname{$[-0.220, -0.069]$}
\expandafter\def\csname N@gss-gemma-pool-sel-rw-corr\endcsname{2.18}
\expandafter\def\csname N@gss-gemma-pool-sel-rw-rel\endcsname{1.19}
\expandafter\def\csname N@gss-gemma-pool-sel-rw-cross\endcsname{2.06}
\expandafter\def\csname N@gss-gemma-pool-sel-rw-d\endcsname{$-$1.10}
\expandafter\def\csname N@gss-gemma-pool-sel-rw-ci\endcsname{$[-1.40, -0.80]$}
\expandafter\def\csname N@gss-gemma-pool-sel-rw-pwin\endcsname{100\%}
\expandafter\def\csname N@gss-gemma-pool-sel-rw-edp\endcsname{0.272}
\expandafter\def\csname N@gss-gemma-pool-sel-rw-edpci\endcsname{$[-0.334, -0.099]$}
\expandafter\def\csname N@gss-gemma-pool-sel-base-corr\endcsname{2.12}
\expandafter\def\csname N@gss-gemma-pool-sel-base-rel\endcsname{1.12}
\expandafter\def\csname N@gss-gemma-pool-sel-base-cross\endcsname{2.01}
\expandafter\def\csname N@gss-gemma-pool-sel-base-d\endcsname{$-$1.04}
\expandafter\def\csname N@gss-gemma-pool-sel-base-ci\endcsname{$[-1.27, -0.79]$}
\expandafter\def\csname N@gss-gemma-pool-sel-base-pwin\endcsname{100\%}
\expandafter\def\csname N@gss-gemma-pool-sel-base-edp\endcsname{0.248}
\expandafter\def\csname N@gss-gemma-pool-sel-base-edpci\endcsname{$[-0.262, -0.096]$}
\expandafter\def\csname N@gss-gemma-pool-sel-div-corr\endcsname{1.70}
\expandafter\def\csname N@gss-gemma-pool-sel-div-rel\endcsname{0.97}
\expandafter\def\csname N@gss-gemma-pool-sel-div-cross\endcsname{1.38}
\expandafter\def\csname N@gss-gemma-pool-sel-div-d\endcsname{$-$0.62}
\expandafter\def\csname N@gss-gemma-pool-sel-div-ci\endcsname{$[-0.85, -0.34]$}
\expandafter\def\csname N@gss-gemma-pool-sel-div-pwin\endcsname{100\%}
\expandafter\def\csname N@gss-gemma-pool-sel-div-edp\endcsname{0.210}
\expandafter\def\csname N@gss-gemma-pool-sel-div-edpci\endcsname{$[-0.238, -0.056]$}
\expandafter\def\csname N@gss-gemma-pool-sel-argyle-corr\endcsname{2.47}
\expandafter\def\csname N@gss-gemma-pool-sel-argyle-rel\endcsname{1.22}
\expandafter\def\csname N@gss-gemma-pool-sel-argyle-cross\endcsname{2.37}
\expandafter\def\csname N@gss-gemma-pool-sel-argyle-d\endcsname{$-$1.38}
\expandafter\def\csname N@gss-gemma-pool-sel-argyle-ci\endcsname{$[-1.65, -1.11]$}
\expandafter\def\csname N@gss-gemma-pool-sel-argyle-pwin\endcsname{100\%}
\expandafter\def\csname N@gss-gemma-pool-sel-argyle-edp\endcsname{0.476}
\expandafter\def\csname N@gss-gemma-pool-sel-argyle-edpci\endcsname{$[-0.514, -0.306]$}
\expandafter\def\csname N@gss-gemma-pool-sel-story-corr\endcsname{2.16}
\expandafter\def\csname N@gss-gemma-pool-sel-story-rel\endcsname{1.19}
\expandafter\def\csname N@gss-gemma-pool-sel-story-cross\endcsname{2.20}
\expandafter\def\csname N@gss-gemma-pool-sel-story-d\endcsname{$-$1.07}
\expandafter\def\csname N@gss-gemma-pool-sel-story-ci\endcsname{$[-1.32, -0.78]$}
\expandafter\def\csname N@gss-gemma-pool-sel-story-pwin\endcsname{100\%}
\expandafter\def\csname N@gss-gemma-pool-sel-story-edp\endcsname{0.234}
\expandafter\def\csname N@gss-gemma-pool-sel-story-edpci\endcsname{$[-0.260, -0.071]$}
\expandafter\def\csname N@gss-gemma-pool-sel-anth-corr\endcsname{1.63}
\expandafter\def\csname N@gss-gemma-pool-sel-anth-rel\endcsname{0.55}
\expandafter\def\csname N@gss-gemma-pool-sel-anth-cross\endcsname{0.68}
\expandafter\def\csname N@gss-gemma-pool-sel-anth-d\endcsname{$-$0.54}
\expandafter\def\csname N@gss-gemma-pool-sel-anth-ci\endcsname{$[-0.77, -0.27]$}
\expandafter\def\csname N@gss-gemma-pool-sel-anth-pwin\endcsname{100\%}
\expandafter\def\csname N@gss-gemma-pool-sel-anth-edp\endcsname{0.192}
\expandafter\def\csname N@gss-gemma-pool-sel-anth-edpci\endcsname{$[-0.214, -0.047]$}
\expandafter\def\csname N@gss-gemma-pool-sel-drange\endcsname{$-$1.38 to $-$0.54}
\expandafter\def\csname N@gss-gemma-pool-sel-nsig\endcsname{7}
\expandafter\def\csname N@gss-gemma-pool-sel-relx\endcsname{1.0--2.3}
\expandafter\def\csname N@gss-gemma-pool-sel-pwinmin\endcsname{100\%}
\expandafter\def\csname N@gss-gemma-pool-sel-pwinphrase\endcsname{all}
\expandafter\def\csname N@gss-gemma-pool-shp-corr\endcsname{1.68--2.50}
\expandafter\def\csname N@gss-gemma-pool-shp-rel\endcsname{1.01--1.30}
\expandafter\def\csname N@gss-gemma-pool-shp-n\endcsname{4}
\expandafter\def\csname N@gss-gemma-pool-shp-nworse\endcsname{4}
\expandafter\def\csname N@gss-gemma-pool-shp-base-corr\endcsname{2.17}
\expandafter\def\csname N@gss-gemma-pool-shp-base-rel\endcsname{1.24}
\expandafter\def\csname N@gss-gemma-pool-shp-base-cross\endcsname{2.05}
\expandafter\def\csname N@gss-gemma-pool-shp-base-d\endcsname{$-$1.09}
\expandafter\def\csname N@gss-gemma-pool-shp-base-ci\endcsname{$[-1.35, -0.84]$}
\expandafter\def\csname N@gss-gemma-pool-shp-base-pwin\endcsname{100\%}
\expandafter\def\csname N@gss-gemma-pool-shp-base-edp\endcsname{0.259}
\expandafter\def\csname N@gss-gemma-pool-shp-base-edpci\endcsname{$[-0.288, -0.108]$}
\expandafter\def\csname N@gss-gemma-pool-shp-div-corr\endcsname{1.68}
\expandafter\def\csname N@gss-gemma-pool-shp-div-rel\endcsname{1.01}
\expandafter\def\csname N@gss-gemma-pool-shp-div-cross\endcsname{1.38}
\expandafter\def\csname N@gss-gemma-pool-shp-div-d\endcsname{$-$0.60}
\expandafter\def\csname N@gss-gemma-pool-shp-div-ci\endcsname{$[-0.84, -0.38]$}
\expandafter\def\csname N@gss-gemma-pool-shp-div-pwin\endcsname{100\%}
\expandafter\def\csname N@gss-gemma-pool-shp-div-edp\endcsname{0.205}
\expandafter\def\csname N@gss-gemma-pool-shp-div-edpci\endcsname{$[-0.220, -0.064]$}
\expandafter\def\csname N@gss-gemma-pool-shp-argyle-corr\endcsname{2.50}
\expandafter\def\csname N@gss-gemma-pool-shp-argyle-rel\endcsname{1.30}
\expandafter\def\csname N@gss-gemma-pool-shp-argyle-cross\endcsname{2.38}
\expandafter\def\csname N@gss-gemma-pool-shp-argyle-d\endcsname{$-$1.41}
\expandafter\def\csname N@gss-gemma-pool-shp-argyle-ci\endcsname{$[-1.68, -1.15]$}
\expandafter\def\csname N@gss-gemma-pool-shp-argyle-pwin\endcsname{100\%}
\expandafter\def\csname N@gss-gemma-pool-shp-argyle-edp\endcsname{0.494}
\expandafter\def\csname N@gss-gemma-pool-shp-argyle-edpci\endcsname{$[-0.524, -0.290]$}
\expandafter\def\csname N@gss-gemma-pool-shp-story-corr\endcsname{2.17}
\expandafter\def\csname N@gss-gemma-pool-shp-story-rel\endcsname{1.27}
\expandafter\def\csname N@gss-gemma-pool-shp-story-cross\endcsname{2.23}
\expandafter\def\csname N@gss-gemma-pool-shp-story-d\endcsname{$-$1.08}
\expandafter\def\csname N@gss-gemma-pool-shp-story-ci\endcsname{$[-1.37, -0.80]$}
\expandafter\def\csname N@gss-gemma-pool-shp-story-pwin\endcsname{100\%}
\expandafter\def\csname N@gss-gemma-pool-shp-story-edp\endcsname{0.239}
\expandafter\def\csname N@gss-gemma-pool-shp-story-edpci\endcsname{$[-0.259, -0.088]$}
\expandafter\def\csname N@gss-gemma-pool-shp-drange\endcsname{$-$1.41 to $-$0.60}
\expandafter\def\csname N@gss-gemma-pool-shp-nsig\endcsname{4}
\expandafter\def\csname N@gss-gemma-pool-shp-relx\endcsname{1.0--1.2}
\expandafter\def\csname N@gss-gemma-pool-shp-pwinmin\endcsname{100\%}
\expandafter\def\csname N@gss-gemma-pool-shp-pwinphrase\endcsname{all}
\expandafter\def\csname N@gss-gemma-pool-abl-k16-corr\endcsname{1.15}
\expandafter\def\csname N@gss-gemma-pool-abl-k16-cross\endcsname{0.67}
\expandafter\def\csname N@gss-gemma-pool-abl-k16-pwin\endcsname{72.0\%}
\expandafter\def\csname N@gss-gemma-pool-abl-k16-gap\endcsname{0.70}
\expandafter\def\csname N@gss-gemma-pool-abl-k16-dcross\endcsname{0.16}
\expandafter\def\csname N@gss-gemma-pool-abl-k16-dcross-ci\endcsname{$[-0.08, 0.40]$}
\expandafter\def\csname N@gss-gemma-pool-abl-k16-dgap\endcsname{0.06}
\expandafter\def\csname N@gss-gemma-pool-abl-k16-dgap-ci\endcsname{$[-0.17, 0.29]$}
\expandafter\def\csname N@gss-gemma-pool-abl-k16-dcorr\endcsname{$-$0.07}
\expandafter\def\csname N@gss-gemma-pool-abl-k16-dcorr-ci\endcsname{$[-0.31, 0.14]$}
\expandafter\def\csname N@gss-gemma-pool-abl-k16-rel\endcsname{1.03}
\expandafter\def\csname N@gss-gemma-pool-abl-nosharp-corr\endcsname{1.07}
\expandafter\def\csname N@gss-gemma-pool-abl-nosharp-cross\endcsname{0.84}
\expandafter\def\csname N@gss-gemma-pool-abl-nosharp-pwin\endcsname{44.5\%}
\expandafter\def\csname N@gss-gemma-pool-abl-nosharp-gap\endcsname{0.76}
\expandafter\def\csname N@gss-gemma-pool-abl-nosharp-dcross\endcsname{$-$0.01}
\expandafter\def\csname N@gss-gemma-pool-abl-nosharp-dcross-ci\endcsname{$[-0.17, 0.15]$}
\expandafter\def\csname N@gss-gemma-pool-abl-nosharp-dgap\endcsname{0.00}
\expandafter\def\csname N@gss-gemma-pool-abl-nosharp-dgap-ci\endcsname{$[-0.09, 0.13]$}
\expandafter\def\csname N@gss-gemma-pool-abl-nosharp-dcorr\endcsname{0.02}
\expandafter\def\csname N@gss-gemma-pool-abl-nosharp-dcorr-ci\endcsname{$[-0.17, 0.21]$}
\expandafter\def\csname N@gss-gemma-pool-abl-nosharp-rel\endcsname{1.21}
\expandafter\def\csname N@gss-gemma-pool-agg-corr\endcsname{1.74}
\expandafter\def\csname N@gss-gemma-pool-agg-cross\endcsname{0.40}
\expandafter\def\csname N@gss-gemma-pool-agg-pwin\endcsname{100.0\%}
\expandafter\def\csname N@gss-gemma-pool-agg-gap\endcsname{0.19}
\expandafter\def\csname N@gss-gemma-pool-agg-dcross\endcsname{0.44}
\expandafter\def\csname N@gss-gemma-pool-agg-dcross-ci\endcsname{$[0.23, 0.67]$}
\expandafter\def\csname N@gss-gemma-pool-agg-dgap\endcsname{0.56}
\expandafter\def\csname N@gss-gemma-pool-agg-dgap-ci\endcsname{$[0.35, 0.78]$}
\expandafter\def\csname N@gss-gemma-pool-agg-dcorr\endcsname{$-$0.66}
\expandafter\def\csname N@gss-gemma-pool-agg-dcorr-ci\endcsname{$[-0.90, -0.37]$}
\expandafter\def\csname N@gss-gemma-pool-agg-rel\endcsname{0.00}
\expandafter\def\csname N@gss-gemma-pool-abl-seed2-corr\endcsname{1.08}
\expandafter\def\csname N@gss-gemma-pool-abl-seed2-cross\endcsname{0.84}
\expandafter\def\csname N@gss-gemma-pool-abl-seed2-pwin\endcsname{54.5\%}
\expandafter\def\csname N@gss-gemma-pool-abl-seed2-gap\endcsname{0.79}
\expandafter\def\csname N@gss-gemma-pool-abl-seed2-dcross\endcsname{$-$0.01}
\expandafter\def\csname N@gss-gemma-pool-abl-seed2-dcross-ci\endcsname{$[-0.26, 0.31]$}
\expandafter\def\csname N@gss-gemma-pool-abl-seed2-dgap\endcsname{$-$0.03}
\expandafter\def\csname N@gss-gemma-pool-abl-seed2-dgap-ci\endcsname{$[-0.25, 0.21]$}
\expandafter\def\csname N@gss-gemma-pool-abl-seed2-dcorr\endcsname{0.01}
\expandafter\def\csname N@gss-gemma-pool-abl-seed2-dcorr-ci\endcsname{$[-0.17, 0.24]$}
\expandafter\def\csname N@gss-gemma-pool-abl-seed2-rel\endcsname{1.26}
\expandafter\def\csname N@gss-gemma-pool-abl-demsharp-corr\endcsname{2.17}
\expandafter\def\csname N@gss-gemma-pool-abl-demsharp-cross\endcsname{2.04}
\expandafter\def\csname N@gss-gemma-pool-abl-demsharp-pwin\endcsname{100.0\%}
\expandafter\def\csname N@gss-gemma-pool-abl-demsharp-gap\endcsname{1.47}
\expandafter\def\csname N@gss-gemma-pool-abl-demsharp-dcross\endcsname{$-$1.21}
\expandafter\def\csname N@gss-gemma-pool-abl-demsharp-dcross-ci\endcsname{$[-1.51, -0.94]$}
\expandafter\def\csname N@gss-gemma-pool-abl-demsharp-dgap\endcsname{$-$0.71}
\expandafter\def\csname N@gss-gemma-pool-abl-demsharp-dgap-ci\endcsname{$[-0.93, -0.50]$}
\expandafter\def\csname N@gss-gemma-pool-abl-demsharp-dcorr\endcsname{$-$1.09}
\expandafter\def\csname N@gss-gemma-pool-abl-demsharp-dcorr-ci\endcsname{$[-1.32, -0.83]$}
\expandafter\def\csname N@gss-gemma-pool-abl-demsharp-rel\endcsname{1.23}
\expandafter\def\csname N@gss-gemma-pool-demo-corr\endcsname{2.16}
\expandafter\def\csname N@gss-gemma-pool-demo-cross\endcsname{2.04}
\expandafter\def\csname N@gss-gemma-pool-demo-pwin\endcsname{100.0\%}
\expandafter\def\csname N@gss-gemma-pool-demo-gap\endcsname{1.48}
\expandafter\def\csname N@gss-gemma-pool-demo-dcross\endcsname{$-$1.21}
\expandafter\def\csname N@gss-gemma-pool-demo-dcross-ci\endcsname{$[-1.50, -0.93]$}
\expandafter\def\csname N@gss-gemma-pool-demo-dgap\endcsname{$-$0.72}
\expandafter\def\csname N@gss-gemma-pool-demo-dgap-ci\endcsname{$[-0.92, -0.51]$}
\expandafter\def\csname N@gss-gemma-pool-demo-dcorr\endcsname{$-$1.08}
\expandafter\def\csname N@gss-gemma-pool-demo-dcorr-ci\endcsname{$[-1.36, -0.80]$}
\expandafter\def\csname N@gss-gemma-pool-demo-rel\endcsname{1.19}
\expandafter\def\csname N@hm-qwen-otp-floor-corr\endcsname{0.26}
\expandafter\def\csname N@hm-qwen-otp-floor-cross\endcsname{0.98}
\expandafter\def\csname N@hm-qwen-otp-lpp-corr\endcsname{0.57}
\expandafter\def\csname N@hm-qwen-otp-lpp-rel\endcsname{1.06}
\expandafter\def\csname N@hm-qwen-otp-lpp-cross\endcsname{0.91}
\expandafter\def\csname N@hm-qwen-otp-lpp-gap\endcsname{1.09}
\expandafter\def\csname N@hm-qwen-otp-lpp-wpsd\endcsname{1.00}
\expandafter\def\csname N@hm-qwen-otp-lpp-edp\endcsname{0.011}
\expandafter\def\csname N@hm-qwen-otp-sel-corr\endcsname{0.57--0.69}
\expandafter\def\csname N@hm-qwen-otp-sel-rel\endcsname{0.36--0.61}
\expandafter\def\csname N@hm-qwen-otp-sel-n\endcsname{7}
\expandafter\def\csname N@hm-qwen-otp-sel-nworse\endcsname{7}
\expandafter\def\csname N@hm-qwen-otp-sel-divbest-corr\endcsname{0.61}
\expandafter\def\csname N@hm-qwen-otp-sel-divbest-rel\endcsname{0.48}
\expandafter\def\csname N@hm-qwen-otp-sel-divbest-cross\endcsname{1.00}
\expandafter\def\csname N@hm-qwen-otp-sel-divbest-d\endcsname{$-$0.05}
\expandafter\def\csname N@hm-qwen-otp-sel-divbest-ci\endcsname{$[-0.19, 0.13]$}
\expandafter\def\csname N@hm-qwen-otp-sel-divbest-pwin\endcsname{73\%}
\expandafter\def\csname N@hm-qwen-otp-sel-divbest-edp\endcsname{0.019}
\expandafter\def\csname N@hm-qwen-otp-sel-divbest-edpci\endcsname{$[-0.029, 0.023]$}
\expandafter\def\csname N@hm-qwen-otp-sel-rw-corr\endcsname{0.61}
\expandafter\def\csname N@hm-qwen-otp-sel-rw-rel\endcsname{0.56}
\expandafter\def\csname N@hm-qwen-otp-sel-rw-cross\endcsname{1.13}
\expandafter\def\csname N@hm-qwen-otp-sel-rw-d\endcsname{$-$0.04}
\expandafter\def\csname N@hm-qwen-otp-sel-rw-ci\endcsname{$[-0.24, 0.14]$}
\expandafter\def\csname N@hm-qwen-otp-sel-rw-pwin\endcsname{67\%}
\expandafter\def\csname N@hm-qwen-otp-sel-rw-edp\endcsname{0.021}
\expandafter\def\csname N@hm-qwen-otp-sel-rw-edpci\endcsname{$[-0.040, 0.019]$}
\expandafter\def\csname N@hm-qwen-otp-sel-base-corr\endcsname{0.58}
\expandafter\def\csname N@hm-qwen-otp-sel-base-rel\endcsname{0.39}
\expandafter\def\csname N@hm-qwen-otp-sel-base-cross\endcsname{0.79}
\expandafter\def\csname N@hm-qwen-otp-sel-base-d\endcsname{$-$0.02}
\expandafter\def\csname N@hm-qwen-otp-sel-base-ci\endcsname{$[-0.22, 0.14]$}
\expandafter\def\csname N@hm-qwen-otp-sel-base-pwin\endcsname{57\%}
\expandafter\def\csname N@hm-qwen-otp-sel-base-edp\endcsname{0.018}
\expandafter\def\csname N@hm-qwen-otp-sel-base-edpci\endcsname{$[-0.031, 0.016]$}
\expandafter\def\csname N@hm-qwen-otp-sel-div-corr\endcsname{0.58}
\expandafter\def\csname N@hm-qwen-otp-sel-div-rel\endcsname{0.49}
\expandafter\def\csname N@hm-qwen-otp-sel-div-cross\endcsname{0.99}
\expandafter\def\csname N@hm-qwen-otp-sel-div-d\endcsname{$-$0.01}
\expandafter\def\csname N@hm-qwen-otp-sel-div-ci\endcsname{$[-0.18, 0.16]$}
\expandafter\def\csname N@hm-qwen-otp-sel-div-pwin\endcsname{55\%}
\expandafter\def\csname N@hm-qwen-otp-sel-div-edp\endcsname{0.018}
\expandafter\def\csname N@hm-qwen-otp-sel-div-edpci\endcsname{$[-0.029, 0.019]$}
\expandafter\def\csname N@hm-qwen-otp-sel-argyle-corr\endcsname{0.61}
\expandafter\def\csname N@hm-qwen-otp-sel-argyle-rel\endcsname{0.36}
\expandafter\def\csname N@hm-qwen-otp-sel-argyle-cross\endcsname{0.72}
\expandafter\def\csname N@hm-qwen-otp-sel-argyle-d\endcsname{$-$0.04}
\expandafter\def\csname N@hm-qwen-otp-sel-argyle-ci\endcsname{$[-0.22, 0.11]$}
\expandafter\def\csname N@hm-qwen-otp-sel-argyle-pwin\endcsname{64\%}
\expandafter\def\csname N@hm-qwen-otp-sel-argyle-edp\endcsname{0.020}
\expandafter\def\csname N@hm-qwen-otp-sel-argyle-edpci\endcsname{$[-0.040, 0.017]$}
\expandafter\def\csname N@hm-qwen-otp-sel-story-corr\endcsname{0.57}
\expandafter\def\csname N@hm-qwen-otp-sel-story-rel\endcsname{0.45}
\expandafter\def\csname N@hm-qwen-otp-sel-story-cross\endcsname{0.88}
\expandafter\def\csname N@hm-qwen-otp-sel-story-d\endcsname{$-$0.00}
\expandafter\def\csname N@hm-qwen-otp-sel-story-ci\endcsname{$[-0.15, 0.19]$}
\expandafter\def\csname N@hm-qwen-otp-sel-story-pwin\endcsname{49\%}
\expandafter\def\csname N@hm-qwen-otp-sel-story-edp\endcsname{0.016}
\expandafter\def\csname N@hm-qwen-otp-sel-story-edpci\endcsname{$[-0.027, 0.026]$}
\expandafter\def\csname N@hm-qwen-otp-sel-anth-corr\endcsname{0.69}
\expandafter\def\csname N@hm-qwen-otp-sel-anth-rel\endcsname{0.61}
\expandafter\def\csname N@hm-qwen-otp-sel-anth-cross\endcsname{0.70}
\expandafter\def\csname N@hm-qwen-otp-sel-anth-d\endcsname{$-$0.13}
\expandafter\def\csname N@hm-qwen-otp-sel-anth-ci\endcsname{$[-0.31, 0.03]$}
\expandafter\def\csname N@hm-qwen-otp-sel-anth-pwin\endcsname{92\%}
\expandafter\def\csname N@hm-qwen-otp-sel-anth-edp\endcsname{0.023}
\expandafter\def\csname N@hm-qwen-otp-sel-anth-edpci\endcsname{$[-0.032, 0.014]$}
\expandafter\def\csname N@hm-qwen-otp-sel-drange\endcsname{$-$0.13 to $-$0.00}
\expandafter\def\csname N@hm-qwen-otp-sel-nsig\endcsname{0}
\expandafter\def\csname N@hm-qwen-otp-sel-relx\endcsname{1.7--2.9}
\expandafter\def\csname N@hm-qwen-otp-sel-pwinmin\endcsname{49\%}
\expandafter\def\csname N@hm-qwen-otp-sel-pwinphrase\endcsname{at least 49\% of}
\expandafter\def\csname N@hm-qwen-otp-shp-corr\endcsname{0.71--0.97}
\expandafter\def\csname N@hm-qwen-otp-shp-rel\endcsname{0.79--0.94}
\expandafter\def\csname N@hm-qwen-otp-shp-n\endcsname{4}
\expandafter\def\csname N@hm-qwen-otp-shp-nworse\endcsname{4}
\expandafter\def\csname N@hm-qwen-otp-shp-base-corr\endcsname{0.97}
\expandafter\def\csname N@hm-qwen-otp-shp-base-rel\endcsname{0.94}
\expandafter\def\csname N@hm-qwen-otp-shp-base-cross\endcsname{1.63}
\expandafter\def\csname N@hm-qwen-otp-shp-base-d\endcsname{$-$0.40}
\expandafter\def\csname N@hm-qwen-otp-shp-base-ci\endcsname{$[-0.58, -0.22]$}
\expandafter\def\csname N@hm-qwen-otp-shp-base-pwin\endcsname{100\%}
\expandafter\def\csname N@hm-qwen-otp-shp-base-edp\endcsname{0.043}
\expandafter\def\csname N@hm-qwen-otp-shp-base-edpci\endcsname{$[-0.056, -0.007]$}
\expandafter\def\csname N@hm-qwen-otp-shp-div-corr\endcsname{0.79}
\expandafter\def\csname N@hm-qwen-otp-shp-div-rel\endcsname{0.83}
\expandafter\def\csname N@hm-qwen-otp-shp-div-cross\endcsname{1.38}
\expandafter\def\csname N@hm-qwen-otp-shp-div-d\endcsname{$-$0.23}
\expandafter\def\csname N@hm-qwen-otp-shp-div-ci\endcsname{$[-0.40, -0.06]$}
\expandafter\def\csname N@hm-qwen-otp-shp-div-pwin\endcsname{99\%}
\expandafter\def\csname N@hm-qwen-otp-shp-div-edp\endcsname{0.037}
\expandafter\def\csname N@hm-qwen-otp-shp-div-edpci\endcsname{$[-0.050, -0.002]$}
\expandafter\def\csname N@hm-qwen-otp-shp-argyle-corr\endcsname{0.74}
\expandafter\def\csname N@hm-qwen-otp-shp-argyle-rel\endcsname{0.79}
\expandafter\def\csname N@hm-qwen-otp-shp-argyle-cross\endcsname{1.25}
\expandafter\def\csname N@hm-qwen-otp-shp-argyle-d\endcsname{$-$0.17}
\expandafter\def\csname N@hm-qwen-otp-shp-argyle-ci\endcsname{$[-0.36, 0.01]$}
\expandafter\def\csname N@hm-qwen-otp-shp-argyle-pwin\endcsname{96\%}
\expandafter\def\csname N@hm-qwen-otp-shp-argyle-edp\endcsname{0.043}
\expandafter\def\csname N@hm-qwen-otp-shp-argyle-edpci\endcsname{$[-0.059, -0.004]$}
\expandafter\def\csname N@hm-qwen-otp-shp-story-corr\endcsname{0.71}
\expandafter\def\csname N@hm-qwen-otp-shp-story-rel\endcsname{0.92}
\expandafter\def\csname N@hm-qwen-otp-shp-story-cross\endcsname{1.38}
\expandafter\def\csname N@hm-qwen-otp-shp-story-d\endcsname{$-$0.14}
\expandafter\def\csname N@hm-qwen-otp-shp-story-ci\endcsname{$[-0.34, 0.07]$}
\expandafter\def\csname N@hm-qwen-otp-shp-story-pwin\endcsname{92\%}
\expandafter\def\csname N@hm-qwen-otp-shp-story-edp\endcsname{0.022}
\expandafter\def\csname N@hm-qwen-otp-shp-story-edpci\endcsname{$[-0.032, 0.013]$}
\expandafter\def\csname N@hm-qwen-otp-shp-drange\endcsname{$-$0.40 to $-$0.14}
\expandafter\def\csname N@hm-qwen-otp-shp-nsig\endcsname{2}
\expandafter\def\csname N@hm-qwen-otp-shp-relx\endcsname{1.1--1.3}
\expandafter\def\csname N@hm-qwen-otp-shp-pwinmin\endcsname{92\%}
\expandafter\def\csname N@hm-qwen-otp-shp-pwinphrase\endcsname{at least 92\% of}
\expandafter\def\csname N@hm-qwen-otp-abl-k16-corr\endcsname{0.58}
\expandafter\def\csname N@hm-qwen-otp-abl-k16-cross\endcsname{0.58}
\expandafter\def\csname N@hm-qwen-otp-abl-k16-pwin\endcsname{56.5\%}
\expandafter\def\csname N@hm-qwen-otp-abl-k16-gap\endcsname{1.10}
\expandafter\def\csname N@hm-qwen-otp-abl-k16-dcross\endcsname{0.34}
\expandafter\def\csname N@hm-qwen-otp-abl-k16-dcross-ci\endcsname{$[0.10, 0.56]$}
\expandafter\def\csname N@hm-qwen-otp-abl-k16-dgap\endcsname{$-$0.00}
\expandafter\def\csname N@hm-qwen-otp-abl-k16-dgap-ci\endcsname{$[-0.21, 0.18]$}
\expandafter\def\csname N@hm-qwen-otp-abl-k16-dcorr\endcsname{$-$0.02}
\expandafter\def\csname N@hm-qwen-otp-abl-k16-dcorr-ci\endcsname{$[-0.20, 0.14]$}
\expandafter\def\csname N@hm-qwen-otp-abl-k16-rel\endcsname{0.97}
\expandafter\def\csname N@hm-qwen-otp-abl-nosharp-corr\endcsname{0.57}
\expandafter\def\csname N@hm-qwen-otp-abl-nosharp-cross\endcsname{0.78}
\expandafter\def\csname N@hm-qwen-otp-abl-nosharp-pwin\endcsname{49.5\%}
\expandafter\def\csname N@hm-qwen-otp-abl-nosharp-gap\endcsname{1.09}
\expandafter\def\csname N@hm-qwen-otp-abl-nosharp-dcross\endcsname{0.13}
\expandafter\def\csname N@hm-qwen-otp-abl-nosharp-dcross-ci\endcsname{$[-0.04, 0.32]$}
\expandafter\def\csname N@hm-qwen-otp-abl-nosharp-dgap\endcsname{0.00}
\expandafter\def\csname N@hm-qwen-otp-abl-nosharp-dgap-ci\endcsname{$[-0.13, 0.14]$}
\expandafter\def\csname N@hm-qwen-otp-abl-nosharp-dcorr\endcsname{$-$0.00}
\expandafter\def\csname N@hm-qwen-otp-abl-nosharp-dcorr-ci\endcsname{$[-0.14, 0.13]$}
\expandafter\def\csname N@hm-qwen-otp-abl-nosharp-rel\endcsname{0.76}
\expandafter\def\csname N@hm-qwen-otp-agg-corr\endcsname{0.78}
\expandafter\def\csname N@hm-qwen-otp-agg-cross\endcsname{0.45}
\expandafter\def\csname N@hm-qwen-otp-agg-pwin\endcsname{98.0\%}
\expandafter\def\csname N@hm-qwen-otp-agg-gap\endcsname{1.10}
\expandafter\def\csname N@hm-qwen-otp-agg-dcross\endcsname{0.46}
\expandafter\def\csname N@hm-qwen-otp-agg-dcross-ci\endcsname{$[0.24, 0.69]$}
\expandafter\def\csname N@hm-qwen-otp-agg-dgap\endcsname{$-$0.01}
\expandafter\def\csname N@hm-qwen-otp-agg-dgap-ci\endcsname{$[-0.20, 0.20]$}
\expandafter\def\csname N@hm-qwen-otp-agg-dcorr\endcsname{$-$0.22}
\expandafter\def\csname N@hm-qwen-otp-agg-dcorr-ci\endcsname{$[-0.40, -0.04]$}
\expandafter\def\csname N@hm-qwen-otp-agg-rel\endcsname{0.21}
\expandafter\def\csname N@hm-qwen-otp-abl-seed2-corr\endcsname{0.58}
\expandafter\def\csname N@hm-qwen-otp-abl-seed2-cross\endcsname{0.96}
\expandafter\def\csname N@hm-qwen-otp-abl-seed2-pwin\endcsname{61.0\%}
\expandafter\def\csname N@hm-qwen-otp-abl-seed2-gap\endcsname{1.10}
\expandafter\def\csname N@hm-qwen-otp-abl-seed2-dcross\endcsname{$-$0.05}
\expandafter\def\csname N@hm-qwen-otp-abl-seed2-dcross-ci\endcsname{$[-0.26, 0.17]$}
\expandafter\def\csname N@hm-qwen-otp-abl-seed2-dgap\endcsname{$-$0.01}
\expandafter\def\csname N@hm-qwen-otp-abl-seed2-dgap-ci\endcsname{$[-0.21, 0.19]$}
\expandafter\def\csname N@hm-qwen-otp-abl-seed2-dcorr\endcsname{$-$0.02}
\expandafter\def\csname N@hm-qwen-otp-abl-seed2-dcorr-ci\endcsname{$[-0.17, 0.14]$}
\expandafter\def\csname N@hm-qwen-otp-abl-seed2-rel\endcsname{1.04}
\expandafter\def\csname N@hm-qwen-otp-abl-demsharp-corr\endcsname{0.96}
\expandafter\def\csname N@hm-qwen-otp-abl-demsharp-cross\endcsname{1.62}
\expandafter\def\csname N@hm-qwen-otp-abl-demsharp-pwin\endcsname{100.0\%}
\expandafter\def\csname N@hm-qwen-otp-abl-demsharp-gap\endcsname{1.09}
\expandafter\def\csname N@hm-qwen-otp-abl-demsharp-dcross\endcsname{$-$0.71}
\expandafter\def\csname N@hm-qwen-otp-abl-demsharp-dcross-ci\endcsname{$[-0.95, -0.49]$}
\expandafter\def\csname N@hm-qwen-otp-abl-demsharp-dgap\endcsname{0.00}
\expandafter\def\csname N@hm-qwen-otp-abl-demsharp-dgap-ci\endcsname{$[-0.20, 0.21]$}
\expandafter\def\csname N@hm-qwen-otp-abl-demsharp-dcorr\endcsname{$-$0.39}
\expandafter\def\csname N@hm-qwen-otp-abl-demsharp-dcorr-ci\endcsname{$[-0.56, -0.19]$}
\expandafter\def\csname N@hm-qwen-otp-abl-demsharp-rel\endcsname{0.87}
\expandafter\def\csname N@hm-qwen-otp-demo-corr\endcsname{0.62}
\expandafter\def\csname N@hm-qwen-otp-demo-cross\endcsname{1.13}
\expandafter\def\csname N@hm-qwen-otp-demo-pwin\endcsname{69.0\%}
\expandafter\def\csname N@hm-qwen-otp-demo-gap\endcsname{1.10}
\expandafter\def\csname N@hm-qwen-otp-demo-dcross\endcsname{$-$0.22}
\expandafter\def\csname N@hm-qwen-otp-demo-dcross-ci\endcsname{$[-0.45, 0.05]$}
\expandafter\def\csname N@hm-qwen-otp-demo-dgap\endcsname{$-$0.00}
\expandafter\def\csname N@hm-qwen-otp-demo-dgap-ci\endcsname{$[-0.18, 0.18]$}
\expandafter\def\csname N@hm-qwen-otp-demo-dcorr\endcsname{$-$0.05}
\expandafter\def\csname N@hm-qwen-otp-demo-dcorr-ci\endcsname{$[-0.25, 0.13]$}
\expandafter\def\csname N@hm-qwen-otp-demo-rel\endcsname{0.56}
\expandafter\def\csname N@hm-qwen-pool-floor-corr\endcsname{0.26}
\expandafter\def\csname N@hm-qwen-pool-floor-cross\endcsname{0.98}
\expandafter\def\csname N@hm-qwen-pool-lpp-corr\endcsname{0.65}
\expandafter\def\csname N@hm-qwen-pool-lpp-rel\endcsname{1.02}
\expandafter\def\csname N@hm-qwen-pool-lpp-cross\endcsname{0.71}
\expandafter\def\csname N@hm-qwen-pool-lpp-gap\endcsname{0.66}
\expandafter\def\csname N@hm-qwen-pool-lpp-wpsd\endcsname{1.07}
\expandafter\def\csname N@hm-qwen-pool-lpp-edp\endcsname{0.058}
\expandafter\def\csname N@hm-qwen-pool-sel-corr\endcsname{0.55--0.85}
\expandafter\def\csname N@hm-qwen-pool-sel-rel\endcsname{0.44--0.68}
\expandafter\def\csname N@hm-qwen-pool-sel-n\endcsname{7}
\expandafter\def\csname N@hm-qwen-pool-sel-nworse\endcsname{3}
\expandafter\def\csname N@hm-qwen-pool-sel-divbest-corr\endcsname{0.73}
\expandafter\def\csname N@hm-qwen-pool-sel-divbest-rel\endcsname{0.59}
\expandafter\def\csname N@hm-qwen-pool-sel-divbest-cross\endcsname{1.30}
\expandafter\def\csname N@hm-qwen-pool-sel-divbest-d\endcsname{$-$0.08}
\expandafter\def\csname N@hm-qwen-pool-sel-divbest-ci\endcsname{$[-0.26, 0.11]$}
\expandafter\def\csname N@hm-qwen-pool-sel-divbest-pwin\endcsname{80\%}
\expandafter\def\csname N@hm-qwen-pool-sel-divbest-edp\endcsname{0.069}
\expandafter\def\csname N@hm-qwen-pool-sel-divbest-edpci\endcsname{$[-0.065, 0.047]$}
\expandafter\def\csname N@hm-qwen-pool-sel-rw-corr\endcsname{0.76}
\expandafter\def\csname N@hm-qwen-pool-sel-rw-rel\endcsname{0.68}
\expandafter\def\csname N@hm-qwen-pool-sel-rw-cross\endcsname{1.41}
\expandafter\def\csname N@hm-qwen-pool-sel-rw-d\endcsname{$-$0.11}
\expandafter\def\csname N@hm-qwen-pool-sel-rw-ci\endcsname{$[-0.31, 0.12]$}
\expandafter\def\csname N@hm-qwen-pool-sel-rw-pwin\endcsname{84\%}
\expandafter\def\csname N@hm-qwen-pool-sel-rw-edp\endcsname{0.082}
\expandafter\def\csname N@hm-qwen-pool-sel-rw-edpci\endcsname{$[-0.104, 0.049]$}
\expandafter\def\csname N@hm-qwen-pool-sel-base-corr\endcsname{0.56}
\expandafter\def\csname N@hm-qwen-pool-sel-base-rel\endcsname{0.44}
\expandafter\def\csname N@hm-qwen-pool-sel-base-cross\endcsname{0.86}
\expandafter\def\csname N@hm-qwen-pool-sel-base-d\endcsname{0.09}
\expandafter\def\csname N@hm-qwen-pool-sel-base-ci\endcsname{$[-0.09, 0.26]$}
\expandafter\def\csname N@hm-qwen-pool-sel-base-pwin\endcsname{16\%}
\expandafter\def\csname N@hm-qwen-pool-sel-base-edp\endcsname{0.037}
\expandafter\def\csname N@hm-qwen-pool-sel-base-edpci\endcsname{$[-0.028, 0.076]$}
\expandafter\def\csname N@hm-qwen-pool-sel-div-corr\endcsname{0.63}
\expandafter\def\csname N@hm-qwen-pool-sel-div-rel\endcsname{0.60}
\expandafter\def\csname N@hm-qwen-pool-sel-div-cross\endcsname{1.18}
\expandafter\def\csname N@hm-qwen-pool-sel-div-d\endcsname{0.02}
\expandafter\def\csname N@hm-qwen-pool-sel-div-ci\endcsname{$[-0.19, 0.25]$}
\expandafter\def\csname N@hm-qwen-pool-sel-div-pwin\endcsname{45\%}
\expandafter\def\csname N@hm-qwen-pool-sel-div-edp\endcsname{0.066}
\expandafter\def\csname N@hm-qwen-pool-sel-div-edpci\endcsname{$[-0.070, 0.050]$}
\expandafter\def\csname N@hm-qwen-pool-sel-argyle-corr\endcsname{0.61}
\expandafter\def\csname N@hm-qwen-pool-sel-argyle-rel\endcsname{0.57}
\expandafter\def\csname N@hm-qwen-pool-sel-argyle-cross\endcsname{1.15}
\expandafter\def\csname N@hm-qwen-pool-sel-argyle-d\endcsname{0.04}
\expandafter\def\csname N@hm-qwen-pool-sel-argyle-ci\endcsname{$[-0.13, 0.27]$}
\expandafter\def\csname N@hm-qwen-pool-sel-argyle-pwin\endcsname{33\%}
\expandafter\def\csname N@hm-qwen-pool-sel-argyle-edp\endcsname{0.115}
\expandafter\def\csname N@hm-qwen-pool-sel-argyle-edpci\endcsname{$[-0.130, 0.018]$}
\expandafter\def\csname N@hm-qwen-pool-sel-story-corr\endcsname{0.55}
\expandafter\def\csname N@hm-qwen-pool-sel-story-rel\endcsname{0.50}
\expandafter\def\csname N@hm-qwen-pool-sel-story-cross\endcsname{0.97}
\expandafter\def\csname N@hm-qwen-pool-sel-story-d\endcsname{0.09}
\expandafter\def\csname N@hm-qwen-pool-sel-story-ci\endcsname{$[-0.08, 0.28]$}
\expandafter\def\csname N@hm-qwen-pool-sel-story-pwin\endcsname{16\%}
\expandafter\def\csname N@hm-qwen-pool-sel-story-edp\endcsname{0.037}
\expandafter\def\csname N@hm-qwen-pool-sel-story-edpci\endcsname{$[-0.035, 0.079]$}
\expandafter\def\csname N@hm-qwen-pool-sel-anth-corr\endcsname{0.85}
\expandafter\def\csname N@hm-qwen-pool-sel-anth-rel\endcsname{0.56}
\expandafter\def\csname N@hm-qwen-pool-sel-anth-cross\endcsname{0.46}
\expandafter\def\csname N@hm-qwen-pool-sel-anth-d\endcsname{$-$0.21}
\expandafter\def\csname N@hm-qwen-pool-sel-anth-ci\endcsname{$[-0.39, -0.05]$}
\expandafter\def\csname N@hm-qwen-pool-sel-anth-pwin\endcsname{100\%}
\expandafter\def\csname N@hm-qwen-pool-sel-anth-edp\endcsname{0.104}
\expandafter\def\csname N@hm-qwen-pool-sel-anth-edpci\endcsname{$[-0.103, 0.007]$}
\expandafter\def\csname N@hm-qwen-pool-sel-drange\endcsname{$-$0.21 to +0.09}
\expandafter\def\csname N@hm-qwen-pool-sel-nsig\endcsname{1}
\expandafter\def\csname N@hm-qwen-pool-sel-relx\endcsname{1.5--2.3}
\expandafter\def\csname N@hm-qwen-pool-sel-pwinmin\endcsname{16\%}
\expandafter\def\csname N@hm-qwen-pool-sel-pwinphrase\endcsname{at least 16\% of}
\expandafter\def\csname N@hm-qwen-pool-shp-corr\endcsname{1.05--1.33}
\expandafter\def\csname N@hm-qwen-pool-shp-rel\endcsname{0.97--1.06}
\expandafter\def\csname N@hm-qwen-pool-shp-n\endcsname{4}
\expandafter\def\csname N@hm-qwen-pool-shp-nworse\endcsname{4}
\expandafter\def\csname N@hm-qwen-pool-shp-base-corr\endcsname{1.32}
\expandafter\def\csname N@hm-qwen-pool-shp-base-rel\endcsname{1.06}
\expandafter\def\csname N@hm-qwen-pool-shp-base-cross\endcsname{2.10}
\expandafter\def\csname N@hm-qwen-pool-shp-base-d\endcsname{$-$0.68}
\expandafter\def\csname N@hm-qwen-pool-shp-base-ci\endcsname{$[-0.85, -0.44]$}
\expandafter\def\csname N@hm-qwen-pool-shp-base-pwin\endcsname{100\%}
\expandafter\def\csname N@hm-qwen-pool-shp-base-edp\endcsname{0.205}
\expandafter\def\csname N@hm-qwen-pool-shp-base-edpci\endcsname{$[-0.228, -0.069]$}
\expandafter\def\csname N@hm-qwen-pool-shp-div-corr\endcsname{1.14}
\expandafter\def\csname N@hm-qwen-pool-shp-div-rel\endcsname{0.97}
\expandafter\def\csname N@hm-qwen-pool-shp-div-cross\endcsname{1.88}
\expandafter\def\csname N@hm-qwen-pool-shp-div-d\endcsname{$-$0.49}
\expandafter\def\csname N@hm-qwen-pool-shp-div-ci\endcsname{$[-0.68, -0.25]$}
\expandafter\def\csname N@hm-qwen-pool-shp-div-pwin\endcsname{100\%}
\expandafter\def\csname N@hm-qwen-pool-shp-div-edp\endcsname{0.183}
\expandafter\def\csname N@hm-qwen-pool-shp-div-edpci\endcsname{$[-0.195, -0.040]$}
\expandafter\def\csname N@hm-qwen-pool-shp-argyle-corr\endcsname{1.33}
\expandafter\def\csname N@hm-qwen-pool-shp-argyle-rel\endcsname{1.06}
\expandafter\def\csname N@hm-qwen-pool-shp-argyle-cross\endcsname{2.14}
\expandafter\def\csname N@hm-qwen-pool-shp-argyle-d\endcsname{$-$0.68}
\expandafter\def\csname N@hm-qwen-pool-shp-argyle-ci\endcsname{$[-0.88, -0.44]$}
\expandafter\def\csname N@hm-qwen-pool-shp-argyle-pwin\endcsname{100\%}
\expandafter\def\csname N@hm-qwen-pool-shp-argyle-edp\endcsname{0.323}
\expandafter\def\csname N@hm-qwen-pool-shp-argyle-edpci\endcsname{$[-0.358, -0.173]$}
\expandafter\def\csname N@hm-qwen-pool-shp-story-corr\endcsname{1.05}
\expandafter\def\csname N@hm-qwen-pool-shp-story-rel\endcsname{1.04}
\expandafter\def\csname N@hm-qwen-pool-shp-story-cross\endcsname{1.86}
\expandafter\def\csname N@hm-qwen-pool-shp-story-d\endcsname{$-$0.41}
\expandafter\def\csname N@hm-qwen-pool-shp-story-ci\endcsname{$[-0.59, -0.16]$}
\expandafter\def\csname N@hm-qwen-pool-shp-story-pwin\endcsname{100\%}
\expandafter\def\csname N@hm-qwen-pool-shp-story-edp\endcsname{0.129}
\expandafter\def\csname N@hm-qwen-pool-shp-story-edpci\endcsname{$[-0.144, 0.005]$}
\expandafter\def\csname N@hm-qwen-pool-shp-drange\endcsname{$-$0.68 to $-$0.41}
\expandafter\def\csname N@hm-qwen-pool-shp-nsig\endcsname{4}
\expandafter\def\csname N@hm-qwen-pool-shp-relx\endcsname{1.0--1.1}
\expandafter\def\csname N@hm-qwen-pool-shp-pwinmin\endcsname{100\%}
\expandafter\def\csname N@hm-qwen-pool-shp-pwinphrase\endcsname{all}
\expandafter\def\csname N@hm-qwen-pool-abl-k16-corr\endcsname{0.86}
\expandafter\def\csname N@hm-qwen-pool-abl-k16-cross\endcsname{0.28}
\expandafter\def\csname N@hm-qwen-pool-abl-k16-pwin\endcsname{99.5\%}
\expandafter\def\csname N@hm-qwen-pool-abl-k16-gap\endcsname{0.40}
\expandafter\def\csname N@hm-qwen-pool-abl-k16-dcross\endcsname{0.42}
\expandafter\def\csname N@hm-qwen-pool-abl-k16-dcross-ci\endcsname{$[0.20, 0.65]$}
\expandafter\def\csname N@hm-qwen-pool-abl-k16-dgap\endcsname{0.26}
\expandafter\def\csname N@hm-qwen-pool-abl-k16-dgap-ci\endcsname{$[0.08, 0.51]$}
\expandafter\def\csname N@hm-qwen-pool-abl-k16-dcorr\endcsname{$-$0.21}
\expandafter\def\csname N@hm-qwen-pool-abl-k16-dcorr-ci\endcsname{$[-0.38, -0.03]$}
\expandafter\def\csname N@hm-qwen-pool-abl-k16-rel\endcsname{0.91}
\expandafter\def\csname N@hm-qwen-pool-abl-nosharp-corr\endcsname{0.68}
\expandafter\def\csname N@hm-qwen-pool-abl-nosharp-cross\endcsname{0.55}
\expandafter\def\csname N@hm-qwen-pool-abl-nosharp-pwin\endcsname{66.5\%}
\expandafter\def\csname N@hm-qwen-pool-abl-nosharp-gap\endcsname{0.65}
\expandafter\def\csname N@hm-qwen-pool-abl-nosharp-dcross\endcsname{0.15}
\expandafter\def\csname N@hm-qwen-pool-abl-nosharp-dcross-ci\endcsname{$[-0.02, 0.34]$}
\expandafter\def\csname N@hm-qwen-pool-abl-nosharp-dgap\endcsname{0.01}
\expandafter\def\csname N@hm-qwen-pool-abl-nosharp-dgap-ci\endcsname{$[-0.12, 0.16]$}
\expandafter\def\csname N@hm-qwen-pool-abl-nosharp-dcorr\endcsname{$-$0.03}
\expandafter\def\csname N@hm-qwen-pool-abl-nosharp-dcorr-ci\endcsname{$[-0.17, 0.10]$}
\expandafter\def\csname N@hm-qwen-pool-abl-nosharp-rel\endcsname{0.69}
\expandafter\def\csname N@hm-qwen-pool-agg-corr\endcsname{1.11}
\expandafter\def\csname N@hm-qwen-pool-agg-cross\endcsname{0.16}
\expandafter\def\csname N@hm-qwen-pool-agg-pwin\endcsname{100.0\%}
\expandafter\def\csname N@hm-qwen-pool-agg-gap\endcsname{0.14}
\expandafter\def\csname N@hm-qwen-pool-agg-dcross\endcsname{0.55}
\expandafter\def\csname N@hm-qwen-pool-agg-dcross-ci\endcsname{$[0.35, 0.74]$}
\expandafter\def\csname N@hm-qwen-pool-agg-dgap\endcsname{0.53}
\expandafter\def\csname N@hm-qwen-pool-agg-dgap-ci\endcsname{$[0.33, 0.75]$}
\expandafter\def\csname N@hm-qwen-pool-agg-dcorr\endcsname{$-$0.46}
\expandafter\def\csname N@hm-qwen-pool-agg-dcorr-ci\endcsname{$[-0.64, -0.29]$}
\expandafter\def\csname N@hm-qwen-pool-agg-rel\endcsname{0.00}
\expandafter\def\csname N@hm-qwen-pool-abl-seed2-corr\endcsname{0.64}
\expandafter\def\csname N@hm-qwen-pool-abl-seed2-cross\endcsname{0.76}
\expandafter\def\csname N@hm-qwen-pool-abl-seed2-pwin\endcsname{45.0\%}
\expandafter\def\csname N@hm-qwen-pool-abl-seed2-gap\endcsname{0.68}
\expandafter\def\csname N@hm-qwen-pool-abl-seed2-dcross\endcsname{$-$0.06}
\expandafter\def\csname N@hm-qwen-pool-abl-seed2-dcross-ci\endcsname{$[-0.32, 0.16]$}
\expandafter\def\csname N@hm-qwen-pool-abl-seed2-dgap\endcsname{$-$0.02}
\expandafter\def\csname N@hm-qwen-pool-abl-seed2-dgap-ci\endcsname{$[-0.24, 0.18]$}
\expandafter\def\csname N@hm-qwen-pool-abl-seed2-dcorr\endcsname{0.01}
\expandafter\def\csname N@hm-qwen-pool-abl-seed2-dcorr-ci\endcsname{$[-0.15, 0.16]$}
\expandafter\def\csname N@hm-qwen-pool-abl-seed2-rel\endcsname{1.01}
\expandafter\def\csname N@hm-qwen-pool-abl-demsharp-corr\endcsname{1.30}
\expandafter\def\csname N@hm-qwen-pool-abl-demsharp-cross\endcsname{2.09}
\expandafter\def\csname N@hm-qwen-pool-abl-demsharp-pwin\endcsname{100.0\%}
\expandafter\def\csname N@hm-qwen-pool-abl-demsharp-gap\endcsname{1.81}
\expandafter\def\csname N@hm-qwen-pool-abl-demsharp-dcross\endcsname{$-$1.38}
\expandafter\def\csname N@hm-qwen-pool-abl-demsharp-dcross-ci\endcsname{$[-1.67, -1.13]$}
\expandafter\def\csname N@hm-qwen-pool-abl-demsharp-dgap\endcsname{$-$1.15}
\expandafter\def\csname N@hm-qwen-pool-abl-demsharp-dgap-ci\endcsname{$[-1.37, -0.94]$}
\expandafter\def\csname N@hm-qwen-pool-abl-demsharp-dcorr\endcsname{$-$0.65}
\expandafter\def\csname N@hm-qwen-pool-abl-demsharp-dcorr-ci\endcsname{$[-0.85, -0.43]$}
\expandafter\def\csname N@hm-qwen-pool-abl-demsharp-rel\endcsname{1.00}
\expandafter\def\csname N@hm-qwen-pool-demo-corr\endcsname{0.77}
\expandafter\def\csname N@hm-qwen-pool-demo-cross\endcsname{1.41}
\expandafter\def\csname N@hm-qwen-pool-demo-pwin\endcsname{85.0\%}
\expandafter\def\csname N@hm-qwen-pool-demo-gap\endcsname{1.52}
\expandafter\def\csname N@hm-qwen-pool-demo-dcross\endcsname{$-$0.70}
\expandafter\def\csname N@hm-qwen-pool-demo-dcross-ci\endcsname{$[-0.99, -0.42]$}
\expandafter\def\csname N@hm-qwen-pool-demo-dgap\endcsname{$-$0.85}
\expandafter\def\csname N@hm-qwen-pool-demo-dgap-ci\endcsname{$[-1.09, -0.64]$}
\expandafter\def\csname N@hm-qwen-pool-demo-dcorr\endcsname{$-$0.12}
\expandafter\def\csname N@hm-qwen-pool-demo-dcorr-ci\endcsname{$[-0.32, 0.10]$}
\expandafter\def\csname N@hm-qwen-pool-demo-rel\endcsname{0.67}
\expandafter\def\csname N@hm-gemma-otp-floor-corr\endcsname{0.26}
\expandafter\def\csname N@hm-gemma-otp-floor-cross\endcsname{0.98}
\expandafter\def\csname N@hm-gemma-otp-lpp-corr\endcsname{0.54}
\expandafter\def\csname N@hm-gemma-otp-lpp-rel\endcsname{1.16}
\expandafter\def\csname N@hm-gemma-otp-lpp-cross\endcsname{0.97}
\expandafter\def\csname N@hm-gemma-otp-lpp-gap\endcsname{1.09}
\expandafter\def\csname N@hm-gemma-otp-lpp-wpsd\endcsname{1.00}
\expandafter\def\csname N@hm-gemma-otp-lpp-edp\endcsname{0.010}
\expandafter\def\csname N@hm-gemma-otp-sel-corr\endcsname{0.62--0.92}
\expandafter\def\csname N@hm-gemma-otp-sel-rel\endcsname{0.22--1.02}
\expandafter\def\csname N@hm-gemma-otp-sel-n\endcsname{7}
\expandafter\def\csname N@hm-gemma-otp-sel-nworse\endcsname{7}
\expandafter\def\csname N@hm-gemma-otp-sel-divbest-corr\endcsname{0.92}
\expandafter\def\csname N@hm-gemma-otp-sel-divbest-rel\endcsname{0.97}
\expandafter\def\csname N@hm-gemma-otp-sel-divbest-cross\endcsname{1.62}
\expandafter\def\csname N@hm-gemma-otp-sel-divbest-d\endcsname{$-$0.38}
\expandafter\def\csname N@hm-gemma-otp-sel-divbest-ci\endcsname{$[-0.54, -0.22]$}
\expandafter\def\csname N@hm-gemma-otp-sel-divbest-pwin\endcsname{100\%}
\expandafter\def\csname N@hm-gemma-otp-sel-divbest-edp\endcsname{0.031}
\expandafter\def\csname N@hm-gemma-otp-sel-divbest-edpci\endcsname{$[-0.044, -0.001]$}
\expandafter\def\csname N@hm-gemma-otp-sel-rw-corr\endcsname{0.92}
\expandafter\def\csname N@hm-gemma-otp-sel-rw-rel\endcsname{1.02}
\expandafter\def\csname N@hm-gemma-otp-sel-rw-cross\endcsname{1.46}
\expandafter\def\csname N@hm-gemma-otp-sel-rw-d\endcsname{$-$0.38}
\expandafter\def\csname N@hm-gemma-otp-sel-rw-ci\endcsname{$[-0.56, -0.22]$}
\expandafter\def\csname N@hm-gemma-otp-sel-rw-pwin\endcsname{100\%}
\expandafter\def\csname N@hm-gemma-otp-sel-rw-edp\endcsname{0.047}
\expandafter\def\csname N@hm-gemma-otp-sel-rw-edpci\endcsname{$[-0.067, -0.007]$}
\expandafter\def\csname N@hm-gemma-otp-sel-base-corr\endcsname{0.91}
\expandafter\def\csname N@hm-gemma-otp-sel-base-rel\endcsname{0.96}
\expandafter\def\csname N@hm-gemma-otp-sel-base-cross\endcsname{1.44}
\expandafter\def\csname N@hm-gemma-otp-sel-base-d\endcsname{$-$0.37}
\expandafter\def\csname N@hm-gemma-otp-sel-base-ci\endcsname{$[-0.54, -0.22]$}
\expandafter\def\csname N@hm-gemma-otp-sel-base-pwin\endcsname{100\%}
\expandafter\def\csname N@hm-gemma-otp-sel-base-edp\endcsname{0.040}
\expandafter\def\csname N@hm-gemma-otp-sel-base-edpci\endcsname{$[-0.051, -0.009]$}
\expandafter\def\csname N@hm-gemma-otp-sel-div-corr\endcsname{0.82}
\expandafter\def\csname N@hm-gemma-otp-sel-div-rel\endcsname{1.00}
\expandafter\def\csname N@hm-gemma-otp-sel-div-cross\endcsname{1.38}
\expandafter\def\csname N@hm-gemma-otp-sel-div-d\endcsname{$-$0.28}
\expandafter\def\csname N@hm-gemma-otp-sel-div-ci\endcsname{$[-0.46, -0.11]$}
\expandafter\def\csname N@hm-gemma-otp-sel-div-pwin\endcsname{100\%}
\expandafter\def\csname N@hm-gemma-otp-sel-div-edp\endcsname{0.033}
\expandafter\def\csname N@hm-gemma-otp-sel-div-edpci\endcsname{$[-0.045, -0.001]$}
\expandafter\def\csname N@hm-gemma-otp-sel-argyle-corr\endcsname{0.62}
\expandafter\def\csname N@hm-gemma-otp-sel-argyle-rel\endcsname{0.84}
\expandafter\def\csname N@hm-gemma-otp-sel-argyle-cross\endcsname{1.20}
\expandafter\def\csname N@hm-gemma-otp-sel-argyle-d\endcsname{$-$0.09}
\expandafter\def\csname N@hm-gemma-otp-sel-argyle-ci\endcsname{$[-0.29, 0.11]$}
\expandafter\def\csname N@hm-gemma-otp-sel-argyle-pwin\endcsname{80\%}
\expandafter\def\csname N@hm-gemma-otp-sel-argyle-edp\endcsname{0.025}
\expandafter\def\csname N@hm-gemma-otp-sel-argyle-edpci\endcsname{$[-0.035, 0.005]$}
\expandafter\def\csname N@hm-gemma-otp-sel-story-corr\endcsname{0.77}
\expandafter\def\csname N@hm-gemma-otp-sel-story-rel\endcsname{1.00}
\expandafter\def\csname N@hm-gemma-otp-sel-story-cross\endcsname{1.37}
\expandafter\def\csname N@hm-gemma-otp-sel-story-d\endcsname{$-$0.24}
\expandafter\def\csname N@hm-gemma-otp-sel-story-ci\endcsname{$[-0.40, -0.08]$}
\expandafter\def\csname N@hm-gemma-otp-sel-story-pwin\endcsname{100\%}
\expandafter\def\csname N@hm-gemma-otp-sel-story-edp\endcsname{0.030}
\expandafter\def\csname N@hm-gemma-otp-sel-story-edpci\endcsname{$[-0.043, 0.007]$}
\expandafter\def\csname N@hm-gemma-otp-sel-anth-corr\endcsname{0.78}
\expandafter\def\csname N@hm-gemma-otp-sel-anth-rel\endcsname{0.22}
\expandafter\def\csname N@hm-gemma-otp-sel-anth-cross\endcsname{0.46}
\expandafter\def\csname N@hm-gemma-otp-sel-anth-d\endcsname{$-$0.24}
\expandafter\def\csname N@hm-gemma-otp-sel-anth-ci\endcsname{$[-0.41, -0.08]$}
\expandafter\def\csname N@hm-gemma-otp-sel-anth-pwin\endcsname{100\%}
\expandafter\def\csname N@hm-gemma-otp-sel-anth-edp\endcsname{0.028}
\expandafter\def\csname N@hm-gemma-otp-sel-anth-edpci\endcsname{$[-0.039, 0.006]$}
\expandafter\def\csname N@hm-gemma-otp-sel-drange\endcsname{$-$0.38 to $-$0.09}
\expandafter\def\csname N@hm-gemma-otp-sel-nsig\endcsname{6}
\expandafter\def\csname N@hm-gemma-otp-sel-relx\endcsname{1.1--5.2}
\expandafter\def\csname N@hm-gemma-otp-sel-pwinmin\endcsname{80\%}
\expandafter\def\csname N@hm-gemma-otp-sel-pwinphrase\endcsname{at least 80\% of}
\expandafter\def\csname N@hm-gemma-otp-shp-corr\endcsname{0.60--0.90}
\expandafter\def\csname N@hm-gemma-otp-shp-rel\endcsname{0.92--1.10}
\expandafter\def\csname N@hm-gemma-otp-shp-n\endcsname{4}
\expandafter\def\csname N@hm-gemma-otp-shp-nworse\endcsname{4}
\expandafter\def\csname N@hm-gemma-otp-shp-base-corr\endcsname{0.90}
\expandafter\def\csname N@hm-gemma-otp-shp-base-rel\endcsname{1.08}
\expandafter\def\csname N@hm-gemma-otp-shp-base-cross\endcsname{1.44}
\expandafter\def\csname N@hm-gemma-otp-shp-base-d\endcsname{$-$0.37}
\expandafter\def\csname N@hm-gemma-otp-shp-base-ci\endcsname{$[-0.53, -0.20]$}
\expandafter\def\csname N@hm-gemma-otp-shp-base-pwin\endcsname{100\%}
\expandafter\def\csname N@hm-gemma-otp-shp-base-edp\endcsname{0.039}
\expandafter\def\csname N@hm-gemma-otp-shp-base-edpci\endcsname{$[-0.048, -0.007]$}
\expandafter\def\csname N@hm-gemma-otp-shp-div-corr\endcsname{0.84}
\expandafter\def\csname N@hm-gemma-otp-shp-div-rel\endcsname{1.08}
\expandafter\def\csname N@hm-gemma-otp-shp-div-cross\endcsname{1.40}
\expandafter\def\csname N@hm-gemma-otp-shp-div-d\endcsname{$-$0.31}
\expandafter\def\csname N@hm-gemma-otp-shp-div-ci\endcsname{$[-0.49, -0.13]$}
\expandafter\def\csname N@hm-gemma-otp-shp-div-pwin\endcsname{100\%}
\expandafter\def\csname N@hm-gemma-otp-shp-div-edp\endcsname{0.035}
\expandafter\def\csname N@hm-gemma-otp-shp-div-edpci\endcsname{$[-0.046, -0.003]$}
\expandafter\def\csname N@hm-gemma-otp-shp-argyle-corr\endcsname{0.60}
\expandafter\def\csname N@hm-gemma-otp-shp-argyle-rel\endcsname{0.92}
\expandafter\def\csname N@hm-gemma-otp-shp-argyle-cross\endcsname{1.18}
\expandafter\def\csname N@hm-gemma-otp-shp-argyle-d\endcsname{$-$0.07}
\expandafter\def\csname N@hm-gemma-otp-shp-argyle-ci\endcsname{$[-0.30, 0.13]$}
\expandafter\def\csname N@hm-gemma-otp-shp-argyle-pwin\endcsname{75\%}
\expandafter\def\csname N@hm-gemma-otp-shp-argyle-edp\endcsname{0.025}
\expandafter\def\csname N@hm-gemma-otp-shp-argyle-edpci\endcsname{$[-0.039, 0.009]$}
\expandafter\def\csname N@hm-gemma-otp-shp-story-corr\endcsname{0.78}
\expandafter\def\csname N@hm-gemma-otp-shp-story-rel\endcsname{1.10}
\expandafter\def\csname N@hm-gemma-otp-shp-story-cross\endcsname{1.39}
\expandafter\def\csname N@hm-gemma-otp-shp-story-d\endcsname{$-$0.24}
\expandafter\def\csname N@hm-gemma-otp-shp-story-ci\endcsname{$[-0.41, -0.05]$}
\expandafter\def\csname N@hm-gemma-otp-shp-story-pwin\endcsname{98\%}
\expandafter\def\csname N@hm-gemma-otp-shp-story-edp\endcsname{0.030}
\expandafter\def\csname N@hm-gemma-otp-shp-story-edpci\endcsname{$[-0.039, 0.004]$}
\expandafter\def\csname N@hm-gemma-otp-shp-drange\endcsname{$-$0.37 to $-$0.07}
\expandafter\def\csname N@hm-gemma-otp-shp-nsig\endcsname{3}
\expandafter\def\csname N@hm-gemma-otp-shp-relx\endcsname{1.1--1.3}
\expandafter\def\csname N@hm-gemma-otp-shp-pwinmin\endcsname{75\%}
\expandafter\def\csname N@hm-gemma-otp-shp-pwinphrase\endcsname{at least 75\% of}
\expandafter\def\csname N@hm-gemma-otp-abl-k16-corr\endcsname{0.52}
\expandafter\def\csname N@hm-gemma-otp-abl-k16-cross\endcsname{0.74}
\expandafter\def\csname N@hm-gemma-otp-abl-k16-pwin\endcsname{46.0\%}
\expandafter\def\csname N@hm-gemma-otp-abl-k16-gap\endcsname{1.10}
\expandafter\def\csname N@hm-gemma-otp-abl-k16-dcross\endcsname{0.23}
\expandafter\def\csname N@hm-gemma-otp-abl-k16-dcross-ci\endcsname{$[-0.01, 0.47]$}
\expandafter\def\csname N@hm-gemma-otp-abl-k16-dgap\endcsname{$-$0.00}
\expandafter\def\csname N@hm-gemma-otp-abl-k16-dgap-ci\endcsname{$[-0.19, 0.16]$}
\expandafter\def\csname N@hm-gemma-otp-abl-k16-dcorr\endcsname{0.01}
\expandafter\def\csname N@hm-gemma-otp-abl-k16-dcorr-ci\endcsname{$[-0.13, 0.15]$}
\expandafter\def\csname N@hm-gemma-otp-abl-k16-rel\endcsname{1.16}
\expandafter\def\csname N@hm-gemma-otp-abl-nosharp-corr\endcsname{0.54}
\expandafter\def\csname N@hm-gemma-otp-abl-nosharp-cross\endcsname{0.97}
\expandafter\def\csname N@hm-gemma-otp-abl-nosharp-pwin\endcsname{53.5\%}
\expandafter\def\csname N@hm-gemma-otp-abl-nosharp-gap\endcsname{1.09}
\expandafter\def\csname N@hm-gemma-otp-abl-nosharp-dcross\endcsname{$-$0.00}
\expandafter\def\csname N@hm-gemma-otp-abl-nosharp-dcross-ci\endcsname{$[-0.13, 0.12]$}
\expandafter\def\csname N@hm-gemma-otp-abl-nosharp-dgap\endcsname{$-$0.00}
\expandafter\def\csname N@hm-gemma-otp-abl-nosharp-dgap-ci\endcsname{$[-0.08, 0.09]$}
\expandafter\def\csname N@hm-gemma-otp-abl-nosharp-dcorr\endcsname{$-$0.01}
\expandafter\def\csname N@hm-gemma-otp-abl-nosharp-dcorr-ci\endcsname{$[-0.11, 0.09]$}
\expandafter\def\csname N@hm-gemma-otp-abl-nosharp-rel\endcsname{1.13}
\expandafter\def\csname N@hm-gemma-otp-agg-corr\endcsname{0.78}
\expandafter\def\csname N@hm-gemma-otp-agg-cross\endcsname{0.45}
\expandafter\def\csname N@hm-gemma-otp-agg-pwin\endcsname{99.5\%}
\expandafter\def\csname N@hm-gemma-otp-agg-gap\endcsname{1.10}
\expandafter\def\csname N@hm-gemma-otp-agg-dcross\endcsname{0.52}
\expandafter\def\csname N@hm-gemma-otp-agg-dcross-ci\endcsname{$[0.31, 0.76]$}
\expandafter\def\csname N@hm-gemma-otp-agg-dgap\endcsname{$-$0.01}
\expandafter\def\csname N@hm-gemma-otp-agg-dgap-ci\endcsname{$[-0.19, 0.18]$}
\expandafter\def\csname N@hm-gemma-otp-agg-dcorr\endcsname{$-$0.25}
\expandafter\def\csname N@hm-gemma-otp-agg-dcorr-ci\endcsname{$[-0.46, -0.05]$}
\expandafter\def\csname N@hm-gemma-otp-agg-rel\endcsname{0.21}
\expandafter\def\csname N@hm-gemma-otp-abl-seed2-corr\endcsname{0.59}
\expandafter\def\csname N@hm-gemma-otp-abl-seed2-cross\endcsname{1.04}
\expandafter\def\csname N@hm-gemma-otp-abl-seed2-pwin\endcsname{75.5\%}
\expandafter\def\csname N@hm-gemma-otp-abl-seed2-gap\endcsname{1.10}
\expandafter\def\csname N@hm-gemma-otp-abl-seed2-dcross\endcsname{$-$0.07}
\expandafter\def\csname N@hm-gemma-otp-abl-seed2-dcross-ci\endcsname{$[-0.28, 0.15]$}
\expandafter\def\csname N@hm-gemma-otp-abl-seed2-dgap\endcsname{$-$0.00}
\expandafter\def\csname N@hm-gemma-otp-abl-seed2-dgap-ci\endcsname{$[-0.19, 0.17]$}
\expandafter\def\csname N@hm-gemma-otp-abl-seed2-dcorr\endcsname{$-$0.06}
\expandafter\def\csname N@hm-gemma-otp-abl-seed2-dcorr-ci\endcsname{$[-0.21, 0.08]$}
\expandafter\def\csname N@hm-gemma-otp-abl-seed2-rel\endcsname{1.16}
\expandafter\def\csname N@hm-gemma-otp-abl-demsharp-corr\endcsname{0.90}
\expandafter\def\csname N@hm-gemma-otp-abl-demsharp-cross\endcsname{1.45}
\expandafter\def\csname N@hm-gemma-otp-abl-demsharp-pwin\endcsname{100.0\%}
\expandafter\def\csname N@hm-gemma-otp-abl-demsharp-gap\endcsname{1.09}
\expandafter\def\csname N@hm-gemma-otp-abl-demsharp-dcross\endcsname{$-$0.48}
\expandafter\def\csname N@hm-gemma-otp-abl-demsharp-dcross-ci\endcsname{$[-0.66, -0.27]$}
\expandafter\def\csname N@hm-gemma-otp-abl-demsharp-dgap\endcsname{0.00}
\expandafter\def\csname N@hm-gemma-otp-abl-demsharp-dgap-ci\endcsname{$[-0.17, 0.18]$}
\expandafter\def\csname N@hm-gemma-otp-abl-demsharp-dcorr\endcsname{$-$0.37}
\expandafter\def\csname N@hm-gemma-otp-abl-demsharp-dcorr-ci\endcsname{$[-0.52, -0.20]$}
\expandafter\def\csname N@hm-gemma-otp-abl-demsharp-rel\endcsname{1.07}
\expandafter\def\csname N@hm-gemma-otp-demo-corr\endcsname{0.91}
\expandafter\def\csname N@hm-gemma-otp-demo-cross\endcsname{1.45}
\expandafter\def\csname N@hm-gemma-otp-demo-pwin\endcsname{100.0\%}
\expandafter\def\csname N@hm-gemma-otp-demo-gap\endcsname{1.09}
\expandafter\def\csname N@hm-gemma-otp-demo-dcross\endcsname{$-$0.48}
\expandafter\def\csname N@hm-gemma-otp-demo-dcross-ci\endcsname{$[-0.70, -0.27]$}
\expandafter\def\csname N@hm-gemma-otp-demo-dgap\endcsname{$-$0.00}
\expandafter\def\csname N@hm-gemma-otp-demo-dgap-ci\endcsname{$[-0.17, 0.18]$}
\expandafter\def\csname N@hm-gemma-otp-demo-dcorr\endcsname{$-$0.38}
\expandafter\def\csname N@hm-gemma-otp-demo-dcorr-ci\endcsname{$[-0.55, -0.22]$}
\expandafter\def\csname N@hm-gemma-otp-demo-rel\endcsname{1.02}
\expandafter\def\csname N@hm-gemma-pool-floor-corr\endcsname{0.26}
\expandafter\def\csname N@hm-gemma-pool-floor-cross\endcsname{0.98}
\expandafter\def\csname N@hm-gemma-pool-lpp-corr\endcsname{0.52}
\expandafter\def\csname N@hm-gemma-pool-lpp-rel\endcsname{1.16}
\expandafter\def\csname N@hm-gemma-pool-lpp-cross\endcsname{0.84}
\expandafter\def\csname N@hm-gemma-pool-lpp-gap\endcsname{0.83}
\expandafter\def\csname N@hm-gemma-pool-lpp-wpsd\endcsname{1.05}
\expandafter\def\csname N@hm-gemma-pool-lpp-edp\endcsname{0.027}
\expandafter\def\csname N@hm-gemma-pool-sel-corr\endcsname{1.07--1.37}
\expandafter\def\csname N@hm-gemma-pool-sel-rel\endcsname{0.02--1.11}
\expandafter\def\csname N@hm-gemma-pool-sel-n\endcsname{7}
\expandafter\def\csname N@hm-gemma-pool-sel-nworse\endcsname{7}
\expandafter\def\csname N@hm-gemma-pool-sel-divbest-corr\endcsname{1.37}
\expandafter\def\csname N@hm-gemma-pool-sel-divbest-rel\endcsname{1.10}
\expandafter\def\csname N@hm-gemma-pool-sel-divbest-cross\endcsname{2.19}
\expandafter\def\csname N@hm-gemma-pool-sel-divbest-d\endcsname{$-$0.84}
\expandafter\def\csname N@hm-gemma-pool-sel-divbest-ci\endcsname{$[-1.00, -0.63]$}
\expandafter\def\csname N@hm-gemma-pool-sel-divbest-pwin\endcsname{100\%}
\expandafter\def\csname N@hm-gemma-pool-sel-divbest-edp\endcsname{0.252}
\expandafter\def\csname N@hm-gemma-pool-sel-divbest-edpci\endcsname{$[-0.299, -0.143]$}
\expandafter\def\csname N@hm-gemma-pool-sel-rw-corr\endcsname{1.19}
\expandafter\def\csname N@hm-gemma-pool-sel-rw-rel\endcsname{1.10}
\expandafter\def\csname N@hm-gemma-pool-sel-rw-cross\endcsname{1.74}
\expandafter\def\csname N@hm-gemma-pool-sel-rw-d\endcsname{$-$0.67}
\expandafter\def\csname N@hm-gemma-pool-sel-rw-ci\endcsname{$[-0.83, -0.48]$}
\expandafter\def\csname N@hm-gemma-pool-sel-rw-pwin\endcsname{100\%}
\expandafter\def\csname N@hm-gemma-pool-sel-rw-edp\endcsname{0.208}
\expandafter\def\csname N@hm-gemma-pool-sel-rw-edpci\endcsname{$[-0.267, -0.111]$}
\expandafter\def\csname N@hm-gemma-pool-sel-base-corr\endcsname{1.18}
\expandafter\def\csname N@hm-gemma-pool-sel-base-rel\endcsname{1.05}
\expandafter\def\csname N@hm-gemma-pool-sel-base-cross\endcsname{1.73}
\expandafter\def\csname N@hm-gemma-pool-sel-base-d\endcsname{$-$0.66}
\expandafter\def\csname N@hm-gemma-pool-sel-base-ci\endcsname{$[-0.82, -0.49]$}
\expandafter\def\csname N@hm-gemma-pool-sel-base-pwin\endcsname{100\%}
\expandafter\def\csname N@hm-gemma-pool-sel-base-edp\endcsname{0.194}
\expandafter\def\csname N@hm-gemma-pool-sel-base-edpci\endcsname{$[-0.227, -0.103]$}
\expandafter\def\csname N@hm-gemma-pool-sel-div-corr\endcsname{1.13}
\expandafter\def\csname N@hm-gemma-pool-sel-div-rel\endcsname{1.11}
\expandafter\def\csname N@hm-gemma-pool-sel-div-cross\endcsname{1.71}
\expandafter\def\csname N@hm-gemma-pool-sel-div-d\endcsname{$-$0.61}
\expandafter\def\csname N@hm-gemma-pool-sel-div-ci\endcsname{$[-0.76, -0.43]$}
\expandafter\def\csname N@hm-gemma-pool-sel-div-pwin\endcsname{100\%}
\expandafter\def\csname N@hm-gemma-pool-sel-div-edp\endcsname{0.157}
\expandafter\def\csname N@hm-gemma-pool-sel-div-edpci\endcsname{$[-0.189, -0.071]$}
\expandafter\def\csname N@hm-gemma-pool-sel-argyle-corr\endcsname{1.34}
\expandafter\def\csname N@hm-gemma-pool-sel-argyle-rel\endcsname{1.06}
\expandafter\def\csname N@hm-gemma-pool-sel-argyle-cross\endcsname{2.14}
\expandafter\def\csname N@hm-gemma-pool-sel-argyle-d\endcsname{$-$0.82}
\expandafter\def\csname N@hm-gemma-pool-sel-argyle-ci\endcsname{$[-1.06, -0.59]$}
\expandafter\def\csname N@hm-gemma-pool-sel-argyle-pwin\endcsname{100\%}
\expandafter\def\csname N@hm-gemma-pool-sel-argyle-edp\endcsname{0.336}
\expandafter\def\csname N@hm-gemma-pool-sel-argyle-edpci\endcsname{$[-0.395, -0.231]$}
\expandafter\def\csname N@hm-gemma-pool-sel-story-corr\endcsname{1.07}
\expandafter\def\csname N@hm-gemma-pool-sel-story-rel\endcsname{1.10}
\expandafter\def\csname N@hm-gemma-pool-sel-story-cross\endcsname{1.65}
\expandafter\def\csname N@hm-gemma-pool-sel-story-d\endcsname{$-$0.55}
\expandafter\def\csname N@hm-gemma-pool-sel-story-ci\endcsname{$[-0.73, -0.39]$}
\expandafter\def\csname N@hm-gemma-pool-sel-story-pwin\endcsname{100\%}
\expandafter\def\csname N@hm-gemma-pool-sel-story-edp\endcsname{0.171}
\expandafter\def\csname N@hm-gemma-pool-sel-story-edpci\endcsname{$[-0.217, -0.087]$}
\expandafter\def\csname N@hm-gemma-pool-sel-anth-corr\endcsname{1.10}
\expandafter\def\csname N@hm-gemma-pool-sel-anth-rel\endcsname{0.02}
\expandafter\def\csname N@hm-gemma-pool-sel-anth-cross\endcsname{0.16}
\expandafter\def\csname N@hm-gemma-pool-sel-anth-d\endcsname{$-$0.58}
\expandafter\def\csname N@hm-gemma-pool-sel-anth-ci\endcsname{$[-0.73, -0.42]$}
\expandafter\def\csname N@hm-gemma-pool-sel-anth-pwin\endcsname{100\%}
\expandafter\def\csname N@hm-gemma-pool-sel-anth-edp\endcsname{0.272}
\expandafter\def\csname N@hm-gemma-pool-sel-anth-edpci\endcsname{$[-0.334, -0.163]$}
\expandafter\def\csname N@hm-gemma-pool-sel-drange\endcsname{$-$0.84 to $-$0.55}
\expandafter\def\csname N@hm-gemma-pool-sel-nsig\endcsname{7}
\expandafter\def\csname N@hm-gemma-pool-sel-relx\endcsname{1.0--61.8}
\expandafter\def\csname N@hm-gemma-pool-sel-pwinmin\endcsname{100\%}
\expandafter\def\csname N@hm-gemma-pool-sel-pwinphrase\endcsname{all}
\expandafter\def\csname N@hm-gemma-pool-shp-corr\endcsname{1.09--1.30}
\expandafter\def\csname N@hm-gemma-pool-shp-rel\endcsname{1.11--1.19}
\expandafter\def\csname N@hm-gemma-pool-shp-n\endcsname{4}
\expandafter\def\csname N@hm-gemma-pool-shp-nworse\endcsname{4}
\expandafter\def\csname N@hm-gemma-pool-shp-base-corr\endcsname{1.18}
\expandafter\def\csname N@hm-gemma-pool-shp-base-rel\endcsname{1.14}
\expandafter\def\csname N@hm-gemma-pool-shp-base-cross\endcsname{1.73}
\expandafter\def\csname N@hm-gemma-pool-shp-base-d\endcsname{$-$0.66}
\expandafter\def\csname N@hm-gemma-pool-shp-base-ci\endcsname{$[-0.80, -0.49]$}
\expandafter\def\csname N@hm-gemma-pool-shp-base-pwin\endcsname{100\%}
\expandafter\def\csname N@hm-gemma-pool-shp-base-edp\endcsname{0.200}
\expandafter\def\csname N@hm-gemma-pool-shp-base-edpci\endcsname{$[-0.238, -0.109]$}
\expandafter\def\csname N@hm-gemma-pool-shp-div-corr\endcsname{1.15}
\expandafter\def\csname N@hm-gemma-pool-shp-div-rel\endcsname{1.19}
\expandafter\def\csname N@hm-gemma-pool-shp-div-cross\endcsname{1.72}
\expandafter\def\csname N@hm-gemma-pool-shp-div-d\endcsname{$-$0.62}
\expandafter\def\csname N@hm-gemma-pool-shp-div-ci\endcsname{$[-0.78, -0.44]$}
\expandafter\def\csname N@hm-gemma-pool-shp-div-pwin\endcsname{100\%}
\expandafter\def\csname N@hm-gemma-pool-shp-div-edp\endcsname{0.157}
\expandafter\def\csname N@hm-gemma-pool-shp-div-edpci\endcsname{$[-0.194, -0.063]$}
\expandafter\def\csname N@hm-gemma-pool-shp-argyle-corr\endcsname{1.30}
\expandafter\def\csname N@hm-gemma-pool-shp-argyle-rel\endcsname{1.11}
\expandafter\def\csname N@hm-gemma-pool-shp-argyle-cross\endcsname{2.09}
\expandafter\def\csname N@hm-gemma-pool-shp-argyle-d\endcsname{$-$0.78}
\expandafter\def\csname N@hm-gemma-pool-shp-argyle-ci\endcsname{$[-1.00, -0.56]$}
\expandafter\def\csname N@hm-gemma-pool-shp-argyle-pwin\endcsname{100\%}
\expandafter\def\csname N@hm-gemma-pool-shp-argyle-edp\endcsname{0.326}
\expandafter\def\csname N@hm-gemma-pool-shp-argyle-edpci\endcsname{$[-0.380, -0.220]$}
\expandafter\def\csname N@hm-gemma-pool-shp-story-corr\endcsname{1.09}
\expandafter\def\csname N@hm-gemma-pool-shp-story-rel\endcsname{1.19}
\expandafter\def\csname N@hm-gemma-pool-shp-story-cross\endcsname{1.68}
\expandafter\def\csname N@hm-gemma-pool-shp-story-d\endcsname{$-$0.56}
\expandafter\def\csname N@hm-gemma-pool-shp-story-ci\endcsname{$[-0.73, -0.38]$}
\expandafter\def\csname N@hm-gemma-pool-shp-story-pwin\endcsname{100\%}
\expandafter\def\csname N@hm-gemma-pool-shp-story-edp\endcsname{0.177}
\expandafter\def\csname N@hm-gemma-pool-shp-story-edpci\endcsname{$[-0.213, -0.093]$}
\expandafter\def\csname N@hm-gemma-pool-shp-drange\endcsname{$-$0.78 to $-$0.56}
\expandafter\def\csname N@hm-gemma-pool-shp-nsig\endcsname{4}
\expandafter\def\csname N@hm-gemma-pool-shp-relx\endcsname{1.0}
\expandafter\def\csname N@hm-gemma-pool-shp-pwinmin\endcsname{100\%}
\expandafter\def\csname N@hm-gemma-pool-shp-pwinphrase\endcsname{all}
\expandafter\def\csname N@hm-gemma-pool-abl-k16-corr\endcsname{0.61}
\expandafter\def\csname N@hm-gemma-pool-abl-k16-cross\endcsname{0.53}
\expandafter\def\csname N@hm-gemma-pool-abl-k16-pwin\endcsname{85.0\%}
\expandafter\def\csname N@hm-gemma-pool-abl-k16-gap\endcsname{0.65}
\expandafter\def\csname N@hm-gemma-pool-abl-k16-dcross\endcsname{0.31}
\expandafter\def\csname N@hm-gemma-pool-abl-k16-dcross-ci\endcsname{$[0.08, 0.53]$}
\expandafter\def\csname N@hm-gemma-pool-abl-k16-dgap\endcsname{0.18}
\expandafter\def\csname N@hm-gemma-pool-abl-k16-dgap-ci\endcsname{$[-0.03, 0.41]$}
\expandafter\def\csname N@hm-gemma-pool-abl-k16-dcorr\endcsname{$-$0.09}
\expandafter\def\csname N@hm-gemma-pool-abl-k16-dcorr-ci\endcsname{$[-0.24, 0.07]$}
\expandafter\def\csname N@hm-gemma-pool-abl-k16-rel\endcsname{1.17}
\expandafter\def\csname N@hm-gemma-pool-abl-nosharp-corr\endcsname{0.53}
\expandafter\def\csname N@hm-gemma-pool-abl-nosharp-cross\endcsname{0.84}
\expandafter\def\csname N@hm-gemma-pool-abl-nosharp-pwin\endcsname{54.0\%}
\expandafter\def\csname N@hm-gemma-pool-abl-nosharp-gap\endcsname{0.83}
\expandafter\def\csname N@hm-gemma-pool-abl-nosharp-dcross\endcsname{$-$0.00}
\expandafter\def\csname N@hm-gemma-pool-abl-nosharp-dcross-ci\endcsname{$[-0.13, 0.12]$}
\expandafter\def\csname N@hm-gemma-pool-abl-nosharp-dgap\endcsname{0.00}
\expandafter\def\csname N@hm-gemma-pool-abl-nosharp-dgap-ci\endcsname{$[-0.08, 0.09]$}
\expandafter\def\csname N@hm-gemma-pool-abl-nosharp-dcorr\endcsname{$-$0.01}
\expandafter\def\csname N@hm-gemma-pool-abl-nosharp-dcorr-ci\endcsname{$[-0.10, 0.09]$}
\expandafter\def\csname N@hm-gemma-pool-abl-nosharp-rel\endcsname{1.13}
\expandafter\def\csname N@hm-gemma-pool-agg-corr\endcsname{1.11}
\expandafter\def\csname N@hm-gemma-pool-agg-cross\endcsname{0.16}
\expandafter\def\csname N@hm-gemma-pool-agg-pwin\endcsname{100.0\%}
\expandafter\def\csname N@hm-gemma-pool-agg-gap\endcsname{0.14}
\expandafter\def\csname N@hm-gemma-pool-agg-dcross\endcsname{0.68}
\expandafter\def\csname N@hm-gemma-pool-agg-dcross-ci\endcsname{$[0.47, 0.89]$}
\expandafter\def\csname N@hm-gemma-pool-agg-dgap\endcsname{0.69}
\expandafter\def\csname N@hm-gemma-pool-agg-dgap-ci\endcsname{$[0.51, 0.93]$}
\expandafter\def\csname N@hm-gemma-pool-agg-dcorr\endcsname{$-$0.59}
\expandafter\def\csname N@hm-gemma-pool-agg-dcorr-ci\endcsname{$[-0.75, -0.38]$}
\expandafter\def\csname N@hm-gemma-pool-agg-rel\endcsname{0.00}
\expandafter\def\csname N@hm-gemma-pool-abl-seed2-corr\endcsname{0.54}
\expandafter\def\csname N@hm-gemma-pool-abl-seed2-cross\endcsname{0.91}
\expandafter\def\csname N@hm-gemma-pool-abl-seed2-pwin\endcsname{61.5\%}
\expandafter\def\csname N@hm-gemma-pool-abl-seed2-gap\endcsname{0.82}
\expandafter\def\csname N@hm-gemma-pool-abl-seed2-dcross\endcsname{$-$0.07}
\expandafter\def\csname N@hm-gemma-pool-abl-seed2-dcross-ci\endcsname{$[-0.28, 0.17]$}
\expandafter\def\csname N@hm-gemma-pool-abl-seed2-dgap\endcsname{0.01}
\expandafter\def\csname N@hm-gemma-pool-abl-seed2-dgap-ci\endcsname{$[-0.21, 0.25]$}
\expandafter\def\csname N@hm-gemma-pool-abl-seed2-dcorr\endcsname{$-$0.02}
\expandafter\def\csname N@hm-gemma-pool-abl-seed2-dcorr-ci\endcsname{$[-0.15, 0.12]$}
\expandafter\def\csname N@hm-gemma-pool-abl-seed2-rel\endcsname{1.16}
\expandafter\def\csname N@hm-gemma-pool-abl-demsharp-corr\endcsname{1.18}
\expandafter\def\csname N@hm-gemma-pool-abl-demsharp-cross\endcsname{1.74}
\expandafter\def\csname N@hm-gemma-pool-abl-demsharp-pwin\endcsname{100.0\%}
\expandafter\def\csname N@hm-gemma-pool-abl-demsharp-gap\endcsname{1.79}
\expandafter\def\csname N@hm-gemma-pool-abl-demsharp-dcross\endcsname{$-$0.90}
\expandafter\def\csname N@hm-gemma-pool-abl-demsharp-dcross-ci\endcsname{$[-1.10, -0.69]$}
\expandafter\def\csname N@hm-gemma-pool-abl-demsharp-dgap\endcsname{$-$0.96}
\expandafter\def\csname N@hm-gemma-pool-abl-demsharp-dgap-ci\endcsname{$[-1.20, -0.75]$}
\expandafter\def\csname N@hm-gemma-pool-abl-demsharp-dcorr\endcsname{$-$0.66}
\expandafter\def\csname N@hm-gemma-pool-abl-demsharp-dcorr-ci\endcsname{$[-0.83, -0.47]$}
\expandafter\def\csname N@hm-gemma-pool-abl-demsharp-rel\endcsname{1.14}
\expandafter\def\csname N@hm-gemma-pool-demo-corr\endcsname{1.19}
\expandafter\def\csname N@hm-gemma-pool-demo-cross\endcsname{1.74}
\expandafter\def\csname N@hm-gemma-pool-demo-pwin\endcsname{100.0\%}
\expandafter\def\csname N@hm-gemma-pool-demo-gap\endcsname{1.78}
\expandafter\def\csname N@hm-gemma-pool-demo-dcross\endcsname{$-$0.90}
\expandafter\def\csname N@hm-gemma-pool-demo-dcross-ci\endcsname{$[-1.12, -0.70]$}
\expandafter\def\csname N@hm-gemma-pool-demo-dgap\endcsname{$-$0.96}
\expandafter\def\csname N@hm-gemma-pool-demo-dgap-ci\endcsname{$[-1.18, -0.75]$}
\expandafter\def\csname N@hm-gemma-pool-demo-dcorr\endcsname{$-$0.66}
\expandafter\def\csname N@hm-gemma-pool-demo-dcorr-ci\endcsname{$[-0.83, -0.49]$}
\expandafter\def\csname N@hm-gemma-pool-demo-rel\endcsname{1.10}
\expandafter\def\csname N@nlowest-otp\endcsname{five}
\expandafter\def\csname N@nlowest-pool\endcsname{four}
\expandafter\def\csname N@all-sel-rel-pct\endcsname{33--98\%}
\expandafter\def\csname N@qwen-sel-rel-pct\endcsname{33--59\%}
\expandafter\def\csname N@churn-human\endcsname{8.1\%}
\expandafter\def\csname N@churn-sel\endcsname{14.9--18.1\%}
\expandafter\def\csname N@churn-sel-ratio\endcsname{1.8--2.2}
\expandafter\def\csname N@churn-lpp\endcsname{7.4\%}
\expandafter\def\csname N@churn-lpp-ratio\endcsname{0.91}
\makeatother
\newcommand{\HMDYNhuman}{0.10}
\newcommand{\HMDYNbaser}{0.85}
\newcommand{\HMDYNbasedirect}{0.64}
\newcommand{\HMDYNlppdirect}{0.39--0.48}
\newcommand{\HMDYNlppr}{0.81--0.85}
\newcommand{\HMDYNbaseraw}{0.44}
\newcommand{\HMDYNlppraw}{0.21}
\newcommand{\HMDYNbasesharp}{0.70}
\newcommand{\HMDYNlppsharp}{0.28}

\newcommand{\HMDYNGbasedirect}{0.45}
\newcommand{\HMDYNGlppdirect}{0.31}
\newcommand{\HMDYNGbaseraw}{0.24}
\newcommand{\HMDYNGlppraw}{0.15}
\newcommand{\HMDYNGbasesharp}{0.31}
\newcommand{\HMDYNGlppsharp}{0.19}

\newcommand{\MATCHrt}{0.9}
\newcommand{\MATCHnun}{16}
\newcommand{\MATCHnb}{36}
\newcommand{\MATCHminp}{76\%}

\expandafter\def\csname R@off-qwen\endcsname{1.74}
\expandafter\def\csname R@off-qwen-maxrel\endcsname{1.20}
\expandafter\def\csname R@off-qwen-lpp\endcsname{1.65}
\expandafter\def\csname R@off-qwen-base\endcsname{2.51}
\expandafter\def\csname R@off-gemma\endcsname{2.46}
\expandafter\def\csname R@off-gemma-maxrel\endcsname{1.07}
\expandafter\def\csname R@off-gemma-lpp\endcsname{1.70}
\expandafter\def\csname R@off-gemma-base\endcsname{2.51}
\expandafter\def\csname R@offu-qwen\endcsname{1.75}
\expandafter\def\csname R@offu-qwen-maxrel\endcsname{1.21}
\expandafter\def\csname R@offu-gemma\endcsname{2.45}
\expandafter\def\csname R@offu-gemma-maxrel\endcsname{1.07}
\expandafter\def\csname R@seed-qwen-mean\endcsname{1.67}
\expandafter\def\csname R@seed-qwen-sd\endcsname{0.06}
\expandafter\def\csname R@seed-qwen-max\endcsname{1.76}
\expandafter\def\csname R@seed-qwen-min\endcsname{1.60}
\expandafter\def\csname R@seed-gemma-mean\endcsname{1.84}
\expandafter\def\csname R@seed-gemma-sd\endcsname{0.10}
\expandafter\def\csname R@seed-gemma-max\endcsname{2.00}
\expandafter\def\csname R@seed-gemma-min\endcsname{1.70}
\expandafter\def\csname R@para-qwen-lpp\endcsname{0.83}
\expandafter\def\csname R@para-qwen-lppmax\endcsname{0.90}
\expandafter\def\csname R@para-qwen-blmax\endcsname{0.86}
\expandafter\def\csname R@para-qwen-same\endcsname{0.95}
\expandafter\def\csname R@para-qwen-nbl\endcsname{6}
\expandafter\def\csname R@para-gemma-lpp\endcsname{1.01}
\expandafter\def\csname R@para-gemma-lppmax\endcsname{1.01}
\expandafter\def\csname R@para-gemma-blmax\endcsname{0.91}
\expandafter\def\csname R@para-gemma-same\endcsname{1.23}
\expandafter\def\csname R@para-gemma-nbl\endcsname{6}
\expandafter\def\csname R@qwen-bestbase\endcsname{2.15}
\expandafter\def\csname R@gemma-bestbase\endcsname{2.16}
\expandafter\def\csname R@ideo-qwen-rel\endcsname{0.48}
\expandafter\def\csname R@ideo-gemma-rel\endcsname{1.00}
\providecommand{\RB}[1]{\ifcsname R@#1\endcsname\csname R@#1\endcsname\else\textbf{??}\fi}